\documentclass[twoside,11pt]{article}

\usepackage{jmlr2e}

\usepackage[utf8]{inputenc} % allow utf-8 input
\usepackage[T1]{fontenc}    % use 8-bit T1 fonts
\usepackage{hyperref}       % hyperlinks
\usepackage{url}            % simple URL typesetting
\usepackage{booktabs}       % professional-quality tables
\usepackage{amsfonts}       % blackboard math symbols
\usepackage{nicefrac}       % compact symbols for 1/2, etc.
\usepackage{microtype}      % microtypography
\usepackage{xcolor}         % colors

\usepackage{microtype}
\usepackage{graphicx}
\usepackage{booktabs} % for professional tables

\usepackage{microtype}
\usepackage{graphicx}

\usepackage{natbib}
\usepackage{algorithm}
\usepackage{algorithmic}
\usepackage{paralist}
\usepackage{multirow}
\usepackage{times}
\usepackage{wrapfig}
\usepackage{url,enumerate}
\usepackage{color,xcolor}
\usepackage{makeidx}  % allows for indexgeneration
\usepackage{mathtools}
\usepackage{xspace}
\usepackage{epstopdf}
\usepackage{cite}

\usepackage{mathrsfs}
\usepackage{times}
\usepackage{enumerate}
\usepackage{color}
\usepackage{graphicx,epsfig}
\usepackage{url}
\usepackage{hyperref}
\usepackage{bm}
\usepackage{bbm}
\usepackage{upgreek}
\usepackage{cleveref}
\usepackage{multirow}
\usepackage{ulem}
\usepackage{cancel}
\usepackage{subcaption}
\usepackage{dsfont}
\usepackage{adjustbox}

\usepackage{amsmath,amssymb,xspace}

\renewcommand{\a}{{\bf a}}

\renewcommand{\c}{{\bf c}}
\renewcommand{\d}{{\rm d}}  % for derivatives

\newcommand{\n}{{\bf n}}

\newcommand{\x}{{\bf x}}

\newcommand{\Dcal}{\mathcal{D}}
\newcommand{\gp}{\mathcal{GP}}

\newcommand{\Lcal}{{\mathcal{L}}}

\newcommand{\Acal}{{\mathcal{A}}}
\newcommand{\Ocal}{{\mathcal{O}}}
\newcommand{\Pcal}{{\mathcal{P}}}
\newcommand{\Fcal}{{\mathcal{F}}}
\newcommand{\uhat}{{\widehat{u}}}

\newcommand{\N}{\mathcal{N}}  % for normal density

\renewcommand{\P}{{\bf P}}

\newcommand{\Scal}{{\mathcal{S}}}

\newcommand{\Xcal}{{\mathcal{X}}}

\newcommand{\bbeta}{\boldsymbol{\beta}}

\newcommand{\btheta}{\boldsymbol{\theta}}

\newcommand{\ben}{\begin{enumerate}}
\newcommand{\een}{\end{enumerate}}

\newcommand{\argmin}{\operatornamewithlimits{argmin}}
\newcommand{\argmax}{\operatornamewithlimits{argmax}}

\newcommand{\EE}{\mathbb{E}}
\newcommand\blfootnote[1]{%
  \begingroup
  \renewcommand\thefootnote{}\footnote{#1}%
  \addtocounter{footnote}{-1}%
  \endgroup
}

\newcommand{\ours}{{METALIC}\xspace}

\newcommand{\xpinns}{{{XPINNs}}\xspace}
\newcommand{\pinn}{{{PINN}}\xspace}
\newcommand{\pinns}{{{PINNs}}\xspace}
\newcommand{\cmt}[1]{}
\newcommand{\eg}{{\textit{e.g.},}\xspace}
\newcommand{\ie}{{\textit{i.e.},}\xspace}
\newcommand{\etc}{{\textit{etc}.}\xspace}

\newtheorem{Th}{Theorem}[section]

\newtheorem{?}[Th]{Problem}

\firstpageno{1}

\begin{document}

\title{Meta Learning of Interface Conditions for Multi-Domain Physics-Informed Neural Networks}

\author{\name Shibo Li$^\ast$ \email shibo@cs.utah.edu \\
       \addr Kahlert School of Computing\\
       University of Utah
       \AND
       \name Michael Penwarden$^\ast$ \email mpenwarden@sci.utah.edu \\
       \addr Kahlert School of Computing, Scientific Computing and Imaging Institute\\
       University of Utah
       \AND
       \name Yiming Xu \email yxu@math.utah.edu \\
		\addr Department of Mathematics\\
		University of Utah
		\AND
       \name Conor Tillinghast \email ctilling@math.utah.edu\\
		\addr Department of Mathematics\\
		University of Utah
		\AND
       \name Akil Narayan \email mpenwarden@sci.utah.edu \\
		\addr Department of Mathematics, Scientific Computing and Imaging Institute\\
		University of Utah
		\AND
       \name Robert M. Kirby \email kirby@cs.utah.edu \\
       \addr Kahlert School of Computing, Scientific Computing and Imaging Institute\\
       University of Utah
       \AND
       \name Shandian Zhe \email zhe@cs.utah.edu \\
       \addr Kahlert School of Computing\\
       University of Utah }

% \editor{Leslie Pack Kaelbling}

\maketitle

\blfootnote{$^\ast$ Equal Contribution.}

%high-dimensional output learning & multi-fdielity --> we consider the challenge task 
\begin{abstract}
Physics-informed neural networks (PINNs) are emerging as popular mesh-free solvers for partial differential equations (PDEs). Recent extensions decompose the domain,  apply different PINNs to solve the problem in each subdomain, and stitch  the subdomains at the interface. Thereby, they can further alleviate the problem complexity, reduce the computational cost, and allow parallelization. However, the performance of multi-domain PINNs is sensitive to the choice of the interface conditions. While quite a few conditions have been proposed, there is no suggestion about how to select the conditions according to specific problems. To address this gap, we propose META Learning of Interface Conditions (\ours),  a simple, efficient yet powerful approach to dynamically determine appropriate interface conditions for solving a family of parametric PDEs.  Specifically, we develop two contextual multi-arm bandit (MAB) models. The first one applies to the entire training course, and online updates a Gaussian process (GP) reward that given the PDE parameters and interface conditions predicts the performance. We prove a sub-linear regret bound for both UCB and Thompson sampling, which in theory guarantees the effectiveness of our MAB. The second one partitions the training into two stages, one is the stochastic phase and the other deterministic phase; we update a GP reward for each phase to enable different condition selections at the two stages to further bolster the flexibility and performance. We have shown the advantage of \ours on  four bench-mark PDE families. 
\end{abstract}
\section{Introduction}
Physics-informed neural networks (PINNs)~\citep{raissi2019physics} have become a popular mesh-free approach for solving partial differential equations (PDEs). They have shown successful in many scientific and engineering problems, \eg~\citep{sahli2020physics,kissas2020machine,sun2020surrogate}.
 %such as in bio-engineering~\citep{sahli2020physics,kissas2020machine}, fluids mechanics~\citep{raissi2019deep,sun2020surrogate,raissi2020hidden}, and material design~\citep{fang2019deep,liu2019multi}.  
 The recent multi-domain versions, \eg~\citep{jagtap2020conservative,jagtap2021extended},  extend PINNs with a divide-and-conquer strategy, and have attracted considerable attention.  Specifically,  they decompose the domain of interest into several subdomains, place a separate PINN to solve the PDE at each subdomain, and stitch the subdomains at the interface. In this way, the multi-domain \pinns can alleviate problem complexity, adopt simpler architectures, reduce the training cost, and enable parallel computation~\citep{shukla2021parallel}.  

However, the performance of multi-domain \pinns is sensitive to the choice of interface conditions (or regularizers) that unify the PINN solutions across different subdomains.  Quite a few conditions have been proposed, such as those that encourage solution and residual continuity~\citep{jagtap2021extended}, flux conservation~\citep{jagtap2020conservative}, gradient continuity~\citep{de2022error} and residual gradient continuity~\citep{yu2022gradient}. On one hand, naively combining all possible interface conditions does not necessarily give the optimal performance; instead, it can complicate the loss landscape, making optimization more costly and challenging. On the other hand, the best conditions can vary across problems, which are up to the problem properties. Currently, there is no suggestion about how to select interface conditions according to specific problems, bringing inconvenience and difficulties to the multi-domain practice.

To address this issue, in this paper we propose \ours,  a simple, efficient and powerful method that can dynamically determine the appropriate interface conditions for solving a family of parametric PDEs. The contributions of our work are summarized as follows. 
\begin{compactitem}
	\item \textbf{Problem Formulation and Strategy.} We formulate interface selection as a novel meta learning problem, and propose to use the multi-arm bandit (MAB) framework to address the problem. Compared with other complex/popular approaches, \eg reinforcement learning with policy gradients, MAB is simple and efficient, requiring much less training trajectories and almost no hyper-parameter tuning.  Due to the online nature, we can apply the learned MAB to select the conditions for solving new PDEs while continuously improving the model. To our knowledge, this is the first time of using MAB to address important meta learning tasks.
	\item \textbf{Method.} We develop two contextual MAB models, where we view the PDE parameters as the context, sets of interface conditions  as the arms, and the solution accuracy as the reward. The first model applies to the entire training course, and online updates a Gaussian process (GP) reward model with  Upper Confidence Bound (UCB) or Thompson sampling (TS). The second model, according to the common PINN practice,  partitions the training into two phases, the stochastic (ADAM)  and deterministic (LBFGS) phase. 	We sequentially learn a separate contextual bandit for each phase. In this way, we can select appropriate conditions for different training stages to enhance flexibility and to further improve the performance.
	\item \textbf{Theory.} We prove that our first MAB model, under the PDE parameters context and with our mixed continuous and categorical kernel, enjoys a sublinear regret bound with both UCB and TS. It means that over the course of online training, our MAB enables increasingly better interface condition selections,  and guarantees to eventually find the optimal conditions. 
	\item \textbf{Results and Analysis.} We evaluated \ours with four commonly used benchmark PDEs in PINN literature. Both of our MAB models exhibit a sublinear growth of the accumulated solution error, which is much slower than the linear growth of random arm selection.  We then examined \ours after online training. With the interface conditions determined by \ours, the multi-domain \pinns achieve solution errors one order of magnitude smaller than using randomly selected conditions, which also outperform standard single-domain \pinns with more neurons (we obtained as much as 88.5\% error reduction). Finally, we conducted a thorough analysis of the  conditions selected by \ours and found many interesting results. Many selected conditions not only reflect the physical properties of the problem, but also are tied to the specific optimization procedure for PINN training.
\end{compactitem}

\section{Background}
%\vspace{-0.13in}

\noindent\textbf{Physics-Informed Neural Networks (PINNs)} estimate PDE solutions with (deep) neural networks.\cmt{, in light of the their universal approximation ability.} Consider a PDE of the following general form, 
\begin{align}
	\Fcal[u](\x) &= h(\x), \;\;\x \in \Omega, \notag  \\ 
	u(\x) &= g(\x), \;\;\x \in \partial\Omega, \label{eq:pde}
\end{align}
where $\Fcal$ is the differential operator for the PDE, $\Omega$ is the domain, $\partial \Omega$ is the boundary of the domain, $h: \Omega \rightarrow \mathbb{R}$, and $g: \partial \Omega \rightarrow \mathbb{R}$ are the given source and boundary functions, respectively. To solve the PDE, the PINN uses a deep neural network $\uhat_{\btheta}(\x)$ to represent the solution $u$, samples $N$ collocation points $\{\x_c^i\}_{i=1}^{N}$ from $\Omega$ and $M$ points $\{\x_b^i\}_{i=1}^M$ from $\partial \Omega$, and minimizes the loss, 
\begin{align}
	\btheta^* = {\argmin}_{\btheta} \;\;\lambda_b L_b(\btheta) + L_r(\btheta), \label{eq:pinn-loss}%\btheta^* = \underset{\btheta}{\argmin} \;\;L_b(\btheta) + L_r(\btheta) \label{eq:pinn-loss}
\end{align}
where $L_b(\btheta) = \frac{1}{M}\sum_{i=1}^M \left(\uhat_{\btheta}(\x_b^i) - g(\x_b^i)\right)^2$  is the boundary term to fit the boundary condition,  $L_r(\btheta) = \frac{1}{N}\sum_{i=1}^N\left(\Fcal[\uhat_{\btheta}](\x_c^i) - h(\x_c^i)\right)^2$ is the residual term to fit the equation, and $\lambda_b>0$ is the weight of the boundary term.  One can also add an initial condition term in the loss function to fit the initial conditions (when needed).

\noindent\textbf{Multi-Domain PINNs} decompose the domain $\Omega$ into several subdomains $\Omega_1, \ldots, \Omega_K$, and assign a separate PINN $\uhat_{\btheta_k}$ to solve the PDE in each subdomain $\Omega_k$.  The loss for each PINN includes a boundary term $L_b^k(\btheta_k)$ and residual term $L_r^k(\btheta_k)$ similar to those in \eqref{eq:pinn-loss}, based on the boundary and collocation points sampled from $\partial \Omega_k$ and $\Omega_k$, respectively. In addition, to stitch together the subdomains to obtain an overall solution over $\Omega$, we introduce interface conditions into the loss to align different PINNs at the intersection of the subdomains. There have been quite a few interface conditions. Suppose $\Omega_{k} \cap \Omega_{k'} \neq \emptyset$. We sample a set of interface points $\{\x_{k,k'}^i\}_{i=1}^{J_{k,k'}} \in \Omega_{k} \cap \Omega_{k'}$. One commonly used interface condition is to encourage the solution continuity~\citep{jagtap2021extended}, 
\begin{align}
	I_1(\btheta_k, \btheta_{k'}) = \frac{1}{J_{k,k'}} \sum_{i=1}^{J_{k,k'}} \left(\uhat_{\btheta_k}(\x_{k,k'}^i) - \uhat^{\text{avg}}_{k,k'}(\x_{k,k'}^i)\right)^2 \notag 
\end{align}
where $\uhat^{\text{avg}}_{k,k'}(\x_{k,k'}^i) = \frac{1}{2}\left(\uhat_{\btheta_k}(\x_{k,k'}^i) + \uhat_{\btheta_{k'}}(\x_{k,k'}^i)\right)$. \cmt{A second one is the residual continuity~\citep{jagtap2021extended}, 
\begin{align}
	&I_2(\btheta_k, \btheta_{k'}) = \frac{1}{J_{k,k'}} \sum_{i=1}^{J_{k,k'}} \big( \left(\Fcal[\uhat_{\btheta_k}](\x_{k,k'}^i) - f(\x_{k,k'}^i)\right) \notag \\
	&- \left(\Fcal[\uhat_{\btheta_{k'}}](\x_{k,k'}^i) - f(\x_{k,k'}^i)\right) \big)^2.
\end{align}}
Other choices include the residual continuity~\citep{jagtap2021extended}, gradient continuity~\citep{de2022error}\cmt{ that encourages $\partial \uhat_{\btheta_k}(\x_{k,k'}^i)/\partial \x_{k,k'}^i$ and $\partial \uhat_{\btheta_{k'}}(\x_{k,k'}^i)/\partial \x_{k,k'}^i$ to be identical},  residual gradient continuity~\citep{yu2022gradient}, flux conservation~\citep{jagtap2020conservative}, \etc In general, the loss for each subdomain $k$ has the following form,
\begin{align}
	\Lcal^k = \lambda_b L_b^k(\btheta_k) +  L_r^k(\btheta_k) + \lambda_I  \sum_{k': \Omega_{k'} \cap \Omega_{k} \neq \emptyset} \sum_{n \in \Scal}  I_n(\btheta_k, \btheta_{k'})  \notag 
\end{align}
where $\Scal$ is the set of interface conditions, and $\lambda_I>0$ is the weight of the interface term.  The training is to minimize $\Lcal = \sum_{k=1}^K \Lcal^k$. The final solution inside each sub-domain $k$ is given by the associated PINN $\uhat_{\btheta_k}$, while on the interface, by the average of the PINNs that share the interface. 

\section{Meta Learning of Interface Conditions}
%\vspace{-0.1in}
While multi-domain \pinns have shown successes, the selection of the interface conditions remains an open and difficult problem. On one hand, naively adding all possible conditions together will complicate the loss landscape, making the optimization challenging and expensive, yet not necessarily giving the best performance.  On the other hand, different problems can demand a different set of the interface conditions as the best choice, which is up to the properties of the problem itself.  For example, in our preliminary study about a specific 2D Poisson equation (see Sec \ref{sect:prelim} in Appendix), we found the best accuracy is obtained with a novel combination of three interface conditions, giving 43.7\% error reduction as compared with combining all the interface conditions together; see Table \ref{tb:combine} in Appendix. 

Currently, there is a lack of methodologies to identify conditions for different PDEs. To address this issue, we first formulate it as a novel meta learning problem.  Specifically, we consider a parametric PDE family $\Acal$, where each PDE  in $\Acal$ is parameterized by $\bbeta \in \Xcal \subset \mathbb{R}^d$. The parameters can come from the operator $\Fcal$, the source term $h$ and/or the boundary function $g$ (see \eqref{eq:pde}). Denote by $\Scal = \{I_1, \ldots, I_{s}\}$ the full set of interface conditions. Our goal is, given a PDE parameterized by arbitrary $\bbeta \in \Xcal$,  to determine $I(\bbeta) \subseteq \Scal$ --- the best set of interface conditions --- for multi-domain \pinns to solve that PDE.

We propose to use the multi-arm bandit (MAB) framework~\citep{slivkins2019introduction} to address this problem. One might consider other complex and prevalent approaches, such as deep neural network prediction and reinforcement learning with policy gradients. However, to get well trained, these methods usually demand massive running trajectories of PINNs, which can be extremely costly. In addition, the success of these methods also rely on elaborate architecture design and intensive tuning of many hyper-parameters. By contrast,  MAB is simple and efficient, requiring much less training trajectories and (almost) no architecture design and hyper-parameter tuning.  The online nature makes the MAB straightforward to  update incrementally with new data, and is much more convenient than those heavy-duty models. 
%intro of MAB. Then talk about our setting 
\subsection{Multi-Arm Bandit for Entire Training} \label{sect:mab-1}
%\vspace{-0.1in}
%\textcolor{blue}{motivation} We first propose a multi-arm bandit (MAB) model~\citep{slivkins2019introduction} to select the interface conditions for  the entire training procedure of multi-domain \pinns. 
We first propose a MAB model to select the interface conditions for  the entire training procedure of multi-domain \pinns. 
In general, MAB considers a gambler playing $q$ slot machines (\ie arms). Pulling the lever of each machine will return  a random reward from a machine-specific probabilistic distribution, which is unknown apriori. The gambler aims to maximize the total reward earned from a series of lever-pulls across the $q$ machines.  For each play, the gambler needs to decide the tradeoff between exploiting the machine that has observed the largest expected payoff so far and exploring the payoffs of other machines. To determine PDE-specific interface conditions,  we build a contextual MAB model. We consider the PDE parameters $\bbeta \in \Xcal$ as the context, all possible combinations of the interface conditions (\ie the power set of $\Scal$)  as the arms, and the negative solution error as the reward. The problem space can be represented by a triplet $(\Xcal, \Pcal, f(\cdot, \cdot))$, where $\Xcal$ is the context space, $\Pcal$ is the action space (the power set of $\Scal$), and $f: \Xcal \times \Pcal \rightarrow \mathbb{R}$ is the reward function.  We represent each action  by an $s$-dimensional binary vector $\a$, where each element corresponds to a particular interface condition in $\Scal$. The $i$-th element $a_i = 1$ means the interface condition $i$ is selected in the action. 

%binary encoddings, kernel 
To estimate the unknown reward function $f(\cdot, \cdot)$, we assign a Gaussian process (GP) prior,
\begin{align}
f \sim \gp\left(0, \kappa\left([\bbeta, \a], [\bbeta', \a']\right)\right)  
\end{align}
where $\kappa(\cdot, \cdot)$ is a kernel (covariance) function. Considering the categorical nature of the action input, we design a product kernel, 
\begin{align}
	 \kappa\left([\bbeta, \a], [\bbeta', \a']\right)  = \kappa_1(\bbeta, \bbeta') \kappa_2(\a, \a')
\end{align}
where $\kappa_1(\bbeta, \bbeta') = \exp(-\tau_1 \|\bbeta - \bbeta'\|^2)$ is the square exponential (SE) kernel for continuous PDE parameters, and 
\begin{align}
	\kappa_2(\a, \a') = \exp\left(\tau_2 \cdot \frac{1}{s}\sum_{i=1}^s \mathds{1}(a_i = a'_i)\right)
\end{align}
where $\mathds{1}(\cdot)$ is the indicator function. Hence, the similarity between actions is based on the overlap ratio of the selected interface conditions, which is natural and intuitive. As in the standard GP regression, the observed reward $r$ is then sampled from a Gaussian noise model, 
\begin{align}
	r \sim \N(r | f, \sigma^2_0)
\end{align}
where $\sigma^2_0>0$. The noise model is to capture the extraneous randomness such as stochastic training and float rounding. 

To learn the MAB, each step we randomly sample a context $\bbeta$ from $\Xcal$, and then select an action $\a$, \ie a set of interface conditions,  according to the current GP reward model. We then run the multi-domain \pinns with the interface conditions to solve the PDE parameterized by $\bbeta$. We evaluate the negative solution error $\xi$ as the received reward. We add the new data point $([\bbeta, \a], -\xi)$ into the current training set, and retrain (update) the GP reward model. We repeat this procedure until a given maximum number of trials (plays) is done. To fulfill a good exploration-exploitation tradeoff, we use the Upper Confidence Bound (UCB)~\citep{auer2002using,srinivas2010gaussian} or Thompson sampling (TS)~\citep{thompson1933likelihood,chapelle2011empirical} to select the action at each step. Specifically, denote the current predictive distribution of the GP surrogate by $$p(\widehat{f}|\Dcal, \bbeta, \a) = \N\left(\widehat{f}|\mu(\a, \bbeta), \sigma^2(\a, \bbeta)\right)$$ where $\Dcal$ is the accumulated training set so far. The UCB score is $$\text{UCB}(\a) = \mu(\a, \bbeta) + {c_t}^{1/2} \cdot \sigma(\a, \bbeta)$$ where $c_t>0$ is a coefficient at step $t$, and the TS score is sampled from the predictive distribution, $$\text{TS}(\a) \sim p(\widehat{f}|\Dcal, \bbeta, \a).$$ We can see that both scores integrate the predictive mean (which reflects the exploitation part)  and the variance information (exploration part). We evaluate the score for each action $\a \in \Pcal$ (UCB or TS), and select the one with the highest score (corresponding to the best tradeoff). 

In the online scenario, we can keep using UCB or TS to select the action for incoming new PDEs while improving the reward  model according to the measured solution error. When the online playing is done and we no longer conduct exploration to update our model, given a new PDE (say, indexed by $\bbeta^*$), we evaluate the predictive mean $\mu$ given $\bbeta^*$ and every $\a \in \Pcal$. We then  select the one with the largest predictive mean (reward estimate). We use the corresponding interface conditions to run the multi-domain \pinns to solve the PDE. Our MAB learning is summarized in Algorithm \ref{algo:mab-1}.

To ensure our MAB is capable of finding the optimal interface conditions for different PDEs, we perform regret analysis. Consider a sequence of PDE parameters (\ie context) in the online playing, $\{\bbeta_t\}$. Denote by $\a^*_t$ and $\a_t$ the optimal and MAB-selected interface condition set for each $\bbeta_t$, respectively. We define an instantaneous regret $\zeta_t = f(\bbeta_t, \a_t) - f(\bbeta_t, \a^*_t)$.\cmt{, where $f(\cdot)$ is the solution accuracy (negative solution error) of the multi-domain PINN with trained with the given interface conditions.} Then we can analyze the accumulated regret up to step $T$ over the context sequence $\{\bbeta_t\}$, namely, $R_T = \sum_{t=1}^T\zeta_t$. Our MAB guarantees a sublinear regret bound for both UCB and TS.   
\begin{Th}\label{th:regret}
	For $\delta>0$, take $c_t$ in the UCB  as
	\begin{align*}
		&c_t = 2\log\left(\frac{2^s\pi^2t^2}{6\delta}\right)& t\in \mathbb{N}
	\end{align*}
	where $d$ is the number of PDE parameters, and $s$ is the total number of interface conditions. 
	Conditioning on every context sequence $\{\bbeta_t\}$, let $\{\a_t\}$ be the action selected by the UCB score under the above choice of $\{c_t\}$. 
	Then, with probability at least $1-\delta$, the regret $R_T$ satisfies
	\begin{align}
		&R_T\lesssim  \sqrt{\frac{2^{s}T(\log T)^{d+1}\log\left(\frac{2^sT^2}{\delta}\right)}{\log (1+\sigma_0^{-2})}}& T = 1, 2, \cdots, \label{ucb1}
	\end{align}
	where the implicit constant is absolute (does not depend on $\{c_t\}$ but depends on the domain $\mathcal X$). 
	In particular, 
	\begin{align}
		\EE[R_T]\lesssim  \sqrt{\frac{2^{s}T(\log T)^{d+1}\log\left(2^sT^2\right)}{\log(1+\sigma_0^{-2})}},\label{ucb2-main}
	\end{align}
	Moreover, \eqref{ucb2-main} holds also for Thompson sampling. 
\end{Th}
%From Theorem \ref{th:regret}, 
We can see $\lim\limits_{T \rightarrow \infty} \frac{\EE[R_T]}{T} = 0$, \ie the average instantaneous regret converges to zero. Since $R_T\ge0$, it means that our MAB is able to find the optimal interface conditions (almost surely) for every possible sequence of PDE parameters with enough long run.

%At the test stage, we no longer conduct exploration to update our model. Given a new PDE (say, indexed by $\bbeta^*$), we evaluate the predictive mean $\mu$ given $\bbeta^*$ and every $\a \in \Pcal$. We then  select the one with the largest predictive mean (reward). We use the corresponding interface conditions to train the \xpinn to solve the PDE. Our MAB learning is summarized in Algorithm \ref{algo:mab-1}. 
\begin{figure}
	\centering
	\includegraphics[width=\linewidth]{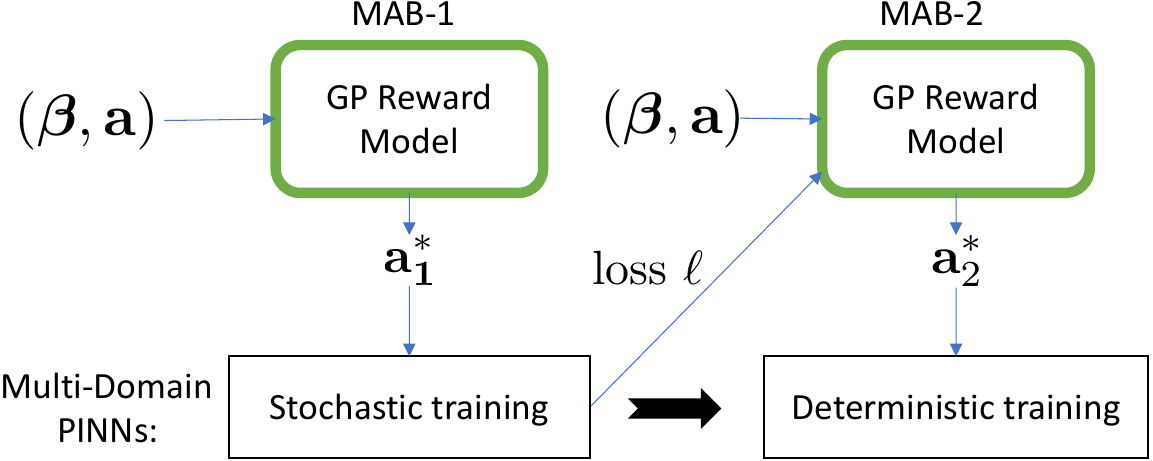}
%	\vspace{-0.25in}
	\caption{\small The illustration of the sequential MAB model.} \label{fig:model}
	%\vspace{-0.25in}
\end{figure}
\subsection{Sequential Multi-Arm Bandits}
%\vspace{-0.1in}
In practice, to achieve good and reliable performance, the training of \pinns is often divided into two stages~\citep{lu2021deepxde}. The first phase is stochastic training, typically with ADAM optimizer~\citep{kingma2014adam}, to find a nice valley of the loss landscape. The second phase is deterministic optimization, typically with L-BFGS, to ensure convergence to the (local) minimum. Due to the different nature of the two phases, the ideal interface conditions can vary as well. To enable more flexible choices so as to further improve the performance, we propose a sequential MAB model, as illustrated in Fig. \ref{fig:model}.  Specifically, for each training phase, we introduce a MAB similar to Sec \ref{sect:mab-1}, which  updates a separate  GP reward model. To coordinate the two MAB's and to optimize the final accuracy, the reward of the first MAB, denoted by $r_1$,   includes not only the negative solution error after the stochastic training phase, but also a discounted error after the  second phase, 
\begin{align}
r_1 = -\xi_1 + \gamma \cdot  (-\xi_2) \label{eq:reward-mab-1}
\end{align}
where $\gamma$ is the discount factor, and $\xi_1$ and $\xi_2$ are the solution errors after the stochastic and deterministic training phases, respectively. In this way, the influence of the interface conditions at the first training phase on the final solution accuracy is also integrated into the learning of the reward model. Next, we expand the context of the second MAB with the training loss value $\ell$ after the first phase. \cmt{The system state includes both the PDE parameters $\bbeta$ and the loss value $\ell$, which together with the action $\a$ constitute the input to the GP reward model of the second MAB.  }In this way, the training status of the first phase is also used to determine the interface conditions for the second phase. The learning of the sequential MAB's is summarized in Algorithm \ref{algo:mab-2}. 

%To learn our sequential MAB model, each step we randomly sample the PDE parameters $\bbeta \in \Xcal$. Then based on the GP reward model of the first MAB, we use UCB (or TS) to select an action (\ie the first set of  interface conditions), with which we conduct the stochastic training of the multi-domain \pinns. We then feed the training loss $\ell$, PDE parameters $\bbeta$ and every possible action into the GP reward model of the second MAB. We select the second set of interface conditions, with which to continue the training of the multi-domain \pinns using deterministic optimization. We evaluate the solution error after each phase, and obtain the reward and new examples for the two GP reward models. We update these models accordingly. \cmt{The test stage is similar to that of our single MAB model in Sec. \ref{sect:mab-1}.} The learning of the sequential MABs is summarized in Algorithm \ref{algo:mab-2}. 

\noindent \textbf{Algorithm Complexity.} The time complexity of both MAB algorithms is $\Ocal(TR + \sum_{t=1}^T t^3)$ where $T$ is the total number of iterations, $R$ is the complexity of multi-domain PINNs, and $\Ocal(\sum_{t=1}^T t^3)$ is the total time complexity of updating the  GP reward model to $T$. In practice,  $T$ is typically chosen as a few hundreds (see the experimental section). Under such a choice, running multi-domain PINNs is much more costly than GP training, and  the complexity is dominant by $\Ocal(TR)$. Hence, the time complexity is linear in the number of iterations. The space complexity of our algorithm is $\Ocal(C + T^2)$, including the storage of the GP reward model and the multi-domain PINN (with constant complexity $\Ocal(C)$). 
\begin{algorithm}[t]
	\small 
	\caption{\ours-single($T$)}                                 
	\begin{algorithmic}[1]                    % enter the algorithmic environment
		\STATE Initialize the GP reward model, and $\Dcal \leftarrow \emptyset$. 
		\FOR{$t=1 \ldots T$}%\REPEAT
		\STATE Randomly sample the PDE parameters $\bbeta \in \Xcal$.
		\STATE For each action $\a \in \Pcal$, compute the predictive distribution of the GP reward model, $\N\left(\mu(\a, \bbeta), \sigma^2(\a, \bbeta)\right)$. %$p(r(\bbeta, \a)|\Dcal) = \N\left(\mu(\a, \bbeta), \sigma^2(\a, \bbeta)\right)$
		\STATE Compute the UCB score: $\text{UCB}(\a) = \mu+ \sqrt{c_t} \cdot \sigma$ or TS score: $\text{TS}(\a) \sim \N(\mu, \sigma^2)$.
		\STATE $\a^* = \argmax_{\a \in \Pcal} \text{UCB}(\a)$ or $\a^* = \argmax_{\a \in \Pcal} \text{TS}(\a)$.
		\STATE Use the interface conditions of $\a^*$ to train the multi-domain \pinns to solve the PDE parameterized by $\bbeta$, and evaluate the solution error $\xi$.
		\STATE $\Dcal \leftarrow \Dcal \cup \{\left([\a^*, \bbeta], -\xi\right)\}$.
		\STATE Retrain the GP reward model on $\Dcal$.
		\ENDFOR %\UNTIL{$T$ iterations are done}
	\end{algorithmic}\label{algo:mab-1}
	%\vspace{-0.05in}
\end{algorithm}

\begin{algorithm}[t]
	\small 
	\caption{\ours-seq($\gamma$, $T$)}                                    
	\begin{algorithmic}[1]                    % enter the algorithmic environment
		\STATE Initialize two GP reward models. Set their training sets $\Dcal_1$ and $\Dcal_2$ to empty. 
		\REPEAT
		\STATE Randomly sample $\bbeta \in \Xcal$.
		\STATE Based on the predictive distribution of the first GP model, use UCB or TS to select the best action $\a^*_1$. 
		\STATE Use the interface conditions of  $\a^*_1$ to train the multi-domain \pinns with ADAM. Evaluate the error $\xi_1$ for solving the PDE parameterized by $\bbeta$. 
		\STATE Given the current training loss $\ell$ and $\bbeta$, compute the predictive distribution of the second GP reward model for each action, and use UCB or TS to select the best action $\a^*_2$. 
		\STATE Use the interface conditions of  $\a^*_2$ to continue the training with L-BFGS. Evaluate the solution error $\xi_2$. 
		\STATE $\Dcal_1 \leftarrow \Dcal_1 \cup \{\left([\a^*_1, \bbeta], -\xi_1 - \gamma \xi_2\right)\}, \Dcal_2 \leftarrow \Dcal_2 \cup \{([\a^*_2, \bbeta, \ell], -\xi_2)\}$.
		\STATE Retrain the two GP models on $\Dcal_1$ and $\Dcal_2$, respectively. 
		\UNTIL{$T$ iterations are done}
	\end{algorithmic}\label{algo:mab-2}
%	\vspace{-0.05in}
\end{algorithm}
%\vspace{-0.1in}
%meta learning,  discussion of our method, the advantage of Reinforcement learning & PINNs and XPINNs 
\section{Related Work}\label{sect:rel}
%\vspace{-0.1in}

As an alternative to mesh-based numerical methods, PINNs have had many success stories, \eg~\citep{raissi2020hidden, chen2020physics, sirignano2018dgm, zhu2019physics, geneva2020modeling, sahli2020physics}. Multi-domain \pinns, \eg~\xpinns\citep{jagtap2021extended} and cPINNs~\citep{jagtap2020conservative}, extend \pinns based on domain decomposition and use a set of \pinns to solve the PDE in different subdomains. To stitch together the subdomains, \xpinns used solution continuity and residual continuity as the interface conditions. Other conditions are also available, such as the flux conservation in cPINNs~\citep{jagtap2020conservative}, the gradient continuity~\citep{de2022error}, and the residual gradient continuity in gPINNs~\citep{yu2022gradient}. Recently, \citet{hu2021extended} developed a theoretical understanding on the convergence and generalization properties of \xpinns, and examined  the trade-off between  \xpinns and \pinns. 

Meta learning \citep{schmidhuber1987evolutionary, naik1992meta,thrun2012learning} is an important topic in machine learning. The existing works can be roughly attributed to  three categories:  (1) metric-learning that learns a metric space with which the tasks can make predictions via matching the training points, \eg nonparametric nearest neighbors \citep{koch2015siamese, vinyals2016matching, snell2017prototypical, oreshkin2018tadam, allen2019infinite}, (2) learning black-box models (\eg neural networks) that map the task dataset and hyperparameters to the optimal model parameters or parameter updating rules, \eg  \citep{andrychowicz2016learning, ravi2016optimization, santoro2016meta, wang2016learning, munkhdalai2017meta, mishra2017simple}, and   (3) bi-level optimization where the outer level optimizes the hyperparameters and the inner level optimizes the model parameters given the hyperparameters~\citep{finn2017model, finn2018learning, bertinetto2018meta, lee2019meta,zintgraf2019fast, li2017meta, zhou2018deep}. The bi-level optimization approaches are often restricted by the high cost of computing the meta-gradient (\eg w.r.t the model initialization) via computational graphs. The issue has been recently addressed by~\citep{li2023meta} that uses gradient flows to model the inner-optimization and adjoint state methods to compute the meta-gradients. Recently, \citet{penwarden2023metalearning} developed the first method to meta learn the initialization for PINNs. 

Multi-arm bandit is a classical online decision making framework~\citep{lai1985asymptotically,auer2002finite,auer2002nonstochastic,mahajan2008multi,bubeck2012regret}, and have numerous applications, such as online advertising~\citep{avadhanula2021stochastic}, collaborative filtering~\citep{li2016collaborative}, clinical trials~\citep{aziz2021multi} and robot control~\citep{7294140}. To our knowledge, our work is the first to use MAB for meta learning of task-specific hyperparameters, which is advantageous in its simplicity and efficiency. MAB can be viewed as an instance of reinforcement learning (RL)~\citep{sutton2018reinforcement}, but it only needs to estimate a reward function online. While one can design more expressive RL models to meanwhile learn a Markov decision process (in MAB, we simply use UCB or TS), it often demands we run a massive number of \pinn training trajectories, which is much more expensive. The model estimation is also much more challenging. 
%\vspace{-0.1in}
\section{Experiment}

To evaluate \ours, we considered four benchmark equation families. We list the equations and domain decomposition settings in the following. 

\noindent \textbf{The Poisson Equation.} First, we considered a 2D Poisson equation with a parameterized source function, 
\begin{align}
	u_{xx} + u_{yy} = \widetilde{f}(x, y;s)
\end{align}
where $(x, y) \in [0, 1] \times [0, 1]$,  $\widetilde{f}(x, y;s) = f(x, y;s)/\max_{x,y} f(x, y; s)$, and
\begin{align}
	&f(x, y; s) = \left[\text{erf}( (x - 0.25)s) - \text{erf}((x- 0.75)s)\right] \notag \\
	&\cdot \left[\text{erf}((y - 0.25 )s) - \text{erf}((y-0.75)s)\right], 
\end{align}
where $\text{erf}(z)= \frac{2}{\sqrt{\pi}}\int_0^z e^{-t^2} \d t$, and $s \in [0, 50]$ is called the sharpness parameter that controls the sharpness of the interior square in the source. We used Dirichlet boundary conditions, and ran a finite difference solver to obtain an accurate ``gold-standard'' solution. To run multi-domain \pinns, we split the domain into two subdomains, where the interface is a line at $y = 0.5$. We visualize an exemplar solution and the subdomains, including the sampled boundary and collocation points in Fig. \ref{fig:poisson-vis}.   
\begin{figure}
	\centering
	\setlength\tabcolsep{0pt}
	\begin{tabular}[c]{cc}
		\setcounter{subfigure}{0}
		\begin{subfigure}[t]{0.4\textwidth}
			\centering
			\includegraphics[width=\textwidth]{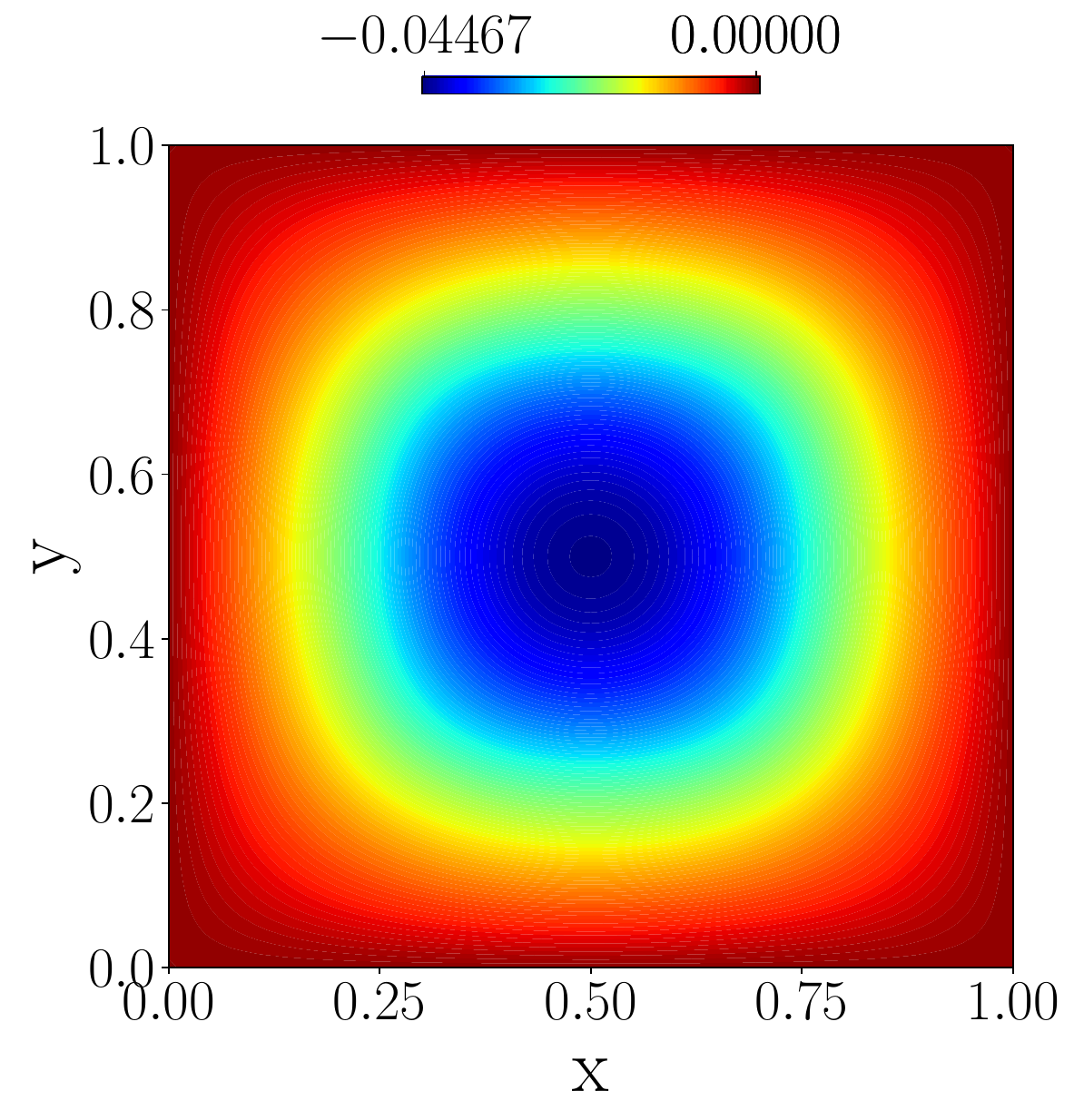}
			\caption{\small Solution at $s=20$}
		\end{subfigure} &
		\begin{subfigure}[t]{0.4\textwidth}
			\centering
			\includegraphics[width=\textwidth]{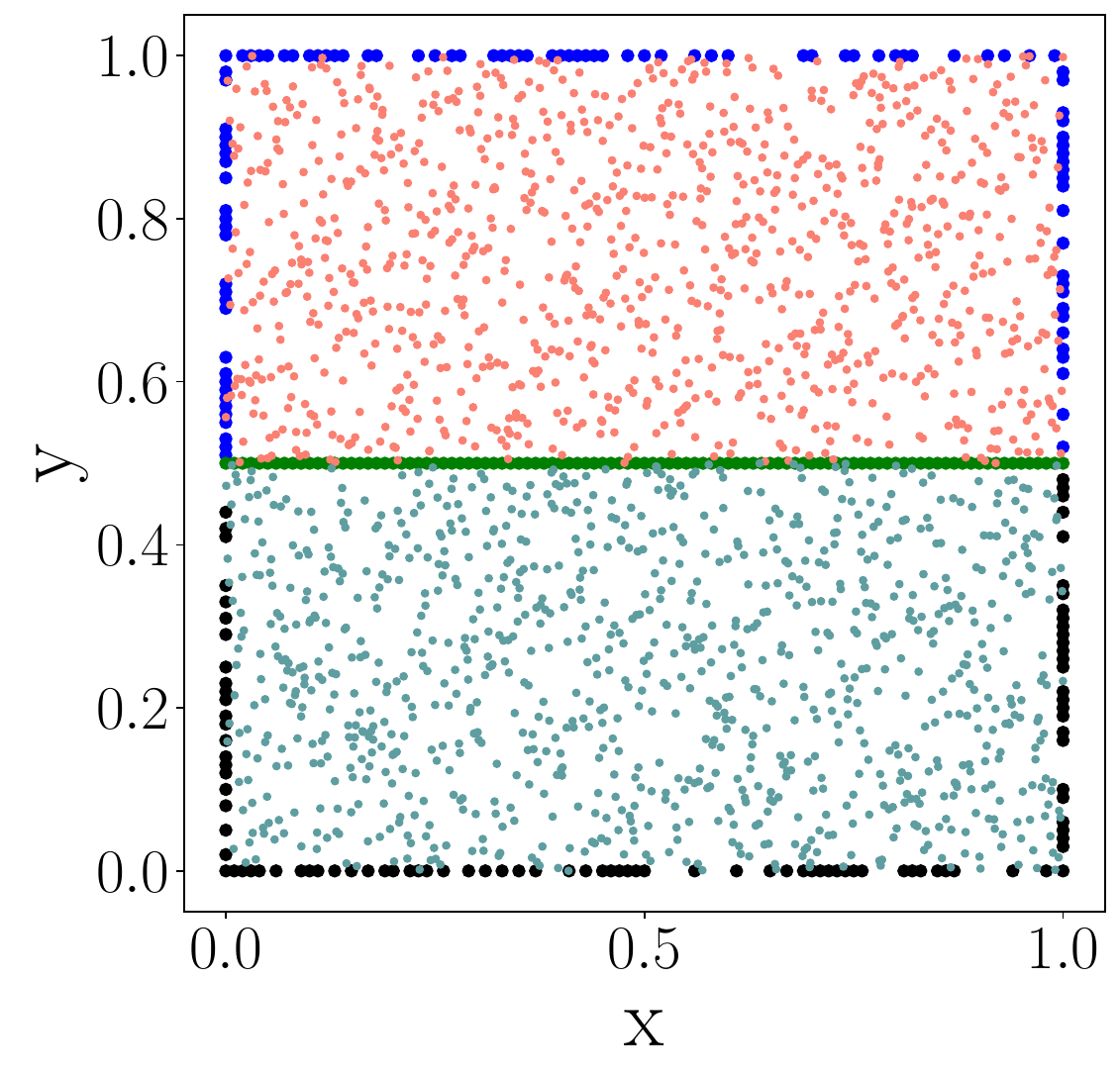}
			\caption{\small Subdomains}
		\end{subfigure}
	\end{tabular}
	\caption{\small The Poisson equation. The interface is the green line at $y=0.5$. Blue and black dots show the sampled boundary points in each subdomain, and the internal dots (red and cyan) are the sampled collocation points inside each subdomain.} \label{fig:poisson-vis}
\end{figure}

%\subsection{Advection Equation}\label{sect:advec}
\noindent \textbf{Advection Equation.} We next considered a 1D advection (one-way wave) equation, $$ u_t + \beta u_x = 0$$ where $x \in [0, 2\pi]$, $t \in [0, 1]$, and $\beta$ is the PDE parameter denoting the wave speed.  We used Dirichlet boundary conditions, and the solution has an analytical form, $u(x, t) = h(x - \beta t)$ were $h(x)$ is the initial condition (which we select as $h(x)=\sin(x)$). For domain decomposition, we split the domain at $t= 0.5$ to obtain two subdomains. Fig. \ref{fig:advec-vis} shows an exemplar solution  and the subdomains with the interface. 
\begin{figure}
	\centering
	\setlength\tabcolsep{0pt}
	\begin{tabular}[c]{cc}
		\setcounter{subfigure}{0}
		\begin{subfigure}[t]{0.4\textwidth}
			\centering
			\includegraphics[width=\textwidth]{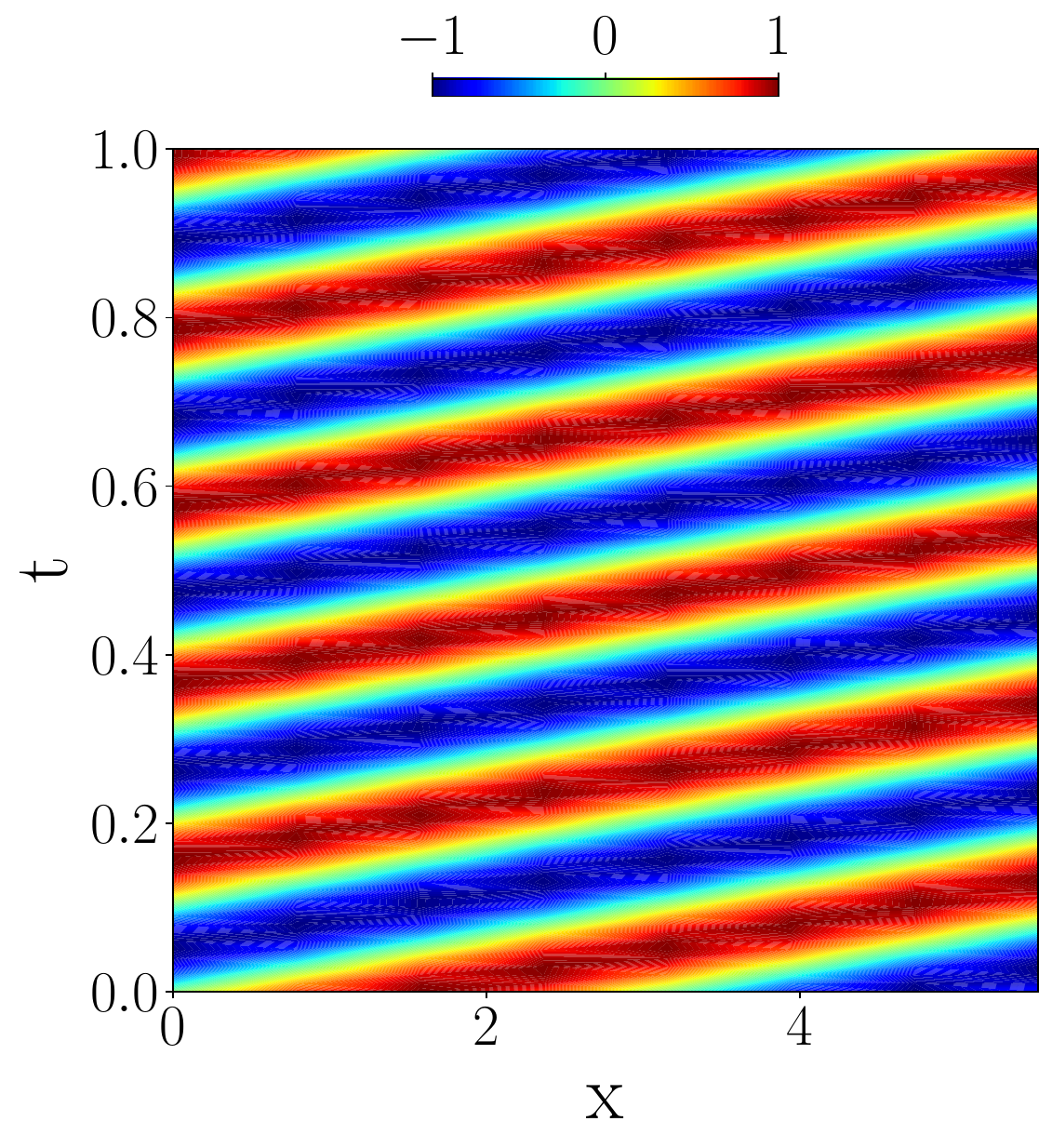}
			\caption{\small Solution at $\beta=30$ }
		\end{subfigure} &
		\begin{subfigure}[t]{0.4\textwidth}
			\centering
			\includegraphics[width=\textwidth]{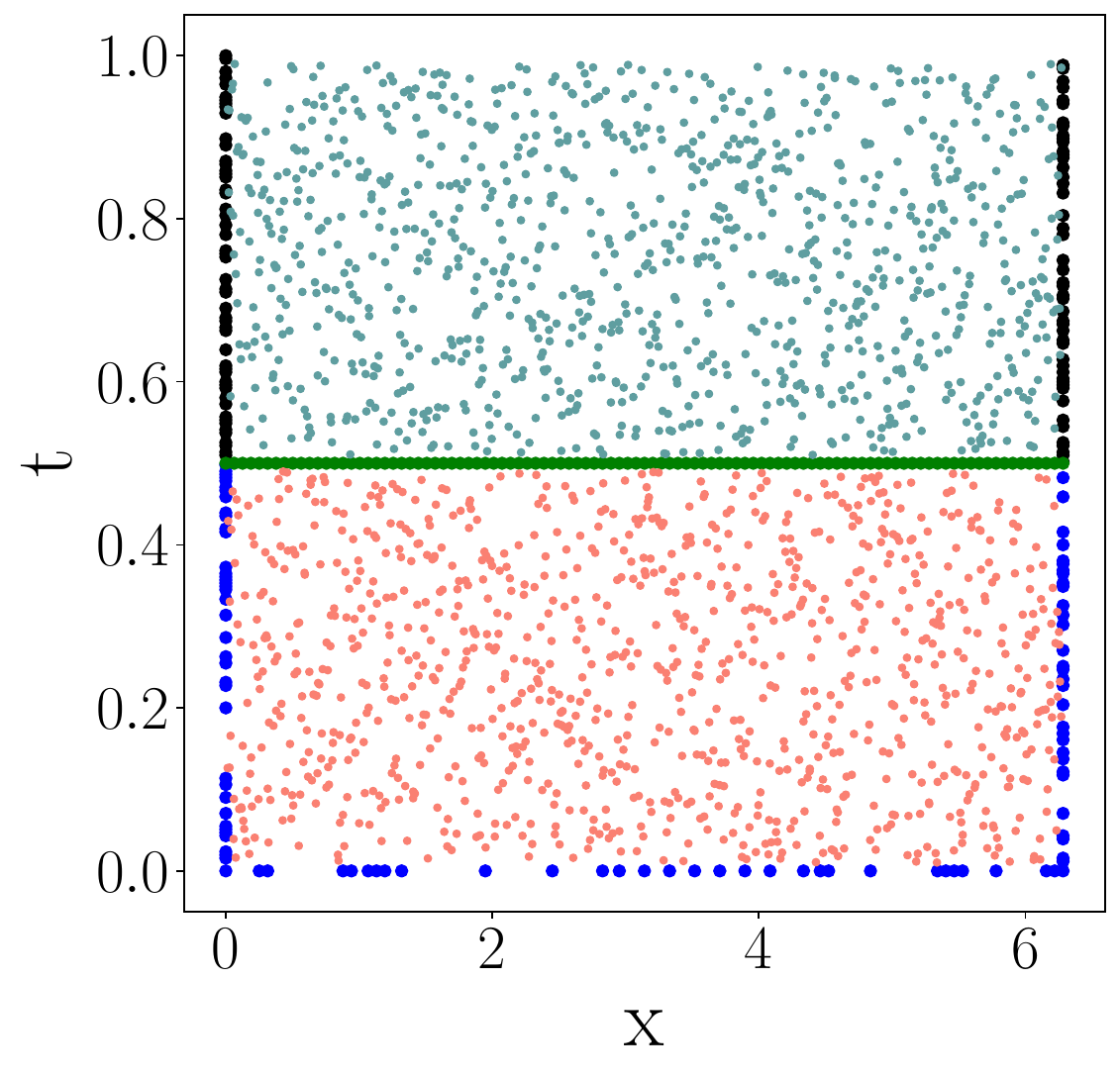}
			\caption{\small Subdomains}
		\end{subfigure}
	\end{tabular}
	\caption{\small Advection equation. The interface is the green line at $t=0.5$.} \label{fig:advec-vis}
\end{figure}

%\subsection{Reaction Equation}\label{sect:react}
\noindent \textbf{Reaction Equation.} Third, we evaluate a 1D reaction equation, $$ u_t - \rho u(1-u) = 0 $$ where $\rho$ is the reaction coefficient (ODE parameter), $x \in [0, 2\pi]$, $t \in [0, 1]$ and $u(x, 0) = e^{-\frac{(x-\pi)^2}{2 (\pi/4)^2}}$. The exact solution is $u(x,t) = u(x, 0)\cdot[e^{\rho t}/\left(u(x,0)e^{\rho t} + 1 - u(x, 0)\right)]$. We split the domain at $t=0.5$ to obtain two subdomains. Although not required for well-posedness of the ODE system, because we are solving for the PINN space-time field $u(x,t)$, we use the exact solution to define a boundary loss term.  This enhances training without compromising the time partitioning we wish to highlight. We show a solution example and the subdomains in Fig. \ref{fig:reaction-vis}.

%\subsection{Burger's Equation}\label{sect:burgers}
\noindent \textbf{Burger's Equation.} Fourth, we considered the viscous Burger's equation,  $$u_t + u u_x = \nu u_{xx}$$ where $\nu \in [0.001, 0.05]$ is the viscosity (PDE parameter), $x \in [-1, 1]$,  $t \in [0, 1]$, and $u(x, 0) = -\sin(\pi x)$. We ran a numerical solver to obtain an accurate ``gold-standard'' solution. To decompose the domain, we take the middle portion that  includes the shock waves as one subdomain, namely, $\Omega_1: x\in [-0.1, 0.1], t \in [0, 1]$, and the remaining as the other subdomain, $\Omega_2: x \in [-1, -0.1] \cap [0.1, 1], t \in [0, 1]$. Hence, the interface consists of two lines. See Fig. \ref{fig:burgers-vis} for the illustration and solution example. 

\begin{figure}
	\centering
	\setlength\tabcolsep{0pt}
	\begin{tabular}[c]{cc}
		\setcounter{subfigure}{0}
		\begin{subfigure}[t]{0.4\textwidth}
			\centering
			\includegraphics[width=\textwidth]{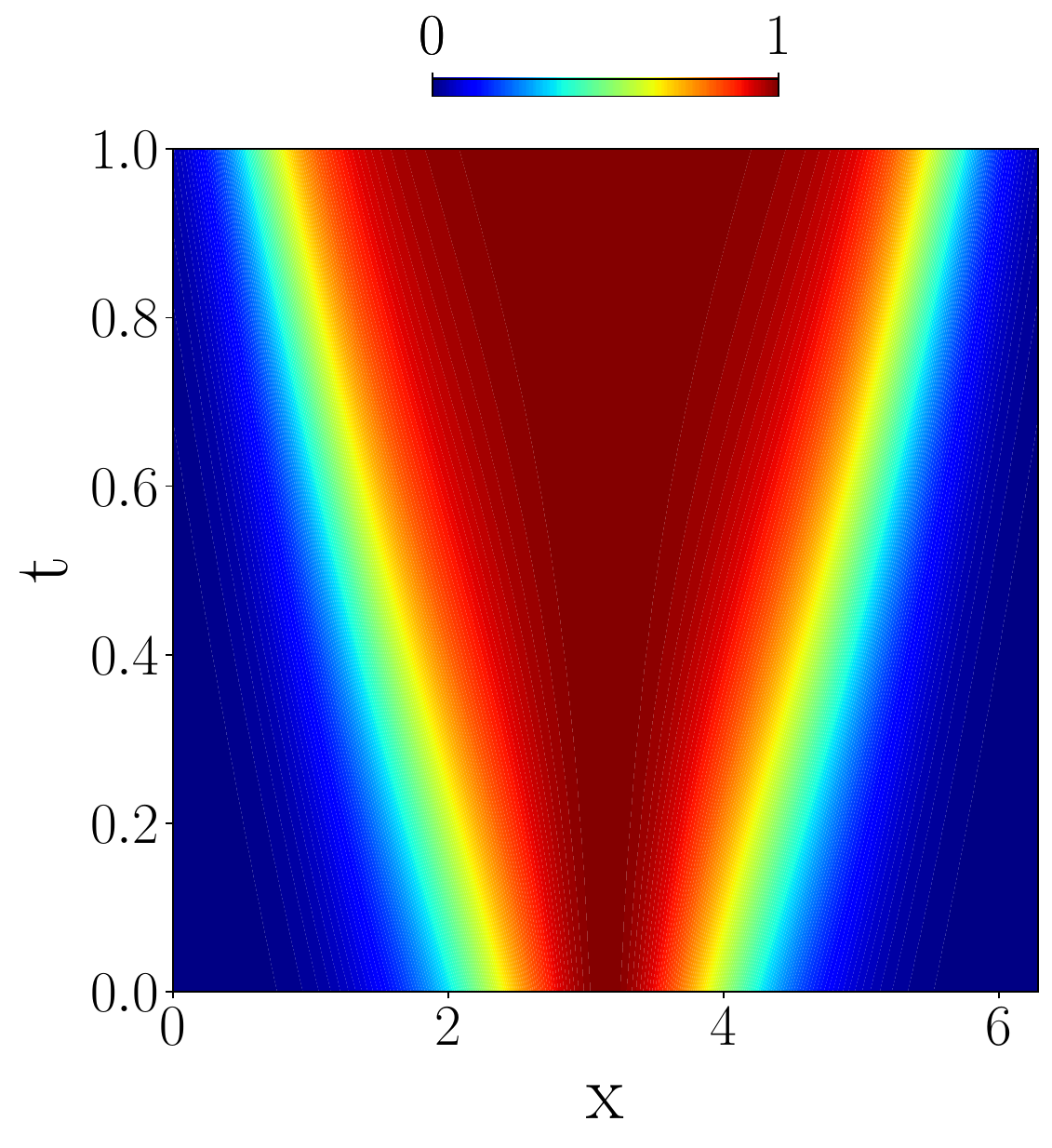}
			\caption{\small Solution at $\rho=5.0$}
		\end{subfigure} &
		\begin{subfigure}[t]{0.4\textwidth}
			\centering
			\includegraphics[width=\textwidth]{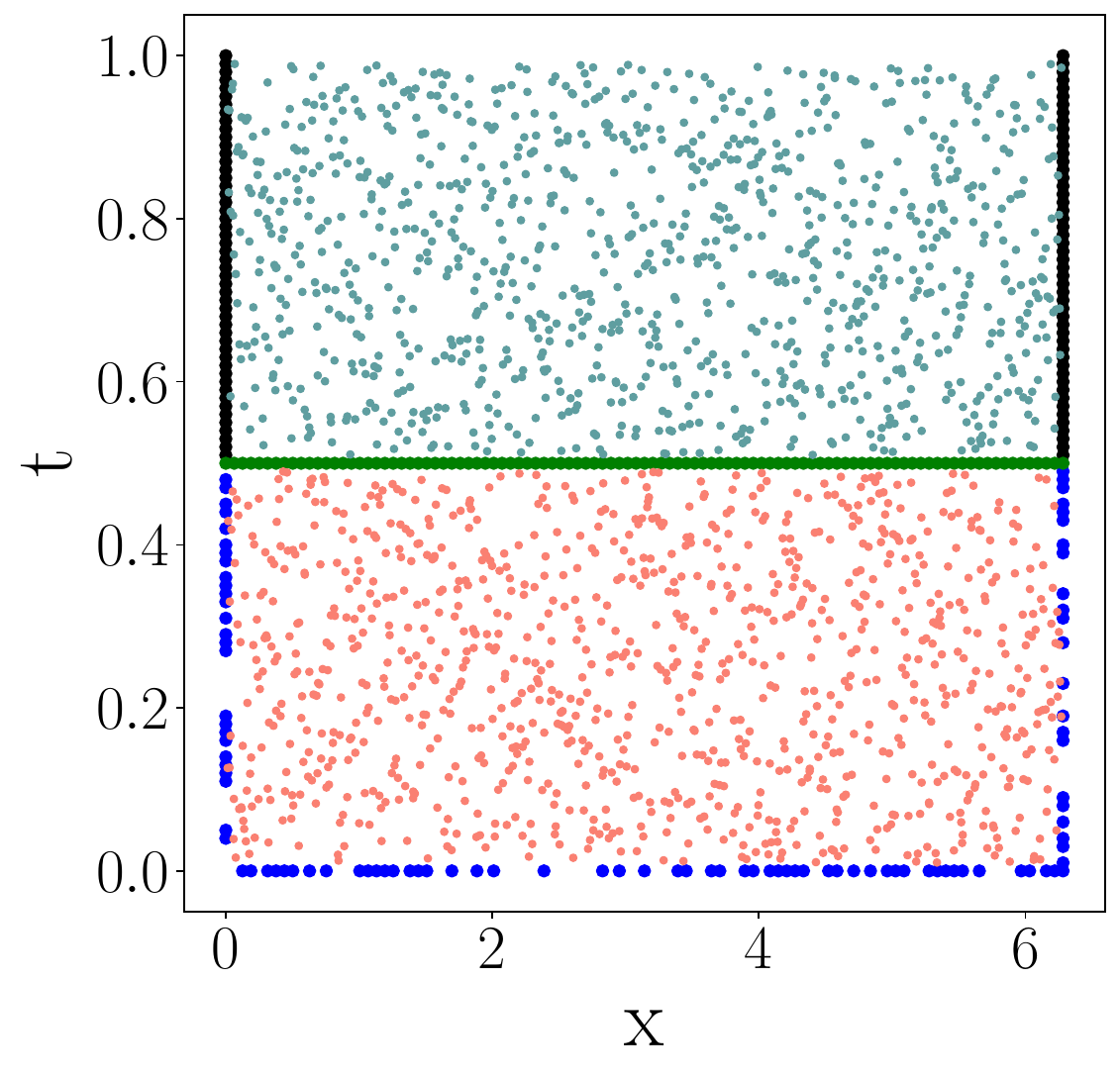}
			\caption{\small Subdomains}
		\end{subfigure}
	\end{tabular}
	\caption{\small Reaction equation. The interface is at $t=0.5$.} \label{fig:reaction-vis}
\end{figure}

\begin{figure}
	\centering
	\setlength\tabcolsep{0pt}
	\begin{tabular}[c]{cc}
		\setcounter{subfigure}{0}
		\begin{subfigure}[t]{0.4\textwidth}
			\centering
			\includegraphics[width=\textwidth]{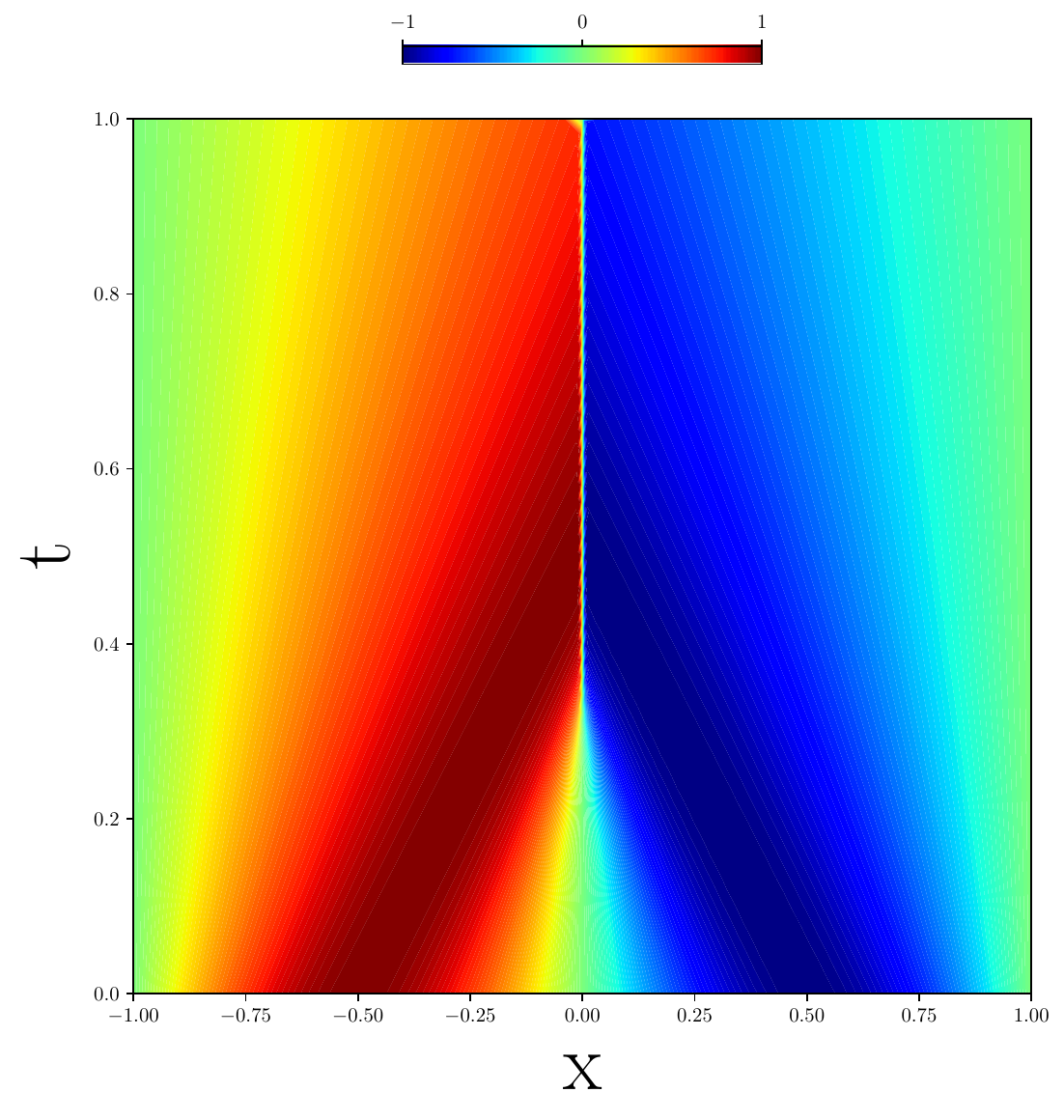}
			\caption{\small Solution at $\nu=0.001$ }
		\end{subfigure} &
		\begin{subfigure}[t]{0.4\textwidth}
			\centering
			\includegraphics[width=\textwidth]{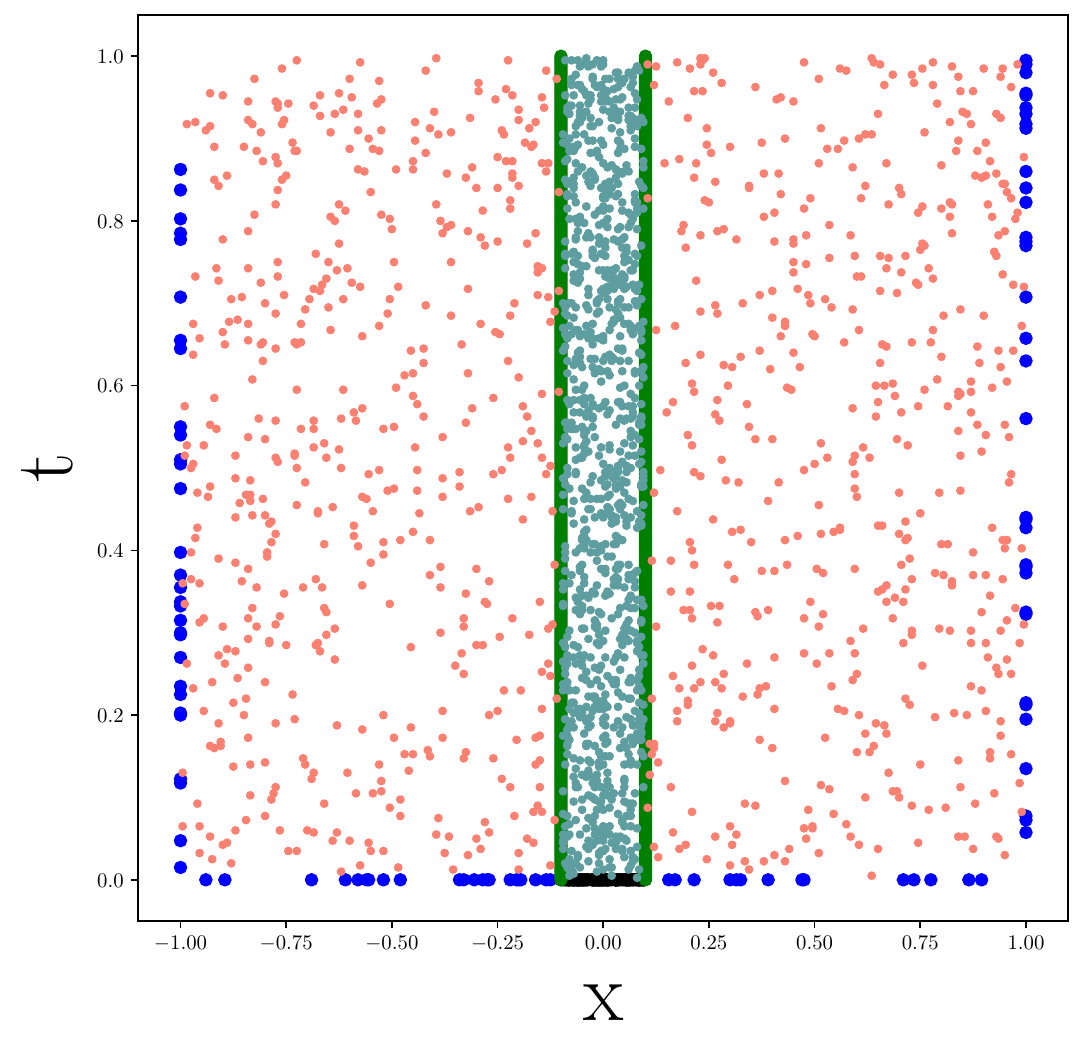}
			\caption{\small Subdomains}
		\end{subfigure}
	\end{tabular}
	\caption{\small Burger's equation. The interfaces are at $x=-0.1$ and $x=0.1$. The middle portion (filled with cyan dots) is the first subdomain, and the remaining parts constitute the  second subdomain.} \label{fig:burgers-vis}
	% 	\vspace{-0.2in}
\end{figure}
\begin{figure*}
	% 	\vspace{-0.1in}
	\centering
	\setlength\tabcolsep{0pt}
	\begin{tabular}[c]{cccc}
		\setcounter{subfigure}{0}
		\begin{subfigure}[t]{0.24\textwidth}
			\centering
			\includegraphics[width=\textwidth]{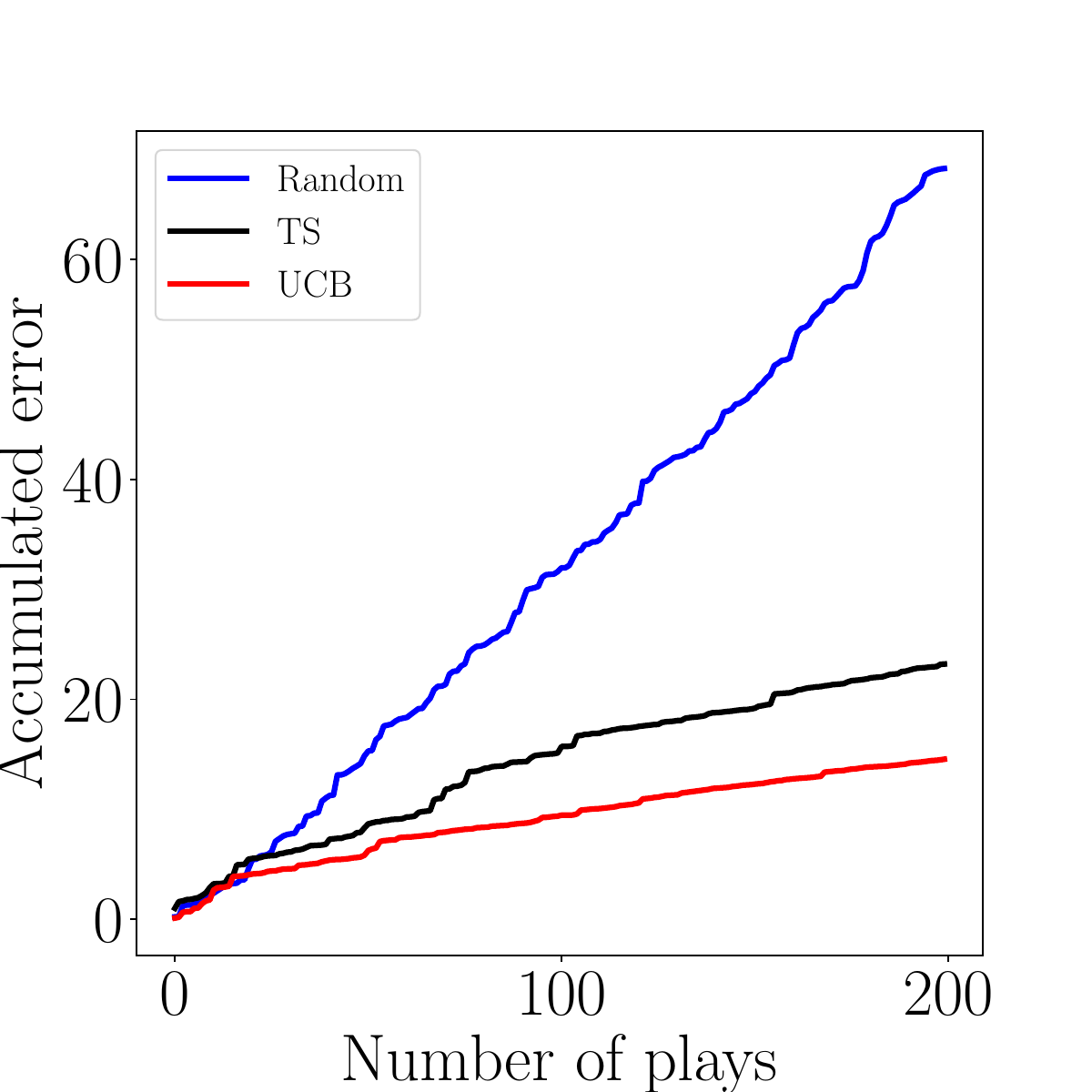}
			\caption{\small Poisson equation}
		\end{subfigure} 
		&
		\begin{subfigure}[t]{0.24\textwidth}
			\centering
			\includegraphics[width=\textwidth]{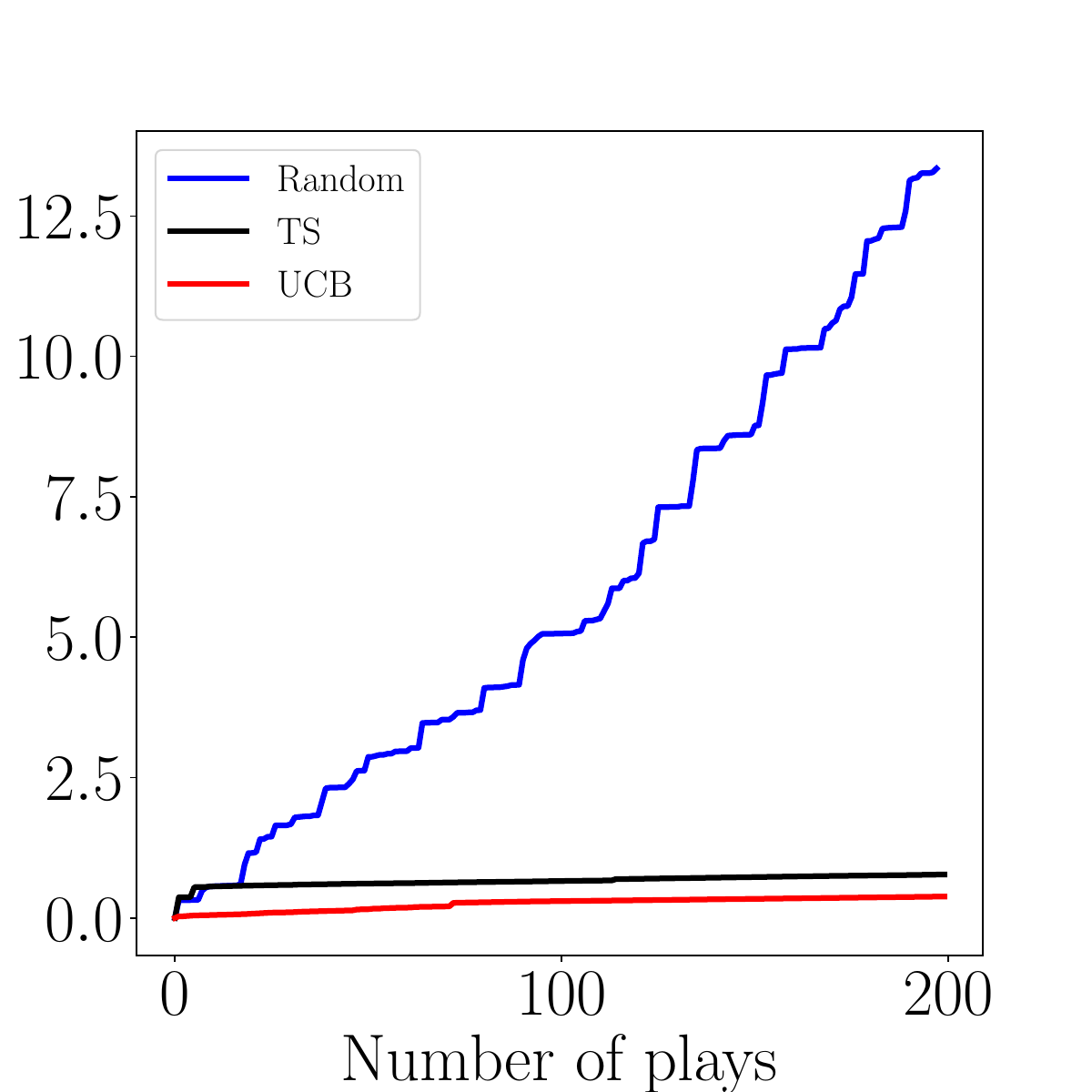}
			\caption{\small Advection equation}
		\end{subfigure}
		&
		\begin{subfigure}[t]{0.24\textwidth}
			\centering
			\includegraphics[width=\textwidth]{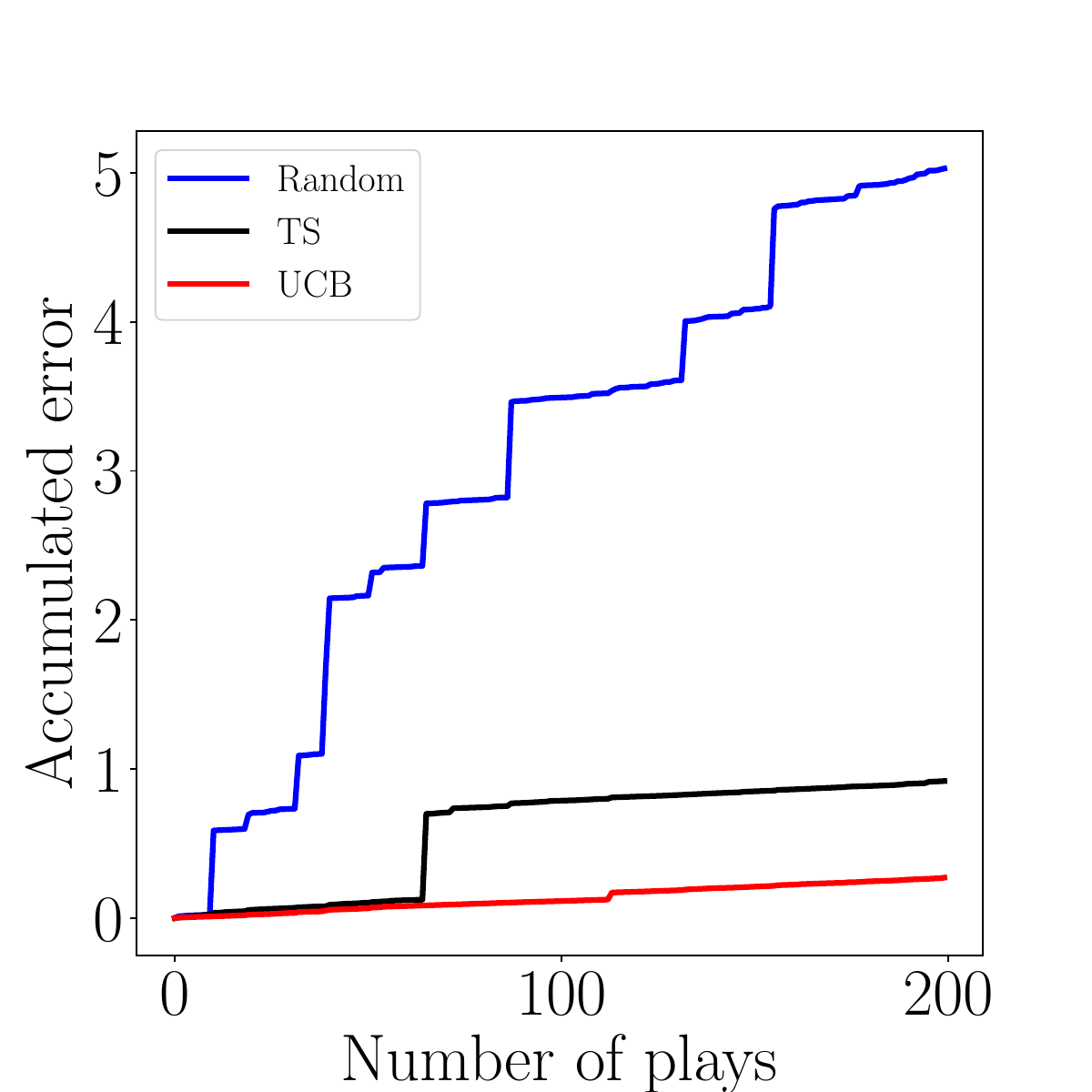}
			\caption{\small Reaction equation}
		\end{subfigure}
		&
		\begin{subfigure}[t]{0.24\textwidth}
			\centering
			\includegraphics[width=\textwidth]{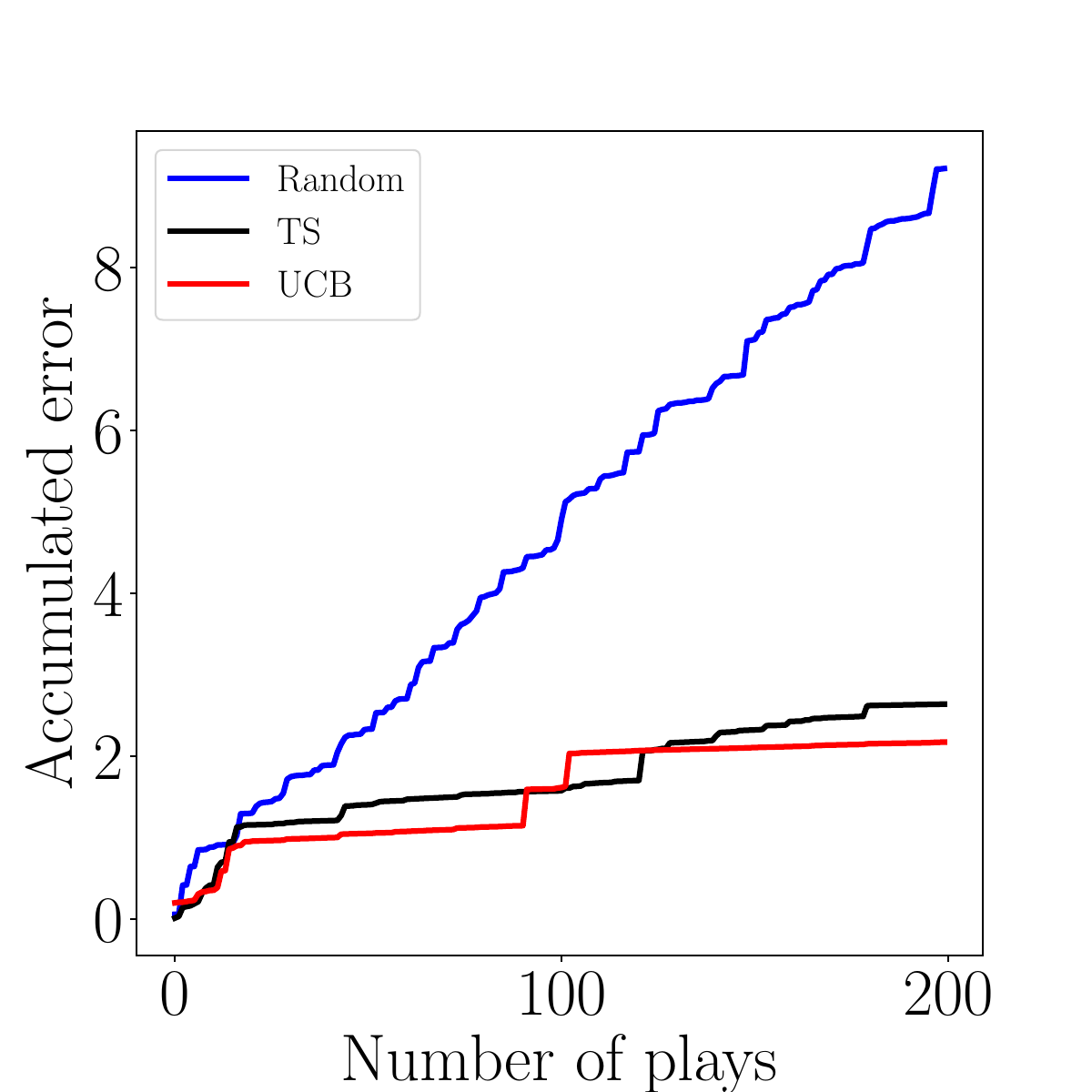}
			\caption{\small Burgers' equation}
		\end{subfigure}
	\end{tabular}
	%\vspace{-0.1in}
	\caption{\small Online performance of \ours-single. } \label{fig:accu-mab-single}
	%\vspace{-0.2in}
\end{figure*}	
\begin{figure*}
	\centering
	\setlength\tabcolsep{0pt}	
	\begin{tabular}[c]{cccc}
		\setcounter{subfigure}{0}
		\begin{subfigure}[t]{0.24\textwidth}
			\centering
			\includegraphics[width=\textwidth]{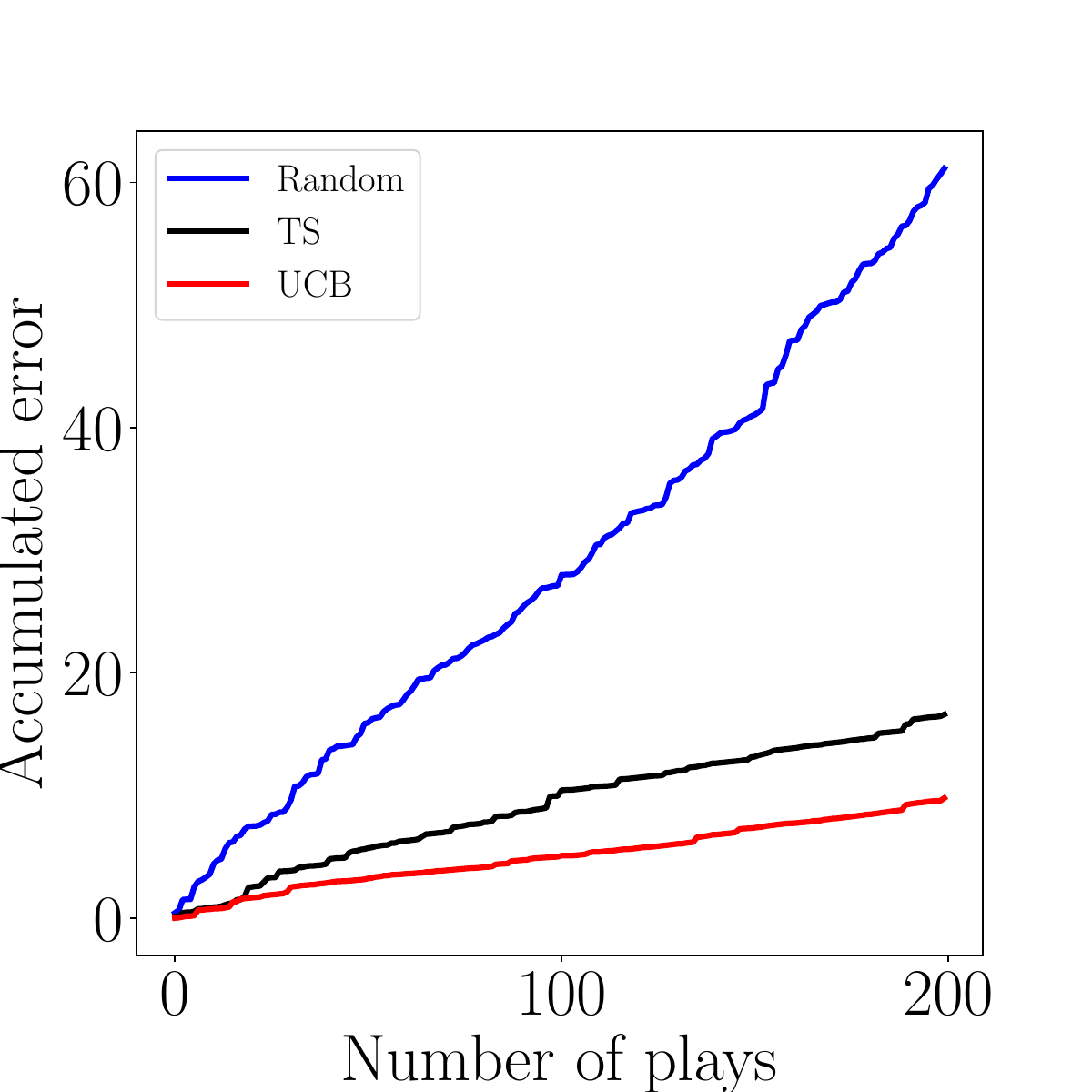}
			\caption{\small Poisson equation}
		\end{subfigure} 
		&
		\begin{subfigure}[t]{0.24\textwidth}
			\centering
			\includegraphics[width=\textwidth]{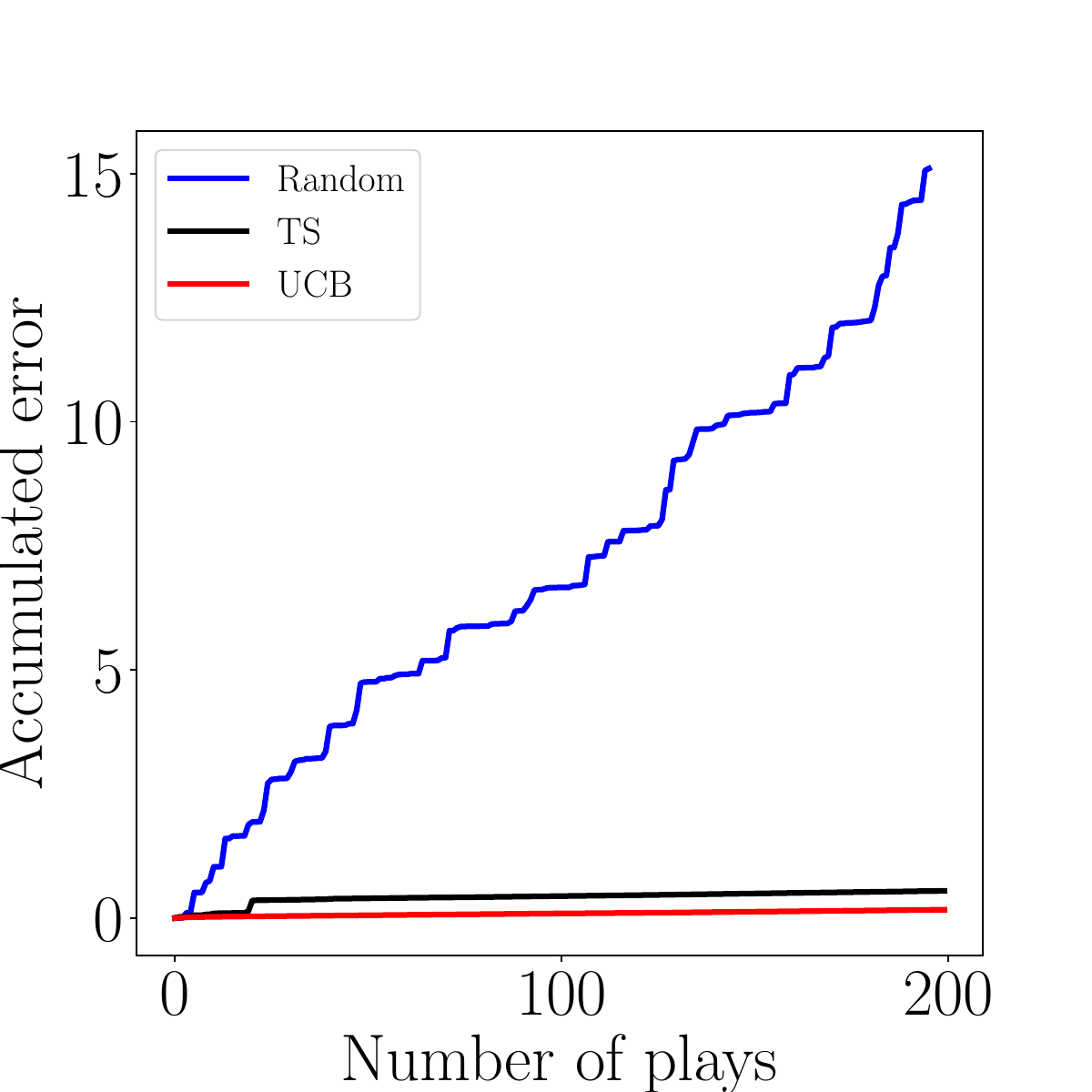}
			\caption{\small Advection equation}
		\end{subfigure}
		&
		\begin{subfigure}[t]{0.24\textwidth}
			\centering
			\includegraphics[width=\textwidth]{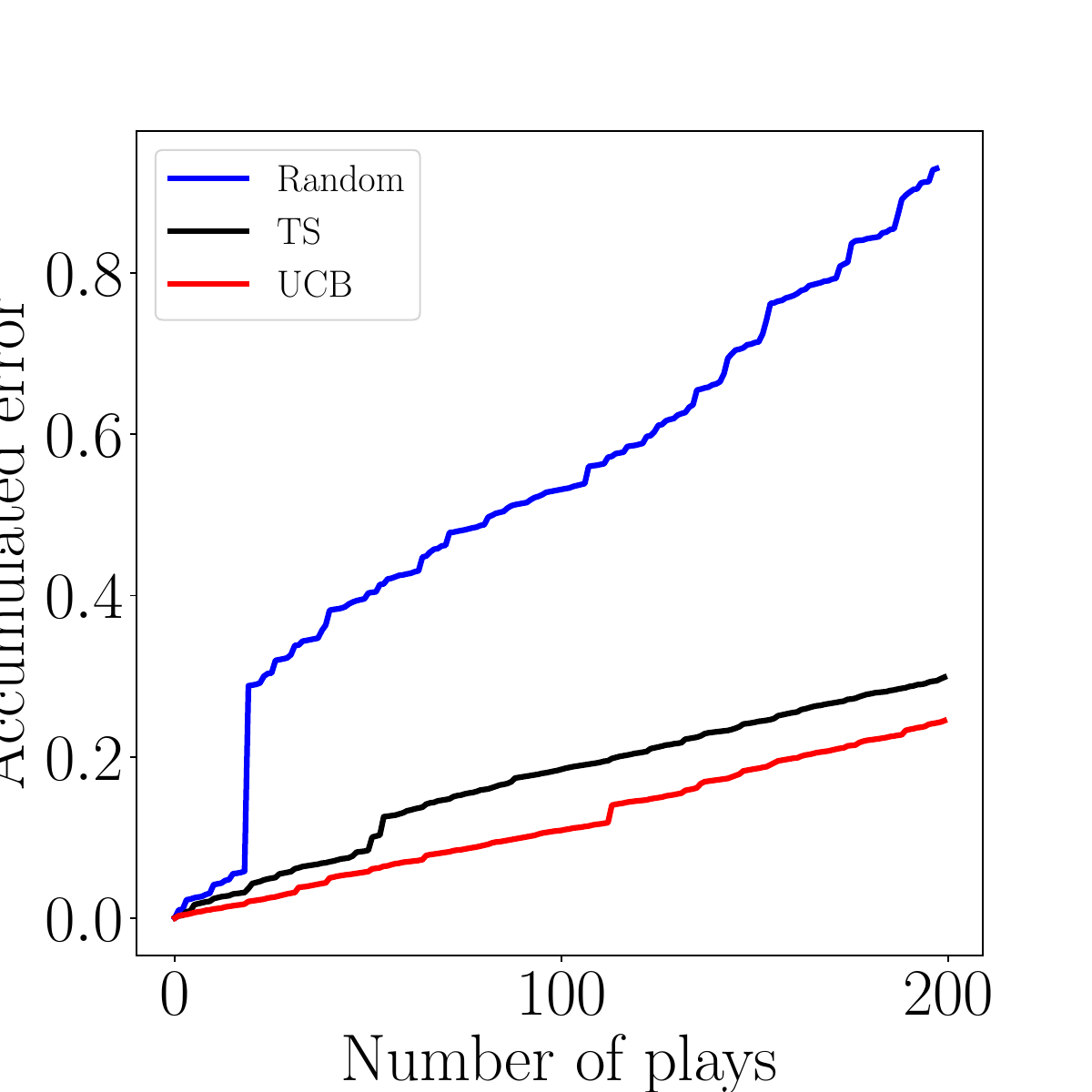}
			\caption{\small Reaction equation}
		\end{subfigure}
		&
		\begin{subfigure}[t]{0.24\textwidth}
			\centering
			\includegraphics[width=\textwidth]{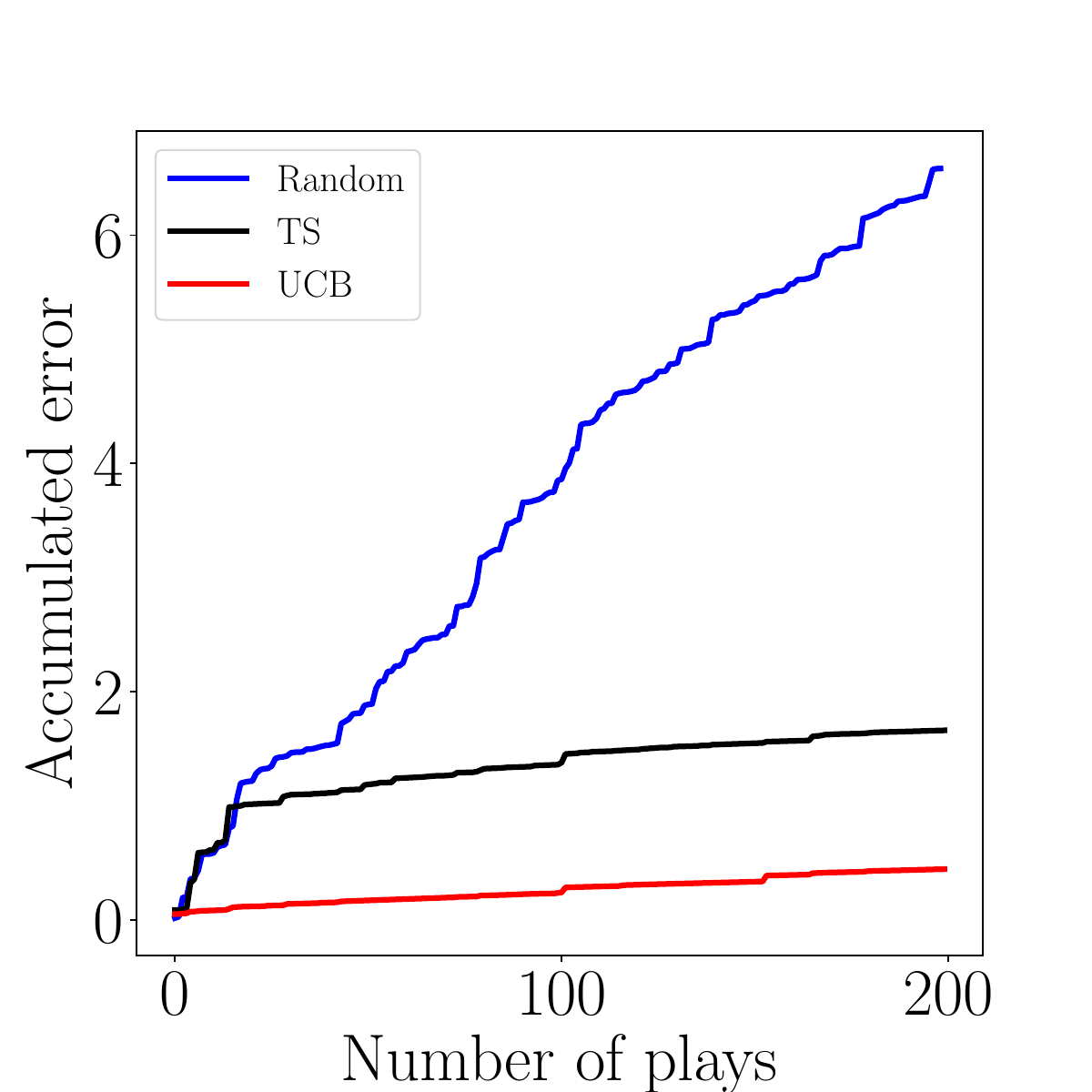}
			\caption{\small Burgers' equation}
		\end{subfigure}
	\end{tabular}
	%	\vspace{-0.05in}
	\caption{\small Online performance of \ours-seq.} \label{fig:accu-mab-seq}
	%	\vspace{-0.1in}
\end{figure*}
\begin{figure}
	\vspace{-0.5in}
	\centering
	\setlength\tabcolsep{0pt}	
	\begin{tabular}[c]{c}
		\begin{subfigure}[b]{0.96\textwidth}
			\centering
			\includegraphics[width=1.0\linewidth]{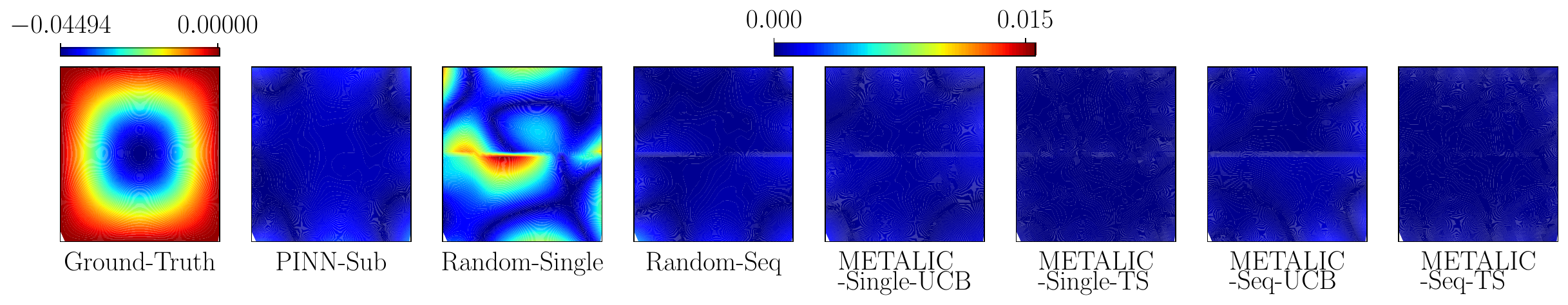}
			\caption{\small Poisson equation ($s=46.1$)}
			\label{fig:poisson}
		\end{subfigure}
		\\
		\begin{subfigure}[b]{0.96\textwidth}
			\centering
			\includegraphics[width=1.0\linewidth]{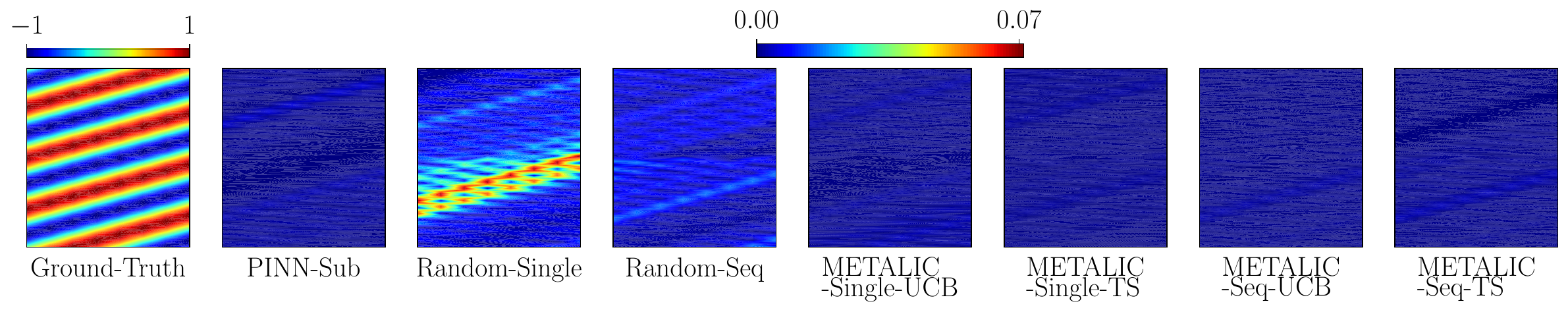}
			\caption{\small Advection equation ($\beta=24.02$)}
			\label{fig:advec}
		\end{subfigure}
		\\
		\begin{subfigure}[b]{0.96\textwidth}
			\centering
			\includegraphics[width=1.0\linewidth]{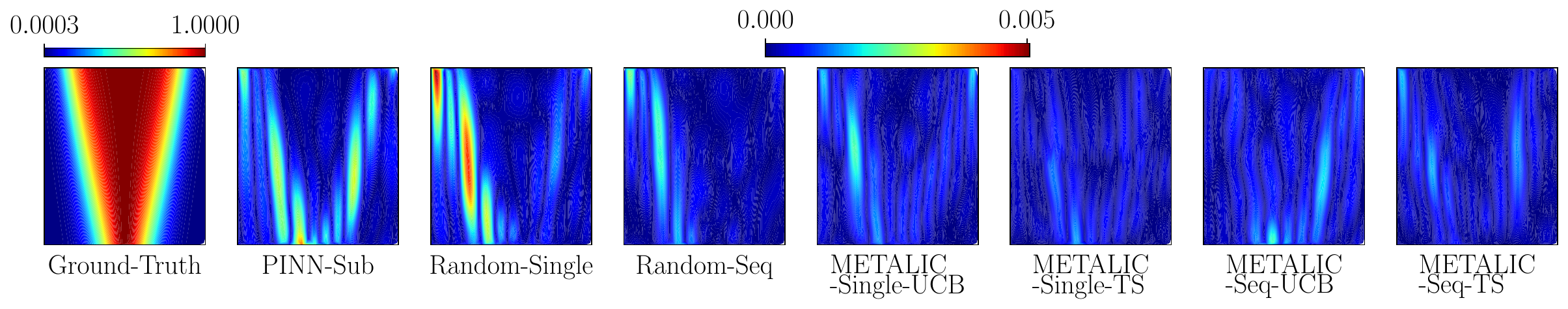}
			\caption{\small Reaction equation ($\rho=4.1$)}
			\label{fig:reaction}
		\end{subfigure}
		\\
		\begin{subfigure}[b]{0.96\textwidth}
			\centering
			\includegraphics[width=1.0\linewidth]{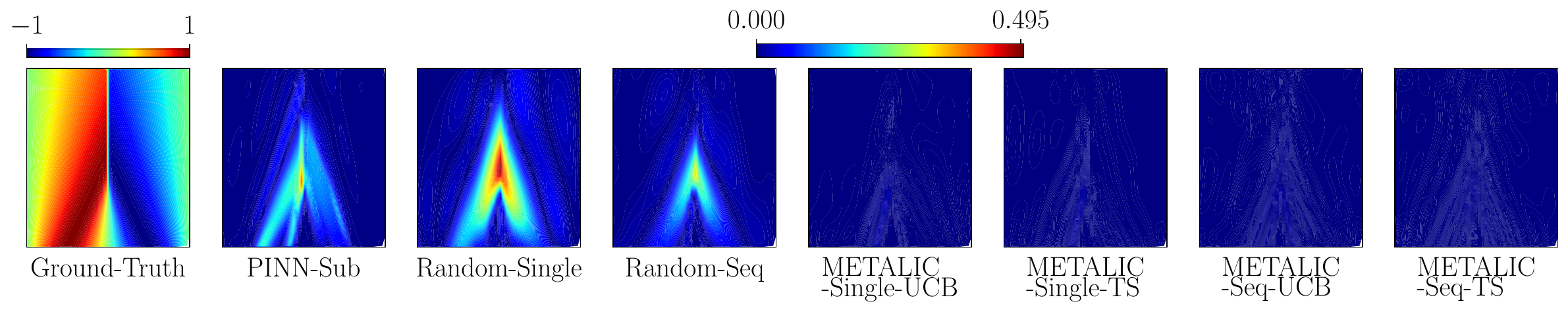}
			\caption{\small Burgers' equation ($\nu=0.0036$)}
			\label{fig:burgers}
		\end{subfigure}
	\end{tabular}
	% 	\vspace{-0.1in}
	\caption{\small Point-wise solution error.}
	\label{fig:point-wise}
	% 	\vspace{-0.5in}
\end{figure}

To evaluate \ours, we used 9 interface conditions, which are listed in the supplementary material. For the \pinn in each subdomain, we used two layers, with 20 neurons per layer and tanh activation function. We randomly sampled 1,000 collocation points and 100 boundary points for each PINN. To inject the interface conditions, we then randomly sampled 101 interface points for the Poisson, advection and reaction equations, and 802 interface points for Burger's equation. 
We set $\lambda_b=20$ and $\lambda_I = 5$, which follows the insight of~\citep{wang2021understanding,wang2022and} to adopt large weights for the boundary and interface terms so as to prevent the training of \pinns from being dominated by the residual term. 
We denote our single MAB by \ours-single, and sequential MABs by \ours-seq. For the latter, we set the discount factor $\gamma = 0.9$ (see \eqref{eq:reward-mab-1}). For better numerical stability, we used the relative $L_2$ error in the log domain to obtain the reward for updating the GP surrogate models. The running of the multi-domain \pinns consists of 10K ADAM epochs (with learning rate $10^{-3}$) and then 50K L-BFGS iterations (the first order optimality and parameter change tolerances set to $10^{-6}$ and $10^{-9}$ respectively). We set $c=1$ to compute the UCB score. We ran 200 plays (trials) for our method.  For static (offline) test, we randomly sampled 100 PDEs (which do not overlap with the PDEs sampled during the online playing). We then used the learned reward model to determine the best interface conditions for each particular PDE (according to the predictive mean), with which we ran the multi-domain \pinns to solve the PDE, and computed the relative $L_2$ error.  
\begin{table*}[t]
	% 	\small
	\centering
	\resizebox{\columnwidth}{!}{%
		\begin{tabular}[c]{ccccc}
			\toprule
			{Method} & {Poisson} & Advection & Reaction & Burger's  \\
			\hline         
			Random-Single & 0.3992 $\pm$ 0.0037 & 0.07042 $\pm$ 0.01575 & 0.00612 $\pm$ 0.000278 & 0.04021 $\pm$ 0.00057\\
			Random-Seq &0.3004 $\pm$ 0.0032 & 0.03922 $\pm$ 0.00919& 0.01284$\pm$ 0.000707 & 0.03486 $\pm$ 0.00066 \\
			PINN-Sub & 0.03078 $\pm$ 0.00177 & 0.00130 $\pm$ 8.108e-5 & 0.00213 $\pm$ 0.00021 & 0.00738 $\pm$ 0.00313 \\
			PINN-Merge-H & 0.02398 $\pm$ 0.00144 & 0.00098 $\pm$ 5.038e-5 & 0.00223 $\pm$ 0.00026 & 0.00951 $\pm$ 0.00390 \\
			PINN-Merge-V & 0.02184 $\pm$ 0.00211 & 0.00079 $\pm$ 3.391e-5 & \textbf{0.00099} $\pm$ \textbf{0.00013} & 0.00276 $\pm$ 0.00041 \\
			\ours-Single-TS &0.02503 $\pm$ 0.0002 & 0.00079 $\pm$ 4.5897e-5 & 0.00204 $\pm$ 1.013e-4 & \textbf{0.00109} $\pm$ \textbf{1.306e-5} \\
			\ours-Single-UCB &0.0245$\pm$ 0.0002 & 0.00078 $\pm$ 3.6771e-5& 0.00102 $\pm$ 8.945e-6 & 0.00161 $\pm$ 2.939e-5 \\
			\ours-Seq-TS &0.01639$\pm$ 9.5384e-5 & 0.00078 $\pm$ 3.6473e-5 & \textbf{0.00099} $\pm$ \textbf{8.4704e-6} & 0.00152 $\pm$ 5.571e-5 \\
			\ours-Seq-UCB & \textbf{0.01406} $\pm$ \textbf{9.1099e-5} & \textbf{0.00070} $\pm$ \textbf{3.2790e-5} & \textbf{0.00099} $\pm$ \textbf{5.999e-6} & 0.00139 $\pm$ 3.948e-5 \\
			\bottomrule
		\end{tabular}
	}
	\caption{\small The average  $L_2$ relative error of single-domain \pinns and multi-domain \pinns for solving 100 test PDEs. The interface conditions of the multi-domain \pinns are provided by \ours and random selection. \{Single, Seq\} indicate using a single set or two sequential sets of interface conditions for the running of the multi-domain \pinns.  \{TS, UCB\} corresponds to our method using TS or UCB score to determine the interface conditions at each play.} \label{tb:test-result}
	% 	\vspace{-0.1in}}}
\end{table*}

First, to examine the online performance of \ours, we looked into the accumulated solution error along with the number of plays. We compared with randomly selecting the arm at each play.  In the case of running \ours-seq, this baseline correspondingly randomly selects the arm twice, one at the stochastic training phase, and the other at the deterministic phase. The results are shown in Fig. \ref{fig:accu-mab-single} and \ref{fig:accu-mab-seq}. As we can see, the accumulated error of \ours with both UCB and TS grows much slower, \ie sublinearly, than the random selection approach (note that the reward of the optimal action is unknown due to the randomness in the running of PINNs, and we cannot compute the regret). This has shown  that our method achieves a much better exploration-exploitation tradeoff in the online interface condition decision and model updating, which is consist with many other MAB applications (see Sec \ref{sect:rel}). The results demonstrate the advantage of our MAB-based approach. First, via effective exploration, \ours can collect valuable training examples (rewards at new actions and context) to improve the learning efficiency and performance of the GP reward surrogate model. Second, the online decision also takes advantage of the predictive abaility of the current reward model, \ie exploitation, to select effective interface conditions, which results in increasingly better solution accuracy of the multi-domain \pinns. The online nature of \ours enables us to keep improving the reward model while utilizing it to solve new equations with promising accuracy. 

Next, we conducted an offline test, namely, without online exploration and model updates any more after 200 plays. We compared with (1) Random-Single, which, for each PDE, randomly selects a set of interface conditions applied to the entire training of the multi-domain PINNs, and (2) Random-Seq, which for each PDE, randomly selects two sets of interface conditions, one for the stochastic training and the other for the deterministic training phase. We also tested single-domain \pinns that do not incorporate interface conditions. Specifically, we compared with (3) \pinn-Sub, which used the same architecture as the PINN in each subdomain, but applied to the entire domain, (4) \pinn-Merge-H, which horizontally pieced all the \pinns in the subdomains, \ie doubling the layer width yet fixing the depth, (5) \pinn-Merge-V, which vertically stacked the \pinns, \ie doubling the depth while fixing the width.  Note that while \pinn-Merge-H and \pinn-Merge-V merge the \pinns in each subdomain, the total number of neurons actually increases (for connecting these \pinns). Hence, the merged \pinn is more expressive. 
Each single-domain \pinn used the union of the boundary points and collocation points from every subdomain. We used the same weight for the boundary term, \ie $\lambda_b= 20$. The running of each single-domain \pinn follows exactly the same setting of the multi-domain \pinns (\ie 10K ADAM epochs and 50K L-BFGS iterations). 

We report the average relative $L_2$ solution error and the standard deviation in Table \ref{tb:test-result}. As we can see, randomly selecting interface conditions, no matter for the whole training procedure or two training phases, result in much worse solution accuracy of multi-domain \pinns. The solution error is one order of magnitude  bigger than \ours in all the settings. It confirms that the success of the multi-domain \pinns is up to appropriate interface conditions. Next, we can observe that while the performance of \ours-Single is similar to \ours-Seq, the best solution accuracy is in most cases obtained by interface conditions selected by \ours-Seq (except in solving the Burger's equation). It demonstrates that our sequential MAB model that can employ different conditions for the two training phases is more flexible and brings additional improvement. We also observe that in most cases using the UCB for online playing can lead to better performance for both \ours-Single and \ours-Seq. This is consistent with the online performance evaluation (see Fig. \ref{fig:accu-mab-single} and \ref{fig:accu-mab-seq}). Third, among the single-domain PINN methods, PINN-Merge-V outperforms PINN-sub in all the equation families and PINN-Merge-H outperforms PINN-sub in the Poisson and advection equations, showing that deeper or wider architectures can help further improve the solution accuracy. However, their performance is still second to the best setting of \ours, which uses simpler  subdomain PINN  architectures and fewer total learnable parameters. These multi-domain PINNs can be further  parallelized to accelerate training. By contrast, if the interface conditions are inferior, such as those selected by Random-Single and Random-Seq, the solution error becomes much worse (orders of magnitude bigger) than single-domain PINNs. Together these results  have shown the importance of the interface conditions for multi-domain \pinns and the advantage of our method. 

For a fine-grained comparison, we visualize the point-wise solution error  of \pinn-Sub, Random-Single, Random-Seq, and our method in solving four random instances of the equations. As shown in Fig. \ref{fig:point-wise}, the point-wise error of both \ours-Single and \ours-Seq  is quite uniform across the domain and close to zero (dark blue). By contrast, the competing methods often exhibit relative large errors in a few local regions, \eg those in the middle (where the shock waves appear) of the domain of the Burger's equation (PINN-Sub, Random-Single, Random-Seq), and the central part of the domain of the Poisson and advection equation (Random-Single). It shows that our method not only can give a superior global accuracy, but locally also better recovers individual solution values.

Finally, we analyzed the selected interface conditions in each equation family by \ours. We found that those conditions are interesting,  physically meaningful, and consistent with the properties of the equations. Due to the space limit, we provide the detailed analysis and discussion in the supplementary material.

\section{Conclusion}
%\vspace{-0.1in}
We have presented \ours, a simple, efficient and powerful meta learning approach to select PDE-specific interface conditions for general multi-domain \pinns. The results at four bench-mark equation families are encouraging. %We have shown that interface configurations for different training phases can outperform a single set of conditions, and 
%In the future, we will extend our method to continuous interface predictions as a function of iterations. 
In the future, we will use the PDE residual as the approximate reward so that our method can be fully unsupervised. We will also extend \ours to meta learn the interface locations along with the conditions, as a function of not only accuracy but training time so as to improve both the solution accuracy and training efficiency of the multi-domain PINNs. 
%so that we can filter out unnecessary conditions that does not contribute to the overall accuracy 

\section*{Acknowledgments}
This work has been supported by MURI AFOSR grant FA9550-20-1-0358, NSF CAREER Award IIS-2046295, and NSF DMS-1848508.

% % Acknowledgements should go at the end, before appendices and references

% \acks{We would like to acknowledge support for this project
% from the National Science Foundation (NSF grant IIS-9988642)
% and the Multidisciplinary Research Program of the Department
% of Defense (MURI N00014-00-1-0637). }

% % Manual newpage inserted to improve layout of sample file - not
% % needed in general before appendices/bibliography.

\newpage

\section*{Appendix}
\section{Interface Conditions} \label{sect:ic-set}
We used a total number of 9 interface conditions throughout all the experiments, which are listed in Table 1. Note that $I_{z}$ and $I_{zz}$ correspond to the first and second-order derivatives w.r.t an input to the PDE solution function. Since all the test PDE problems consist of two spatial or spatiotemporal dimensions, $I_{z}$ and $I_{zz}$ give four interface conditions. There are no mixed derivatives across different input dimensions. In the case that one $I_{z}$ is the same as $I_{c}$, such as in Poisson equation, the de-duplication gives 9 different conditions. In the case that all $I_{z}$'s are different from $I_{c}$, such as in Burger's equation, we used $I_c$ and removed one  $I_z$ ($z=y$ or $z=t$), so that we still maintain 9 interface conditions to be consistent with other experiments. 
%All the test PDE problems consist of two spatiotemporal dimensions, which gives four terms from $I_{\x}$ and $I_{\x \x}$ with no mixed derivatives. In the case that $I_{\x}$ is equal to $I_{c}$, such as Poisson, only one term is used bringing the total to 9 as stated in the manuscript. In the case that they aren't equivalent, such as Burgers, $I_{c}$ is used instead of the first derivative in $y$ or $t$, keeping the total number of terms at 9. It can also be seen that $I_{u_{avg}}$ reduces to $\frac{1}{2}I_{u}$. Given that we run with a constant $\lambda_I = 20$ for all interface terms, this difference can be seen as weakly ($I_{u_{avg}}$) or more strongly ($I_{u}$) enforcing the solution continuity. 

{
\centering
\captionof{table}{Interface Conditions of Multi-domain PINNs } 
\scalebox{1}{
\begin{tabular}{ ll }
\hline
$I_u$ & Solution continuity (\ref{eq:interface-u})\\
$I_{u_{avg}}$ & Average solution continuity (\ref{eq:interface-uavg}) \\
$I_{r}$ & Residual (\ref{eq:interface-r})\\
$I_{rc}$ & Residual continuity (\ref{eq:interface-rc})\\
$I_{gr}$ & Gradient-enhanced residual (\ref{eq:interface-gr})\\
$I_{c}$ & Flux continuity (\ref{eq:interface-c})\\
$I_{z}$ & First-order spatial/temporal derivative continuity (\ref{eq:interface-x})\\
$I_{zz}$ & Second-order spatial/temporal derivative continuity (\ref{eq:interface-xx})\\
\hline
\end{tabular}}\par \label{tb:ic-list}
}
\begin{align}
I_{u}(\btheta_k, \btheta_{k'}) = \frac{1}{J_{k,k'}} \sum_{i=1}^{J_{k,k'}} \left(\uhat_{\btheta_k}(\x_{k,k'}^i) - \uhat_{\btheta_{k'}}(\x_{k,k'}^i)\right)^2 \label{eq:interface-u}
\end{align}

\begin{align}
I_{u_{avg}}(\btheta_k, \btheta_{k'}) = \frac{1}{J_{k,k'}} \sum_{i=1}^{J_{k,k'}} \left(\uhat_{\btheta_k}(\x_{k,k'}^i) - \uhat^{\text{avg}}_{k,k'}(\x_{k,k'}^i)\right)^2 \label{eq:interface-uavg} \\
\text{where} \; \uhat^{\text{avg}}_{k,k'}(\x_{k,k'}^i) = \frac{1}{2}\left(\uhat_{\btheta_k}(\x_{k,k'}^i) + \uhat_{\btheta_{k'}}(\x_{k,k'}^i)\right) \notag
\end{align}

\begin{align}
I_{r}(\btheta_k, \btheta_{k'}) = \frac{1}{J_{k,k'}} \sum_{i=1}^{J_{k,k'}} \left( \left(\Fcal[\uhat_{\btheta_k}](\x_{k,k'}^i) - f(\x_{k,k'}^i)\right)^2 +  \left(\Fcal[\uhat_{\btheta_{k'}}](\x_{k,k'}^i) - f(\x_{k,k'}^i)\right)^2 \right) \label{eq:interface-r}
\end{align}

\begin{align}
I_{rc}(\btheta_k, \btheta_{k'}) = \frac{1}{J_{k,k'}} \sum_{i=1}^{J_{k,k'}} \left( \left(\Fcal[\uhat_{\btheta_k}](\x_{k,k'}^i) - f(\x_{k,k'}^i)\right) - \left(\Fcal[\uhat_{\btheta_{k'}}](\x_{k,k'}^i) - f(\x_{k,k'}^i)\right) \right)^2 \label{eq:interface-rc}
\end{align}

\begin{align}
I_{gr}(\btheta_k, \btheta_{k'}) = \frac{1}{J_{k,k'}} \sum_{i=1}^{J_{k,k'}} \sum_{j=1}^{2} \left( \left|\frac{\partial}{\partial \x_{k,k'}^i[j]} \left(\Fcal[\uhat_{\btheta_k}](\x_{k,k'}^i) - f(\x_{k,k'}^i)\right)\right|^2 + \left|\frac{\partial}{\partial \x_{k,k'}^i[j]} \left(\Fcal[\uhat_{\btheta_{k'}}](\x_{k,k'}^i) - f(\x_{k,k'}^i)\right)\right|^2 \right) \label{eq:interface-gr} %\\ 
%\text{where} \; \x \in \Omega \subset \mathbb{R}^n \notag
\end{align}

\begin{align}
I_{c}(\btheta_k, \btheta_{k'}) = \frac{1}{J_{k,k'}} \sum_{i=1}^{J_{k,k'}} \left( \phi \left( \uhat_{\btheta_k}(\x_{k,k'}^i) \right) \cdot \n - \phi \left( \uhat_{\btheta_{k'}}(\x_{k,k'}^i) \right) \cdot \n \right)^2 \label{eq:interface-c} \\
\text{where} \; \phi(\uhat_{\btheta}) \cdot \n \; \text{are fluxes normal at the interface} \notag
\end{align}

\begin{align}
I_{z}(\btheta_k, \btheta_{k'}) = \frac{1}{J_{k,k'}} \sum_{i=1}^{J_{k,k'}} \left( \frac{\partial}{\partial z^i}\uhat_{\btheta_k}(\x_{k,k'}^i) - \frac{\partial}{\partial z^i} \uhat_{\btheta_{k'}}(\x_{k,k'}^i)\right)^2 \label{eq:interface-x} \\
\text{where}\; z^i = \x^i_{k, k'}[1]\; \text{or}\; z^i = \x^i_{k,k'}[2]. \notag 
\end{align}

\begin{align}
I_{zz}(\btheta_k, \btheta_{k'}) = \frac{1}{J_{k,k'}} \sum_{i=1}^{J_{k,k'}} \left( \frac{\partial^2}{\partial {z^i}^2} \uhat_{\btheta_k}(\x_{k,k'}^i) - \frac{\partial^2}{\partial {z^i}^2} \uhat_{\btheta_{k'}}(\x_{k,k'}^i)\right)^2 \label{eq:interface-xx} \\
\text{where}\;  z^i = \x^i_{k, k'}[1]\; \text{or}\; z^i = \x^i_{k,k'}[2]. \notag 
\end{align}

\begin{figure}
	\centering
	\setlength\tabcolsep{0pt}
	\begin{tabular}[c]{cc}
		\setcounter{subfigure}{0}
		\begin{subfigure}[t]{0.25\textwidth}
			\centering
			\includegraphics[width=\textwidth]{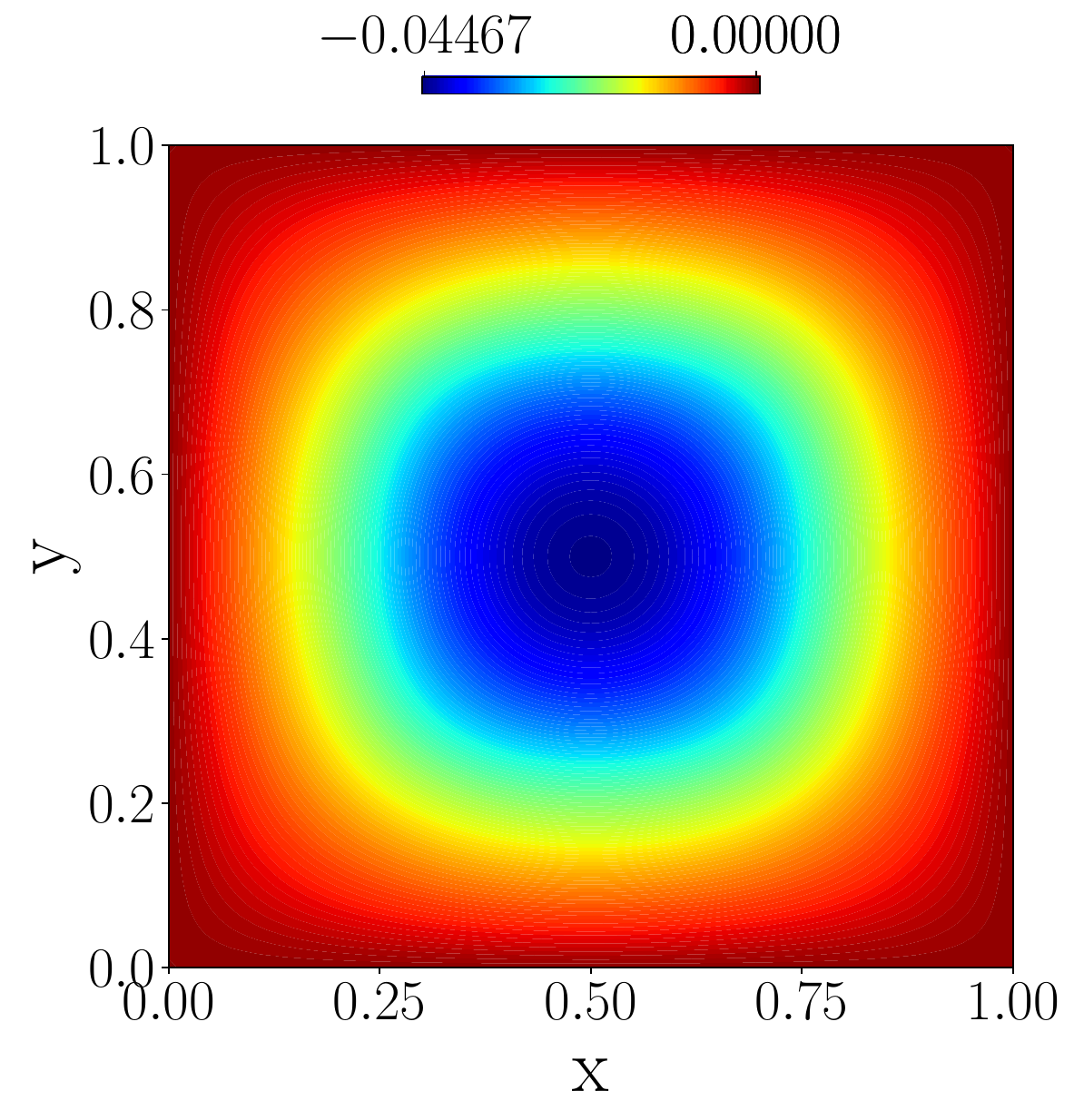}
			\caption{\small Solution at $s=20$}
		\end{subfigure} &
		\begin{subfigure}[t]{0.25\textwidth}
			\centering
			\includegraphics[width=\textwidth]{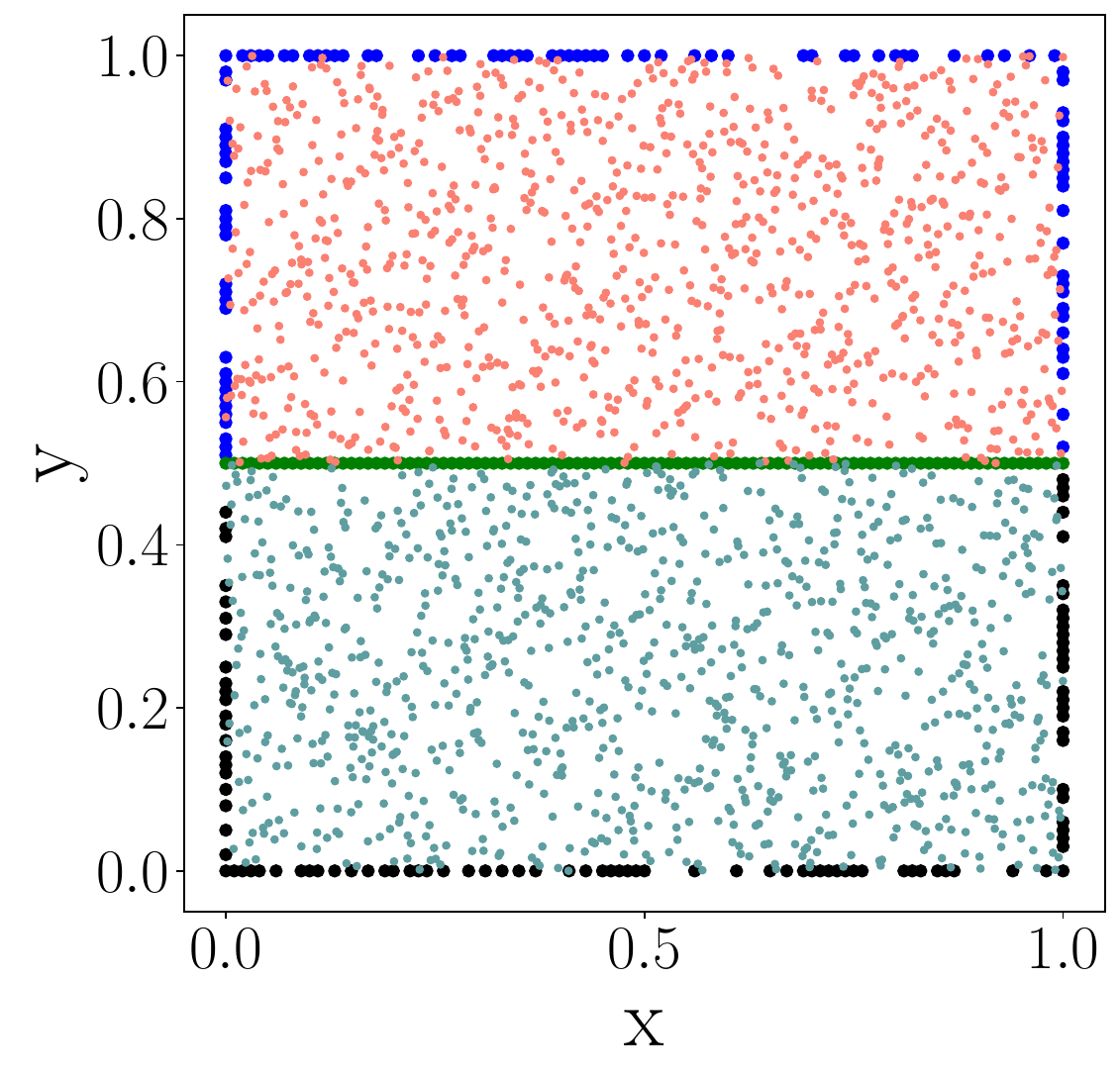}
			\caption{\small Subdomains}
		\end{subfigure}
	\end{tabular}
\vspace{-0.1in}
	\caption{\small The Poisson equation. The interface is the green line at $y=0.5$. Blue and black dots show the sampled boundary points in each subdomain, and the internal dots (red and cyan) are the sampled collocation points inside each subdomain.} \label{fig:poisson-vis}
	\vspace{-0.1in}
\end{figure}
\begin{figure}
	\centering
	\setlength\tabcolsep{0pt}
	\begin{tabular}[c]{cc}
		\setcounter{subfigure}{0}
		\begin{subfigure}[t]{0.25\textwidth}
			\centering
			\includegraphics[width=\textwidth]{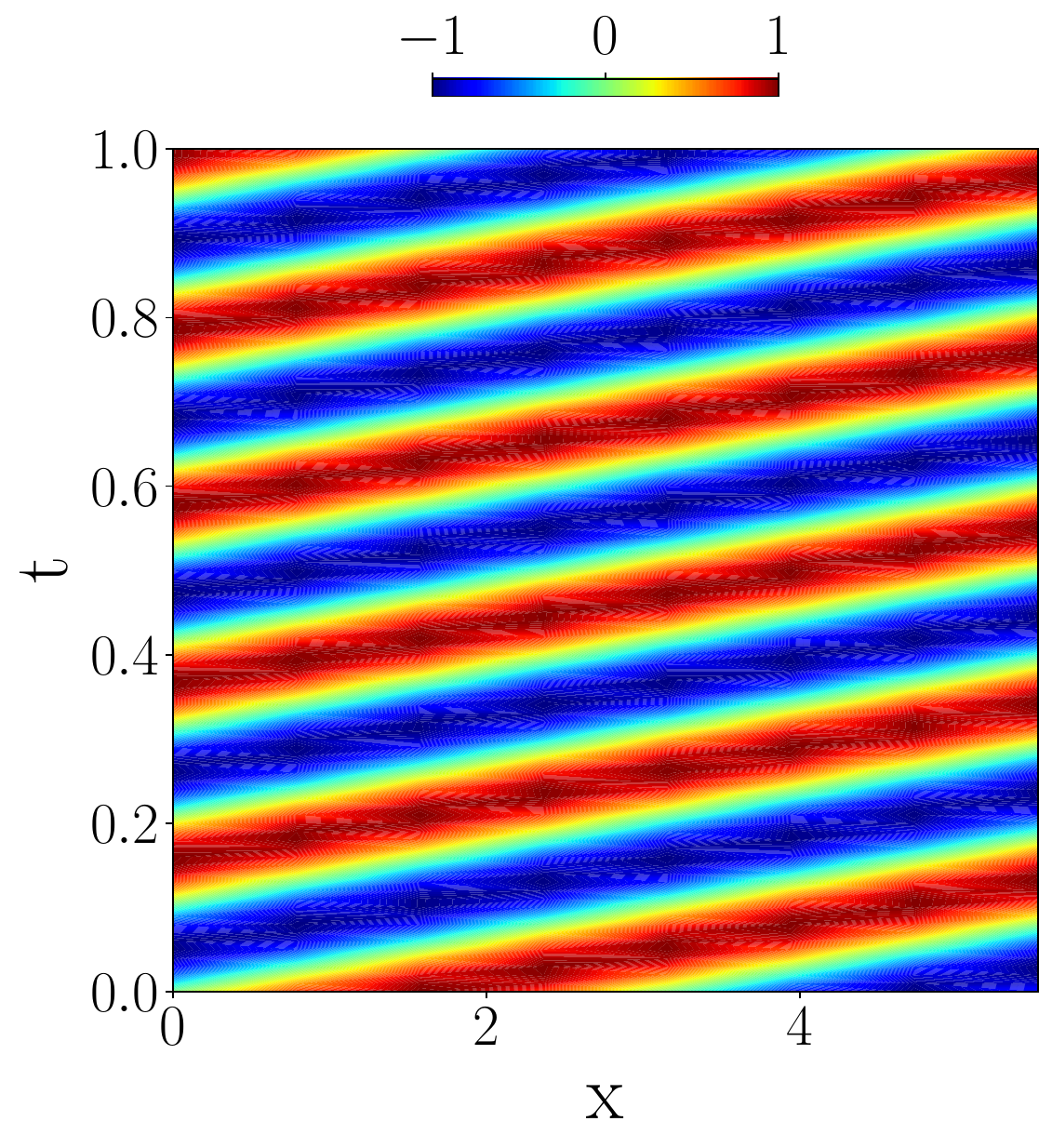}
		%	\vspace{-0.1in}
			\caption{\small Solution at $\beta=30$ }
		%	\vspace{-0.1in}
		\end{subfigure} &
		\begin{subfigure}[t]{0.25\textwidth}
			\centering
			\includegraphics[width=\textwidth]{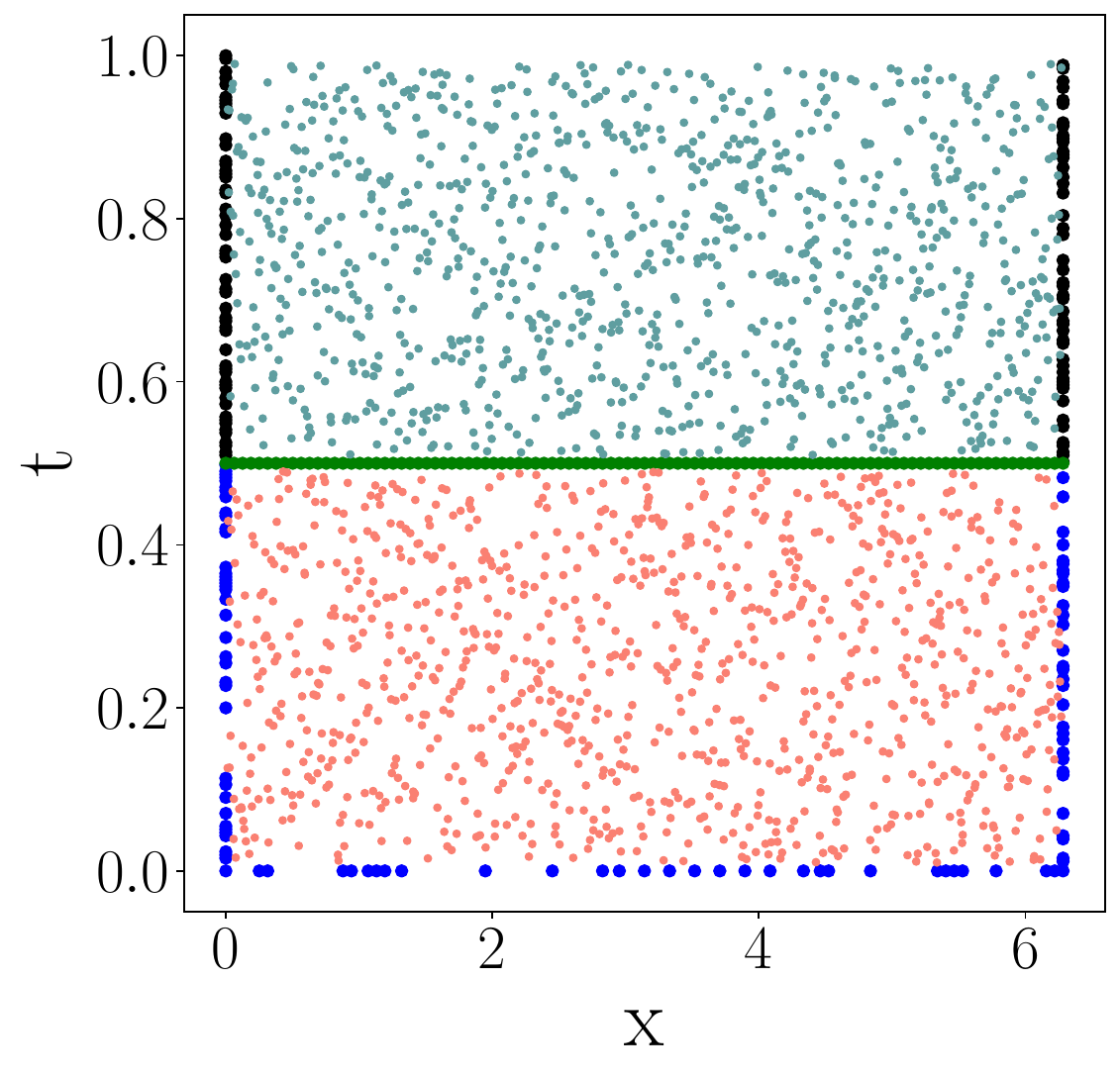}
			\caption{\small Subdomains}
		\end{subfigure}
	\end{tabular}
%\vspace{-0.1in}
	\caption{\small Advection equation. The interface is the green line at $t=0.5$.} \label{fig:advec-vis}
	%\vspace{-0.2in}
\end{figure}
\begin{figure}[H]
	\centering
	\setlength\tabcolsep{0pt}
	\begin{tabular}[c]{cc}
		\setcounter{subfigure}{0}
		\begin{subfigure}[t]{0.25\textwidth}
			\centering
			\includegraphics[width=\textwidth]{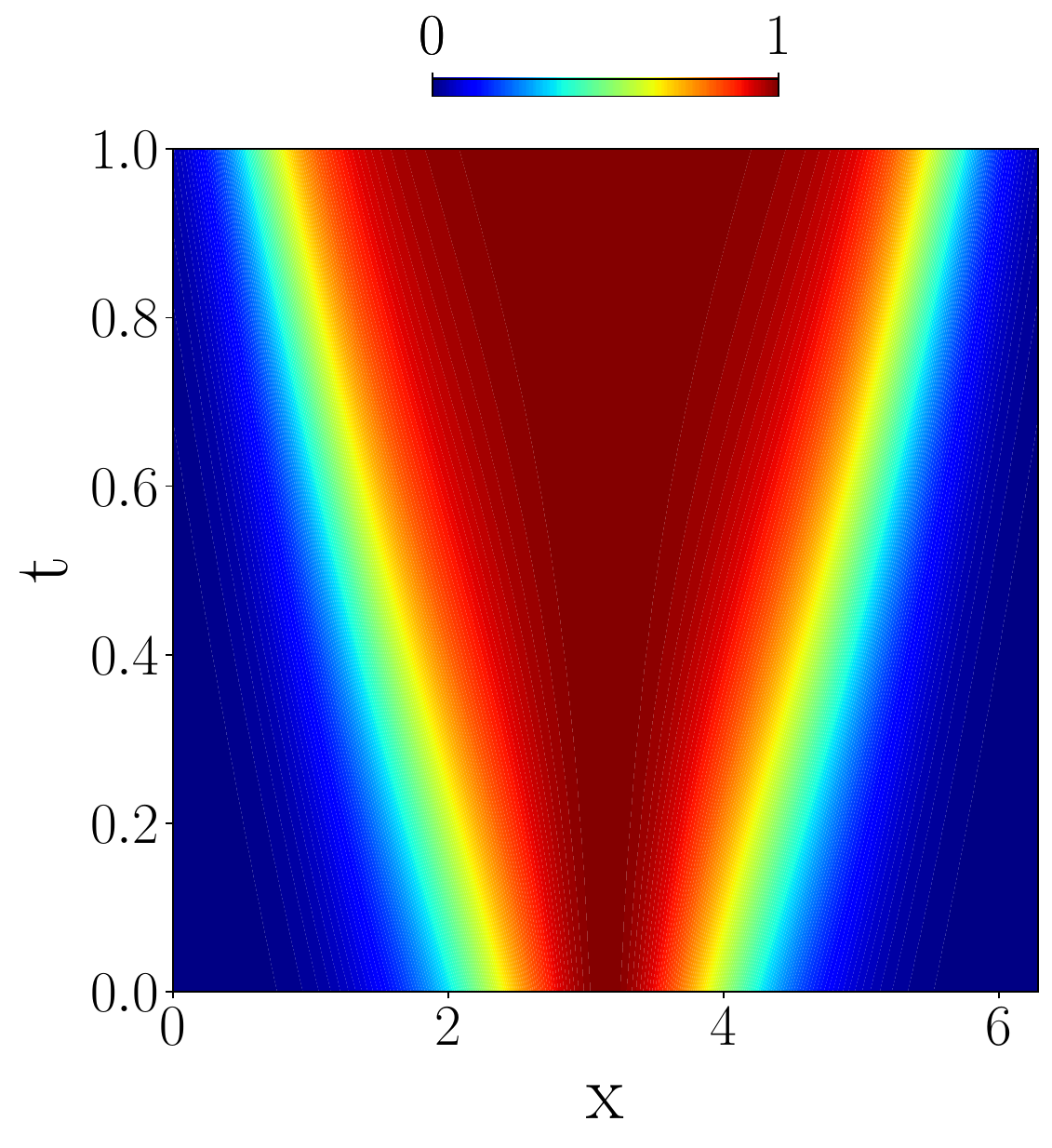}
			\caption{\small Solution at $\rho=5.0$}
		\end{subfigure} &
		\begin{subfigure}[t]{0.25\textwidth}
			\centering
			\includegraphics[width=\textwidth]{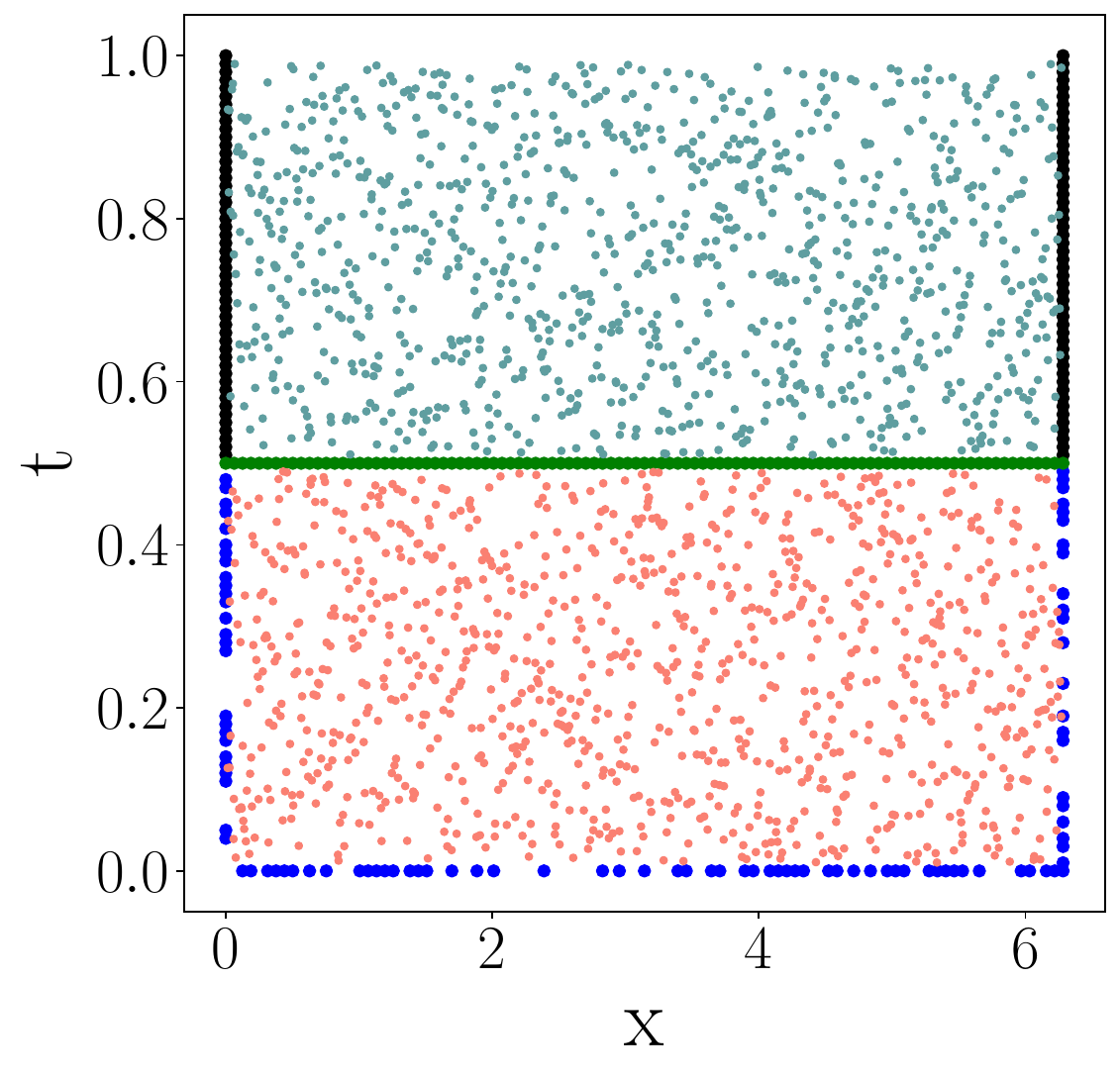}
			\caption{\small Subdomains}
		\end{subfigure}
	\end{tabular}
	\caption{\small Reaction equation. The interface is at $t=0.5$.} \label{fig:reaction-vis}
\end{figure}
\begin{figure}[H]
	\centering
	\setlength\tabcolsep{0pt}
	\begin{tabular}[c]{cc}
		\setcounter{subfigure}{0}
		\begin{subfigure}[t]{0.25\textwidth}
			\centering
			\includegraphics[width=\textwidth]{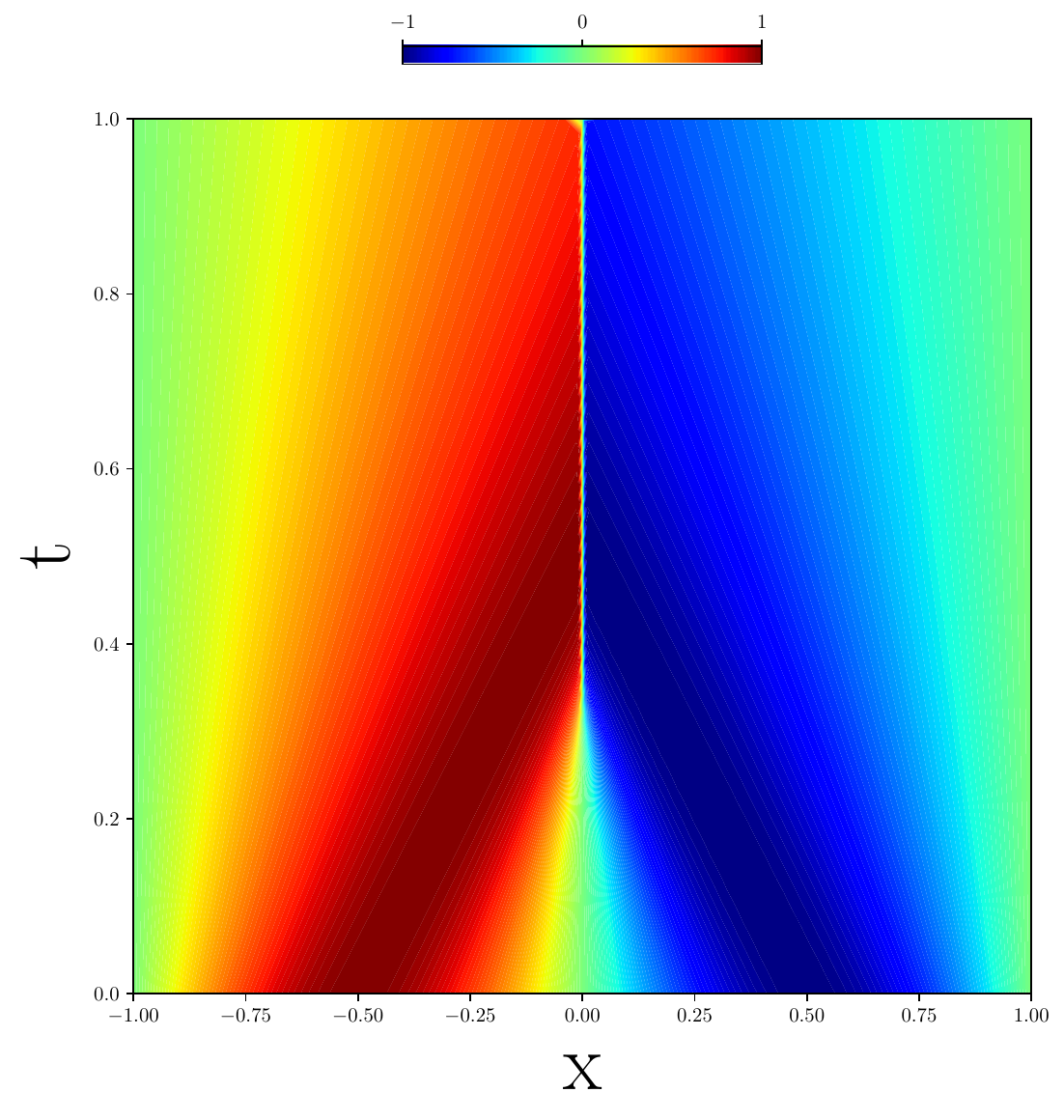}
			\caption{\small Solution at $\nu=0.001$ }
		\end{subfigure} &
		\begin{subfigure}[t]{0.25\textwidth}
			\centering
			\includegraphics[width=\textwidth]{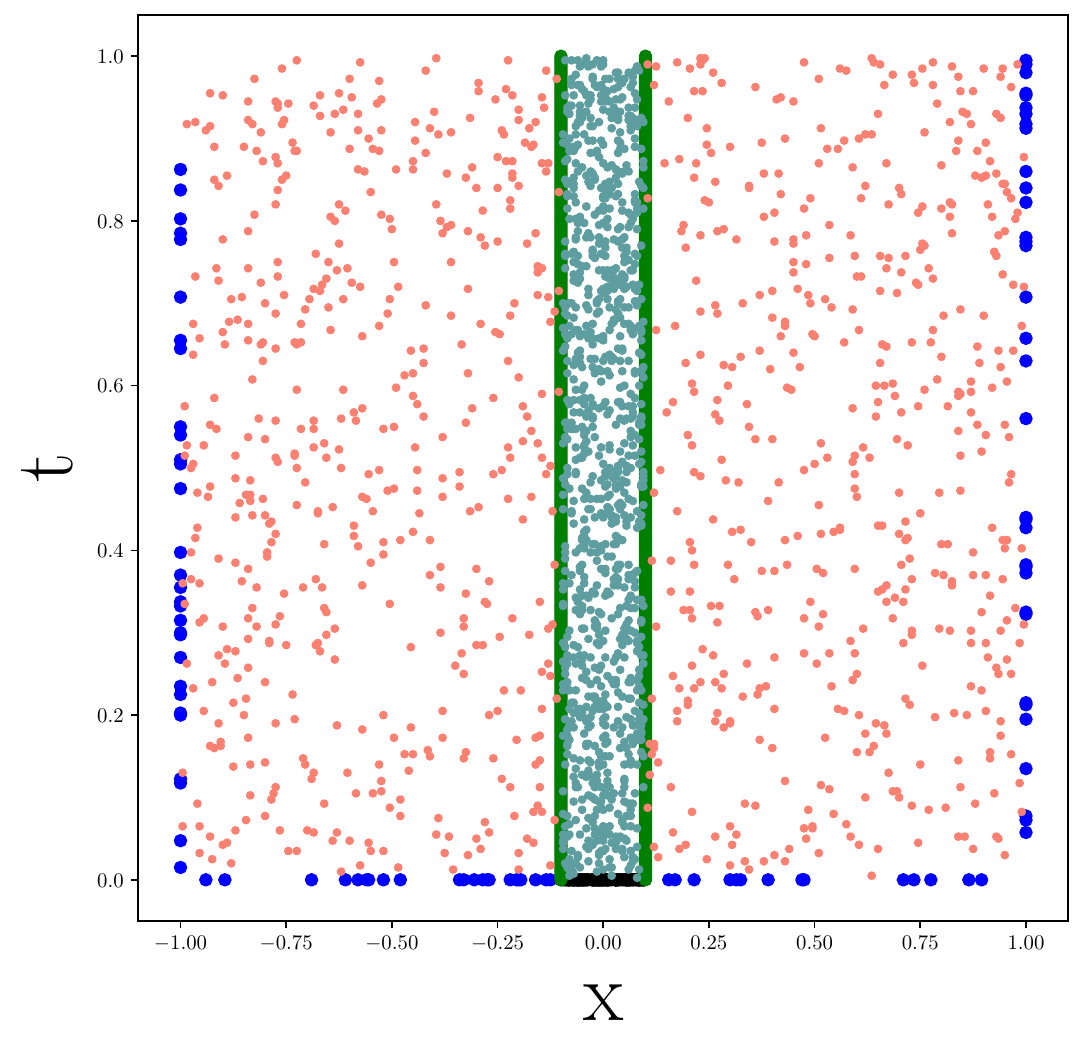}
			\caption{\small Subdomains}
		\end{subfigure}
	\end{tabular}
	\caption{\small Burger's equation. The interfaces are at $x=-0.1$ and $x=0.1$. The middle portion (filled with cyan dots) is the first subdomain, and the remaining parts constitute the  second subdomain.} \label{fig:burgers-vis}
	%\vspace{-0.2in}
\end{figure}

\section{Regret Bound Proof}
Recall that $\mathcal X\subseteq\mathbb{R}^d$ is a compact set denoting the parameter (context) space associated with the parametrized PDE, and $\mathcal S$ is the state space consisting of a finite number of interface conditions, i.e. $|\mathcal S|=s\in\mathbb{N}$. 
The action space $\mathcal P$ is defined as
\begin{align}
	\mathcal P = 2^{\mathcal S} = \{(q_1, \cdots, q_{s})\in \{0, 1\}^{s}\}\subset\mathbb{R}^{s}.  \notag 
\end{align} 
In our paper, the reward at time $t$ is modeled as 
\begin{align}
	&r_t = r(\bbeta_t, a_t) = f(\bbeta_t, a_t) + \eta_t& (\bbeta_t, a_t)\in\mathcal X\times\mathcal P, \label{101}
\end{align}
where $\bbeta_t$ is the context revealed at time $t$, $a_t$ is the selected action, $f$ is a function on $\mathcal X\times\mathcal P$ sampled from an appropriate prior, and $\eta_t$ is a white noise process used to model the extraneous randomness (e.g. neural network implementation, error rounding, etc.)   

In our case, the true reward is the negative error metric computed for the learned PDE solution, which is too complicated for analysis. Alternatively, we use the above model \eqref{101} as a substitute for approximation. As a result, the reward model is misspecified. Nevertheless, we assume that \eqref{101} is reflective of the true reward and do not consider the model misspecification effects in the subsequent analysis for ease of demonstration; ideas from \citep{bogunovic2021misspecified} can be used to obtain refined analysis for misspecified models but we do not pursue them here. 

We now state the technical assumptions on the model parametrization as used in the METALIC algorithm:
\begin{itemize}
	\item $f(\bbeta, a)$ is sampled from a $\gp(0, \kappa)$ prior, where $\kappa$ is a kernel function on $\mathcal X\times\mathcal P\subset\mathbb{R}^{d+s}$:
	\begin{align*}
		\kappa((\bbeta, a), (\bbeta', a')) = \kappa_1(\bbeta, \bbeta')\kappa_2(a, a'),
	\end{align*}
	where $\kappa_1$ and $\kappa_2$ are Gaussian kernels:
	\begin{align*}
		&\kappa_1(\bbeta, \bbeta') = \exp(-\tau_1\|\bbeta-\bbeta'\|_2^2) & \kappa_2(a, a') = \exp\left(-\frac{\tau_2}{s}\|a-a'\|_1\right).
	\end{align*}
	(Note: The key assumption we will be using the is the tensor product structure as well as the form of $\kappa_1$; since $\kappa_2$ is discrete, it does not encode much of geometry and changing to other alternatives should not affect the subsequent analysis. )
	\item $\eta_t$ are i.i.d. Gaussian with variance $\sigma_0^2$:
	\begin{align*}
		\eta_t\stackrel{\text{i.i.d.}}{\sim} N(0, \sigma_0^2).
	\end{align*}
\end{itemize}

Under the above assumptions, for $T\in\mathbb{N}$ and historical observations $\x_t = (\bbeta_t, a_t)\in\mathcal X\times\mathcal P$, $1\leq t\leq T$, $y_T = (r_1, \cdots, r_T)^\top$, 
the posterior distribution of $f$ at $\x = (\bbeta, a)\in\mathbb{R}^{d+s}$ is a normal random variable with mean and variance given below:
\begin{align*}
	\mu_T(\x) &= k^\top_T(\x)(\sigma_0^2 I_T + K_T)^{-1}y_T\\
	\sigma_T^2(\x) &= \kappa(\x, \x) - k^\top_T(\x)(\sigma_0^2 I_T + K_T)^{-1}k_T(\x),
\end{align*}
where 
\begin{align*}
	&K_T = (\kappa(\x_i, \x_j))_{1\leq i, j\leq T}\in\mathbb{R}^{T\times T}&k_T(\x) = (\kappa(\x, \x_1), \cdots, \kappa(\x, \x_T))^\top\in\mathbb{R}^T. 
\end{align*}
The following quantity, which measures the maximum uncertainty reduction of $f_T = (f(\x_1), \cdots, f(\x_T))^\top$ when observing $y_T$, will appear in the regret analysis:
\begin{align*}
	\gamma_T &:= \max_{\{\x_t\}\subset\mathcal X\times\mathcal P}H(f_T) - H(f_T|y_T)\\
	& = \max_{\{\x_t\}\subset\mathcal X\times\mathcal P}H(y_T) - H(y_T|f_T)\ \ \ \ \ \ \ \ \text{$H(y_T|f_T)$ is independent of $\{\x_t\}$}\\
	& = \max_{\{\x_t\}\subset\mathcal X\times\mathcal P}\frac{1}{2}\log | I_T + \sigma_0^{-2}K_T|, 
\end{align*}
where $H$ is the Shannon entropy.

We are now ready to state the main result:

\begin{Th}
	For $\delta>0$, take $c_t$ in the UCB algorithm as
	\begin{align*}
		&c_t = 2\log\left(\frac{2^s\pi^2t^2}{6\delta}\right)& t\in\mathbb{N}.
	\end{align*}
	Conditioning on every context sequence $\{\bbeta_t\}$, let $\{a_t\}$ be the action selected by the UCB algorithm under the above choice of $\{c_t\}$. 
	Then, with probability at least $1-\delta$, the regret $R_T$ satisfies
	\begin{align}
		&R_T\lesssim  \sqrt{\frac{2^{s}T(\log T)^{d+1}\log\left(\frac{2^sT^2}{\delta}\right)}{\log (1+\sigma_0^{-2})}}& T = 1, 2, \cdots, \label{ucb1}
	\end{align}
	where the implicit constant is absolute (does not depend on $\{\bbeta_t\}$ but depends on the domain $\mathcal X$). 
	In particular, 
	\begin{align}
		\EE[R_T]\lesssim  \sqrt{\frac{2^{s}T(\log T)^{d+1}\log\left(2^sT^2\right)}{\log(1+\sigma_0^{-2})}},\label{ucb2}
	\end{align}
	Moreover, \eqref{ucb2} holds also for the Thompson sampling algorithm. 
\end{Th}

\begin{proof}
	We first prove the statement concerning the UCB algorithm. 
	The proof is similar to \citep{krause2011contextual} overall. Owing to a few subtle differences, we provide a sketch of the proof. In our setting, contexts are revealed in a random fashion that is independent of the reward noises. For convenience, we condition on the context sequence $\{\bbeta_t\}_{t\in\mathbb{N}}\subset \mathcal X$ throughout the analysis, i.e., we treat $\{\bbeta_t\}_{t\in\mathbb{N}}\subset \mathcal X$ as deterministic and arbitrary sequence.  
	
	Firstly, a standard application of concentration inequalities and a union bound \citep[supplement, Lemma 5.2]{krause2011contextual} yields that, with probability at least $1-\delta$, 
	\begin{align}
		&|f(\bbeta_t, a)-\mu_t(\bbeta_t, a)|\leq c_t^{1/2}\sigma_{t-1}(\bbeta_t, a)& t\in\mathbb{N}, \ a\in\mathcal P, \label{ou1}
	\end{align}
	which immediately implies an upper bound for the regret:
	\begin{align*}
		R_T &= \sum_{t=1}^T(f(\bbeta_t, a_t^*) - f(\bbeta_t, a_t))\ \ \ \ \ \ \ \ \ a_t^* = \arg\max_{a\in\mathcal P}f(\bbeta_t, a)\\
		& \leq \sum_{t=1}^T\underbrace{(\mu_{t-1}(\bbeta_t, a_t^*) + c_t^{1/2}\sigma_{t-1}(\bbeta_t, a_t^*)) - (\mu_{t-1}(\bbeta_t, a_t) + c_t^{1/2}\sigma_{t-1}(\bbeta_t, a_t))}_{\leq 0} + 2c_t^{1/2}\sigma_{t-1}(\bbeta_t, a_t)\\
		& \leq 2\sum_{t=1}^T c_t^{1/2}\sigma_{t-1}(\bbeta_t, a_t)\\
		&\leq 2\sqrt{T}\left[c_T\sum_{t=1}^T \sigma^2_{t-1}(\bbeta_t, a_t)\right]^{1/2}\ \ \ \ \ \ \ \text{(Cauchy--Schwarz; $c_t$ is increasing in $t$)}\\
		&\stackrel{(\star)}{\leq} \sqrt{\frac{8Tc_T\gamma_T}{\log(1+\sigma_0^{-2})}}\\
		&\lesssim K(\sigma_0)\sqrt{Tc_T\gamma_T}\ \ \ \ \ \ \ \ \ \ \ \ \ K(\sigma_0) = \sqrt{\frac{1}{\log (1+\sigma_0^{-2})}},
	\end{align*}
	where the ($\star$) follows from \citep[Theorem 5]{krause2011contextual} and an intuitive way to understand it is that the total information gain (i.e. the predictive variance term; see \citep[Lemma 5.3]{srinivas2009gaussian}) is bounded by the maximum information gain under the optimal design. 
	
	It remains to bound $\gamma_T$ for the kernel $\kappa$. Since $\kappa$ is a tensor product of $\kappa_1$ and $\kappa_2$, with $\kappa_2$ being a kernel on a discrete set with cardinality $2^s$ (i.e. has rank $2^s$),  according to \citep[Theorem 2]{krause2011contextual}, 
	\begin{align*}
		\gamma_T\leq 2^s(\gamma_T|_{\kappa_1} + \log T),
	\end{align*}
	where $\gamma_T|_{\kappa_1}$ is the maximum information gain defined for the GP with kernel function $\kappa_1$. 
	Note $\kappa_1$ is the Gaussian kernel. \citep[Theorem 5]{srinivas2009gaussian} tells us that $\gamma_T|_{\kappa_1}=\mathcal O((\log T)^{d+1})$, where the implicit constant depends on the domain $\mathcal X$. Hence, $\gamma_T\lesssim 2^s(\log T)^{d+1}$. Plugging this into the above bound for $R_T$ yields the high-probability bound \eqref{ucb1}. 
	For \eqref{ucb2}, note that \eqref{ucb1} and $\sqrt{x+y}\leq\sqrt{x}+\sqrt{y}, x, y\geq 0$ together imply that there exists an absolute constant $C>0$ so that with probability at least $1-\delta$, 
	\begin{align*}
		\widetilde{R}_T:=\frac{\left|R_T - CK(\sigma_0)\sqrt{2^{s}T(\log T)^{d+1}\log\left(2^sT^2\right)}\right|}{CK(\sigma_0)\sqrt{2^sT(\log T)^{d+1}}}\leq\sqrt{\log\left(\frac{1}{\delta}\right)},
	\end{align*} 
	i.e., $\P(\widetilde{R}_T\geq x|\{\bbeta_t\})\leq e^{-x^2}$. Integrating the tail probability yields 
	\begin{align*}
		\EE[\widetilde{R}_T|\{\bbeta_t\}] = \int_0^\infty\P(\widetilde{R}_T\geq x|\{\bbeta_t\})dx\leq \int_0^\infty e^{-x^2}dx = \sqrt{\pi}.
	\end{align*}
	Taking expectation over $\{\bbeta_t\}$ yields $\EE[\widetilde{R}_T]\leq\sqrt{\pi}$. 
	\eqref{ucb2} follows by rearrangement. 
	
	For $a_t$ chosen according to the Thompson sampling, we employ a similar technique that appears in \citep[Theorem 36.1]{lattimore2020bandit}.
	First, note that for any two random variables $Z_t, Z'_t$ with the same mean, 
	\begin{align*}
		\EE[f(\bbeta_t, a_t^*) - f(\bbeta_t, a_t)] = \EE[f(\bbeta_t, a_t^*) - Z_t + Z_t' -f(\bbeta_t, a_t)].
	\end{align*}
	Conditioning on the historical actions $\{a_s\}_{1\leq s\leq t-1}$ and rewards $\{r_s\}_{1\leq s\leq t-1}$ up to $t-1$ (i.e. the $\sigma$-field $\mathcal F_{t-1}$), $a_t$ and $a_t^*$ are the argmax of $f(\bbeta_t, a)$ and $f'(\bbeta_t, a)$, respectively, where $f(\bbeta_t, a)$ and $f'(\bbeta_t, a)$ have the same distribution (i.e. posterior distribution of $f$). 
	As a result, $a_t$ and $a_t^*$ have the same $\mathcal F_{t-1}$-conditional distribution. 
	Now take $Z_t$ and $Z_t'$ as the UCB scores of $a_t^*$ and $a_t$ at $t-1$, respectively:
	\begin{align*}
		&Z_t = \mu_{t-1}(\bbeta_{t}, a_t^*) + c_t^{1/2}\sigma_{t-1}(\bbeta_{t}, a^*_t)&Z'_t = \mu_{t-1}(\bbeta_{t}, a_t) + c_t^{1/2}\sigma_{t-1}(\bbeta_{t}, a_t).
	\end{align*}
	It is easy to verify using the tower property that $\EE[Z_t]=\EE[\EE[Z_t|\mathcal F_{t-1}]] = \EE[\EE[Z'_t|\mathcal F_{t-1}]] = \EE[Z_{t-1}]$. 
	On the other hand, according to \eqref{ou1}, it holds with probability at least $1-2\delta$ that 
	\begin{align*}
		&f(\bbeta_t, a_t^*) - Z_t + Z_t' -f(\bbeta_t, a_t)\leq c_t^{1/2}\sigma_{t-1}(\bbeta_t, a)& t\in\mathbb{N}. 
	\end{align*}
	Using a similar analysis in the UCB case, we conclude that 
	\begin{align*}
		\EE[R_T] = \sum_{t=1}^T\EE[f(\bbeta_t, a_t^*) - Z_t + Z_t' -f(\bbeta_t, a_t)]\lesssim \sqrt{\frac{2^{s}T(\log T)^{d+1}\log\left(2^sT^2\right)}{\log(1+\sigma_0^{-2})}}.
	\end{align*}
	
\end{proof}

\section{Preliminary Study of the Interface Conditions} \label{sect:prelim}
\begin{figure}[H]
	\captionsetup{width=.35\linewidth}
	\centering
	\includegraphics[width=.35\linewidth]{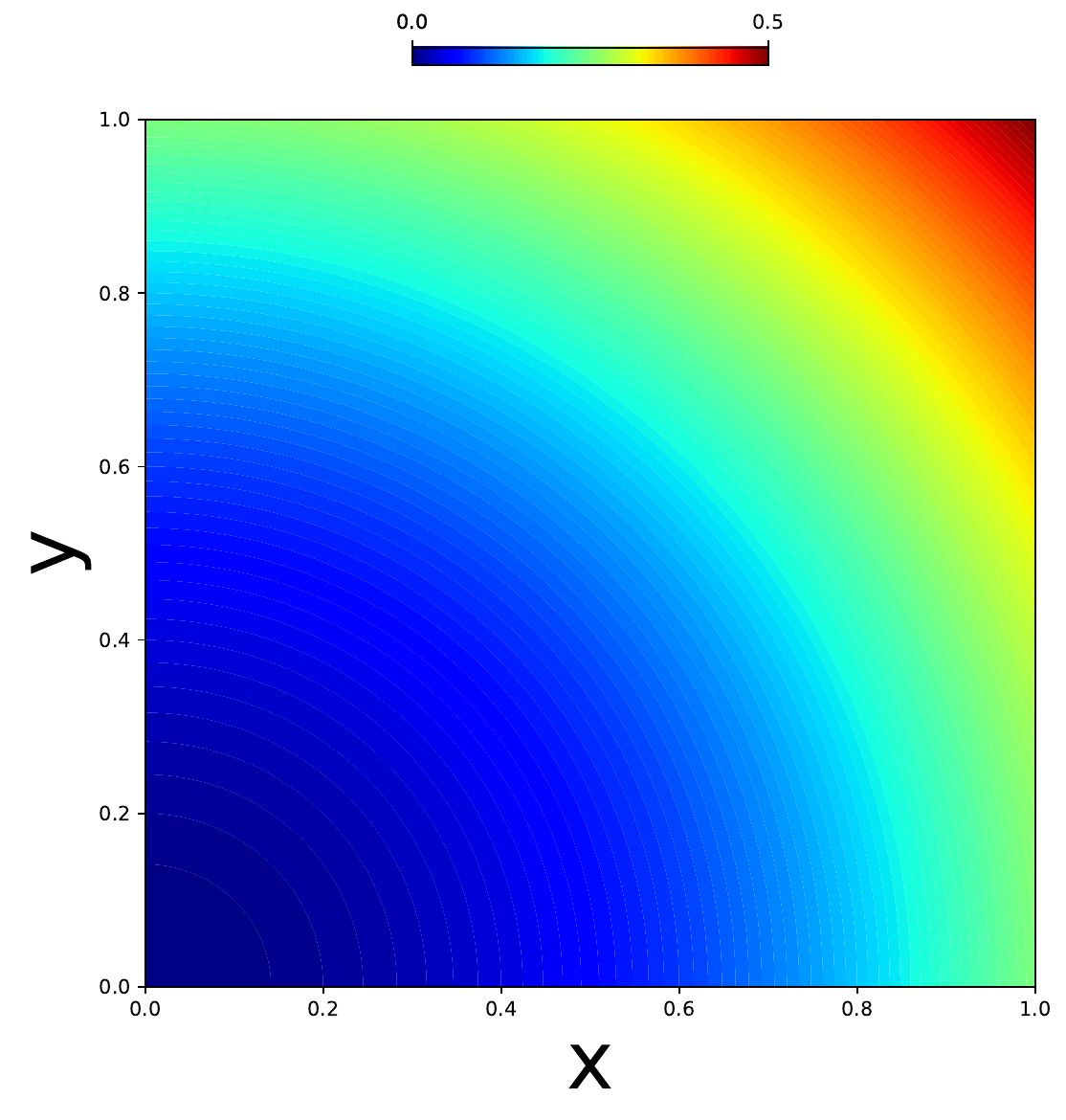}
	\caption{\small Poisson solution.}
	\label{fig:poiss-sol}
\end{figure}

We conducted a preliminary study on a 2D Poisson equation $u_{xx} + u_{yy} = 1$ with the solution shown in Figure \ref{fig:poiss-sol}.% ($x, y \in [0,1] \times [0,1]$).  %Fig. \ref{fig:poiss-force} shows an example of the forcing function. %We consider Consider the 2D Poisson equation $u_{xx} + u_{yy} = 1$ with the solution shown in Figure \ref{fig:poiss-sol} for $x, y \in [0,1] \times [0,1]$. Optimization in this section consists of 500 ADAM epochs (with learning rate $10^-3$) followed by 20K L-BFGS iterations (the first order optimality and parameter change tolerances set to $10^{-9}$ and $10^{-12}$ respectively)
%\textcolor{blue}{need to check if the solution and forcing function match}
\cmt{
\begin{figure}[H]
	\captionsetup{width=.35\linewidth}
	\centering
	\includegraphics[width=.35\linewidth]{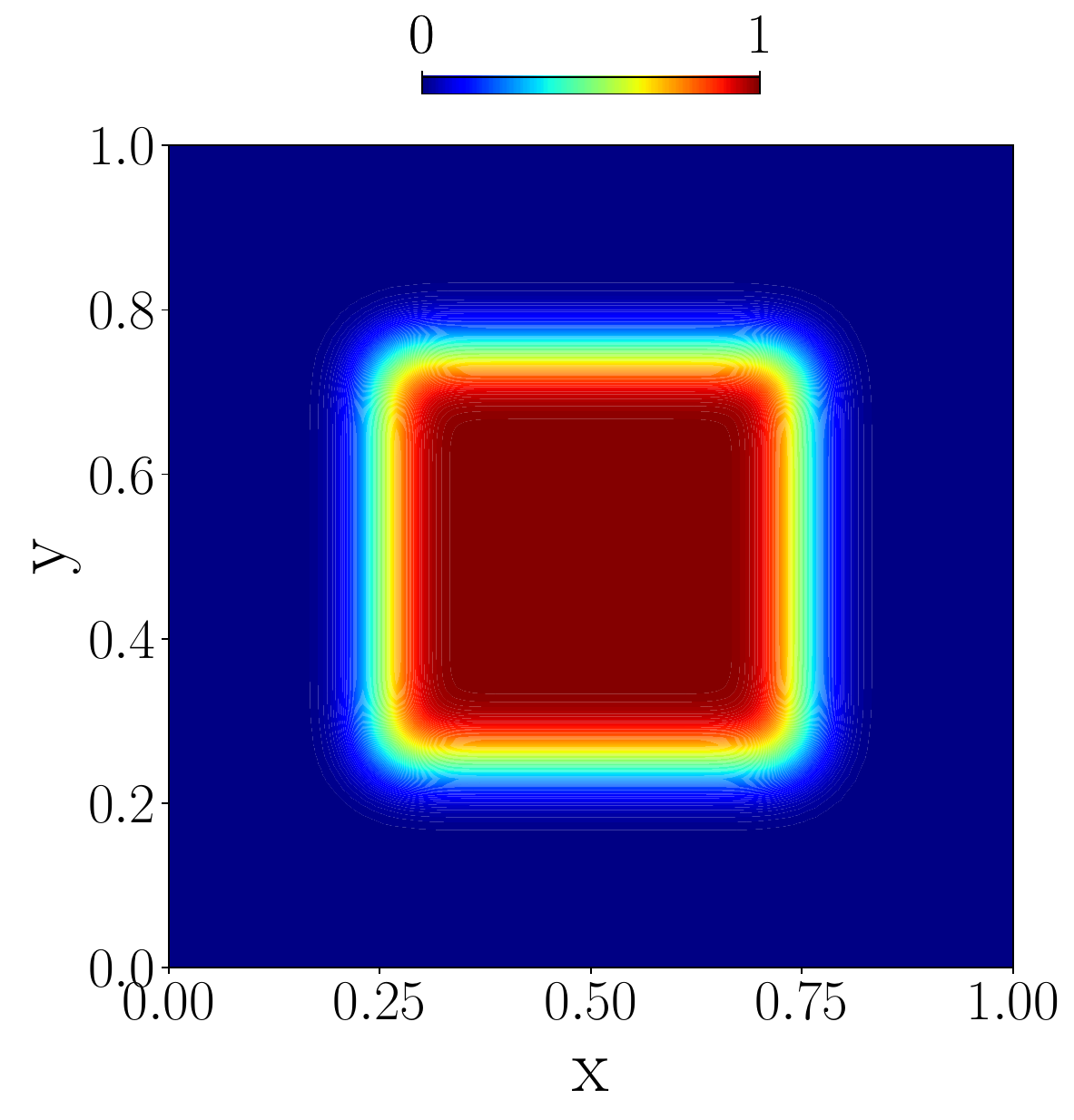}
	\caption{\small Supplementary plot of forcing function $f(x,y;s)$ at $s=20$ for ``Figure 2: Poisson's equation'' in the manuscript.}
	\label{fig:poiss-force}
\end{figure}
}

\begin{figure}[H]
	\captionsetup{width=.9\linewidth}
	\centering
	\setlength\tabcolsep{0pt}
	\begin{tabular}[c]{ccc}
		\setcounter{subfigure}{0}
		\begin{subfigure}[t]{0.3\textwidth}
			\centering
			\includegraphics[width=\textwidth]{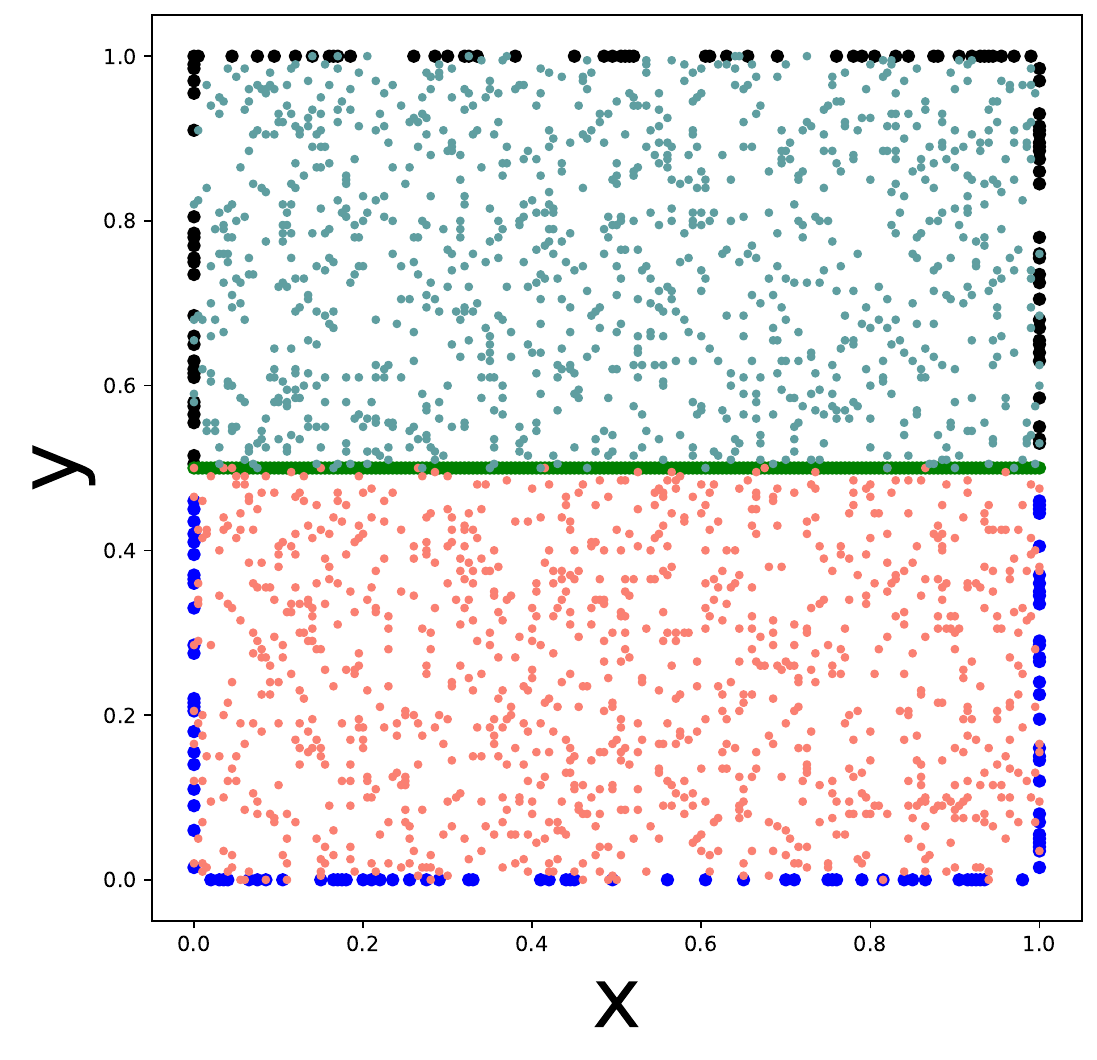}
			\caption{\small Random}
		\end{subfigure} &
		\begin{subfigure}[t]{0.3\textwidth}
			\centering
			\includegraphics[width=\textwidth]{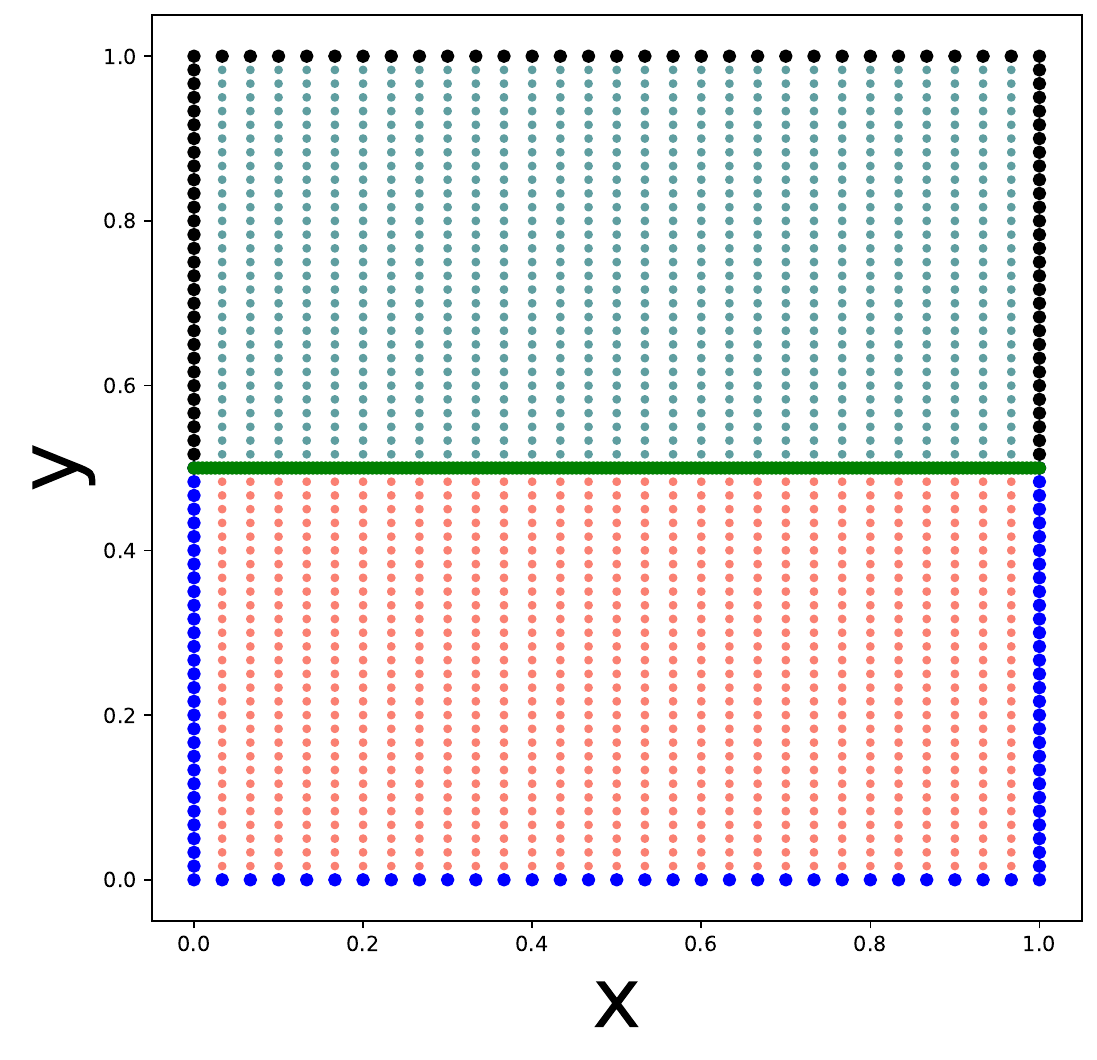}
			\caption{\small Grid}
		\end{subfigure} &
		\begin{subfigure}[t]{0.3\textwidth}
			\centering
			\includegraphics[width=\textwidth]{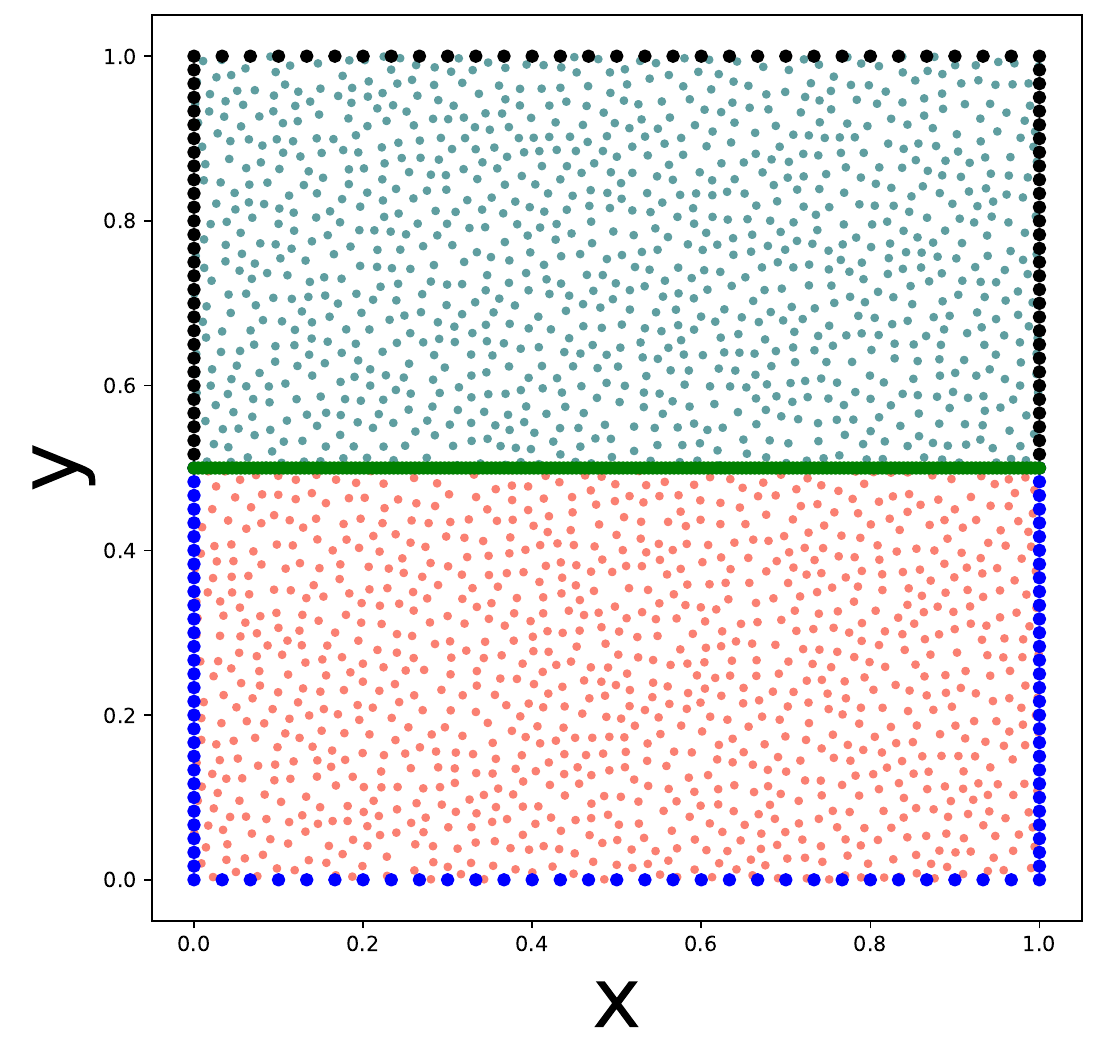}
			\caption{\small Poisson Disc}
		\end{subfigure}
	\end{tabular}
	\caption{\small Random, grid, and Poisson disc  sampling for the Poisson equation problem. The interface is the green line at $y=0.5$. Blue and black dots show the sampled boundary points in each subdomain, and the internal dots (red and cyan) the sampled points inside each subdomain.} \label{fig:poisson-supp-sample}
\end{figure}

%\subsection{Sampling and Loss Term Weights}
\label{sssec:sample}
Given this PDE problem, we compared three types of boundary and collocation point sampling methods: random, grid, and Poisson disc sampling, as shown in Figure \ref{fig:poisson-supp-sample}. The comparison was done between a standard PINN and XPINN, where the number of collocation points in each XPINN subdomain is the same as the total number of collocation points used by the PINN. \cmt{ therefore more points are used in training the decomposition.} We trained the two models with the boundary loss term weight $\lambda_b$ set to  1 and 20. We also varied the interface loss term weight $\lambda_I$ from \{1, 20\}. The interface loss term is computed from \eqref{eq:interface-uavg} and \eqref{eq:interface-rc}. Table \ref{tb:sample} shows the $L_2$ relative error averaged over 10 runs to minimize the variance in network initialization and optimization. %We can see that the performance of PINNs is quite sensitive to the choice of the sampling method and loss term weights. 
We can see that the XPINN performance is relevantly less  variant to differences in sampling and weights, but for PINNs these differences result in order of magnitude changes in error. 
For this reason, we have conducted all the evaluations fairly by using random sampling and larger boundary weights for PINN's which was the best overall setting.  We also make the insight that random sampling allows a PINN to see higher frequencies according to the Nyquist-Shannon sampling theorem which may be the reason for increased performance over the other sampling methods.  The XPINN includes an additional complexity of subdomains and interface conditions which may dominate the training, resulting in less variance as a function of collocation points. 
 %resulting and give less varying performance across all the sampling methods. %The results have shown that differences in hyper-parameters such as sampling and term weights are critical in making fair comparisons between PINNs and domain decomposition methods. %We also conduct all tests using ADAM + L-BFGS optimization schemes which are known to train PINNs to orders of magnitude more accurate than using only ADAM optimization. 

\begin{table}[H]
	\centering
	\begin{tabular}[c]{c | c c | c c | c c}
		\toprule
		\multirow{2}{*}{Model} & \multicolumn{2}{c|}{Random} &  \multicolumn{2}{c|}{Grid} &  \multicolumn{2}{c}{Poisson Disc} \\
		& $\lambda_{b,I}$ = 1 & $\lambda_{b,I}$ = 20 & $\lambda_{b,I}$ = 1 & $\lambda_{b,I}$ = 20 & $\lambda_{b,I}$ = 1 & $\lambda_{b,I}$ = 20 \\
		\hline         
		PINN & 9.0325e-4 & \bf{4.3223e-4} & 6.0278e-3 & 2.3699e-3 & 5.3902e-3 & 2.2375e-3 \\
		XPINN & 5.0884e-3 & 5.3205e-3 & 6.0764e-3 & 4.7829e-3 & 6.5061e-3 & 4.4888e-3 \\
		\bottomrule
	\end{tabular}
	\caption{\small Average $L_2$ relative error over 10 runs for different sampling techniques and loss term weights.} \label{tb:sample}
\end{table}

\subsection{Interface Condition Combination}
\label{sssec:combine}

For the same PDE problem with random sampling, we ran multi-domain PINNs with different sets of conditions. We used the generalized interface condition notations for multi-domain PINNs as described in Table \ref{tb:ic-list}. For example, an XPINN can be described as $I_{u_{avg}} + I_{rc}$. The weights on all terms are unity. As seen by the results in Table \ref{tb:combine}, the multi-domain PINNs with interfaces $I_{u_{avg}} + I_{c} + I_{yy}$ outperforms other combinations as well as the PINN. \cmt{Referring back to Table \ref{tb:sample}, a PINN drastically improves with the correct term weighting} We can see that with the correct interface conditions, the multi-domain PINN can greatly improve upon the standard XPINN.  %In fact, using the additional residual continuity term, a trait of XPINNs, performs slightly better than only using the average solution continuity. 
In fact, the additional residual continuity term, a trait of XPINNs, performs infinitesimally better than only using the average solution continuity. 
These results are the foundation of the METALIC method as we have shown that different combinations of conditions result in drastically different performances. We can also see that multi-domain PINNs are more general and flexible than the existing PINN decomposition models such as XPINN and cPINN. Having only used XPINNs in Table \ref{tb:sample}, one might conclude decomposing this problem is inferior to a standard PINN. However, we have shown that cPINN outperformed XPINN by an order of magnitude and that adding the additional term $I_{yy}$ improved the cPINN even further. Furthermore, naively adding all possible terms such as in the final row, does not necessarily give the best results. This leaves three options for multi-domain PINNs: manually tuning the interface conditions, running all possible permutations such as we have done here, or devise a method to learn the appropriate interfaces such as METALIC. 

\begin{table}[H]
	\centering
	\captionsetup{width=.5\linewidth}
	\begin{tabular}[c]{l | c }
		\toprule
		Model & $L_2$ Relative Error \\
		\hline
		PINN & 1.05e-3 ± 4.38e-4 \\
		$I_{u_{avg}}$ & 4.28e-3 ± 2.63e-3 \\
		$I_{u_{avg}} + I_{rc}$ & 3.92e-3 ± 2.25e-3 \\
		$I_{u_{avg}} + I_{c}$ & 9.45e-4 ± 2.85e-4 \\
		$I_{u_{avg}} + I_{rc} + I_{c}$ & 9.77e-4 ± 3.45e-4 \\
		$I_{u_{avg}} + I_{rc} + I_{gr}$ & 4.57e-3 ± 3.18e-3 \\
		$I_{u_{avg}} + I_{c} + I_{yy}$ & \bf{5.26e-4 ± 1.97e-4} \\
		$I_{u_{avg}} + I_{rc} + I_{gr} + I_{c} + I_{yy}$ & 9.34e-4 ± 3.18e-4 \\
		\bottomrule
	\end{tabular}
	\caption{\small Average $L_2$ relative error over 10 runs for different interface combinations. Note: $I_{c} = I_{y}$ for this problem.} \label{tb:combine}
\end{table}

\section{Meta Learning Result Analysis} \label{sect:analysis}
\begin{figure}[H]
	\captionsetup{width=.75\linewidth}
	\centering
	\includegraphics[width=.75\linewidth]{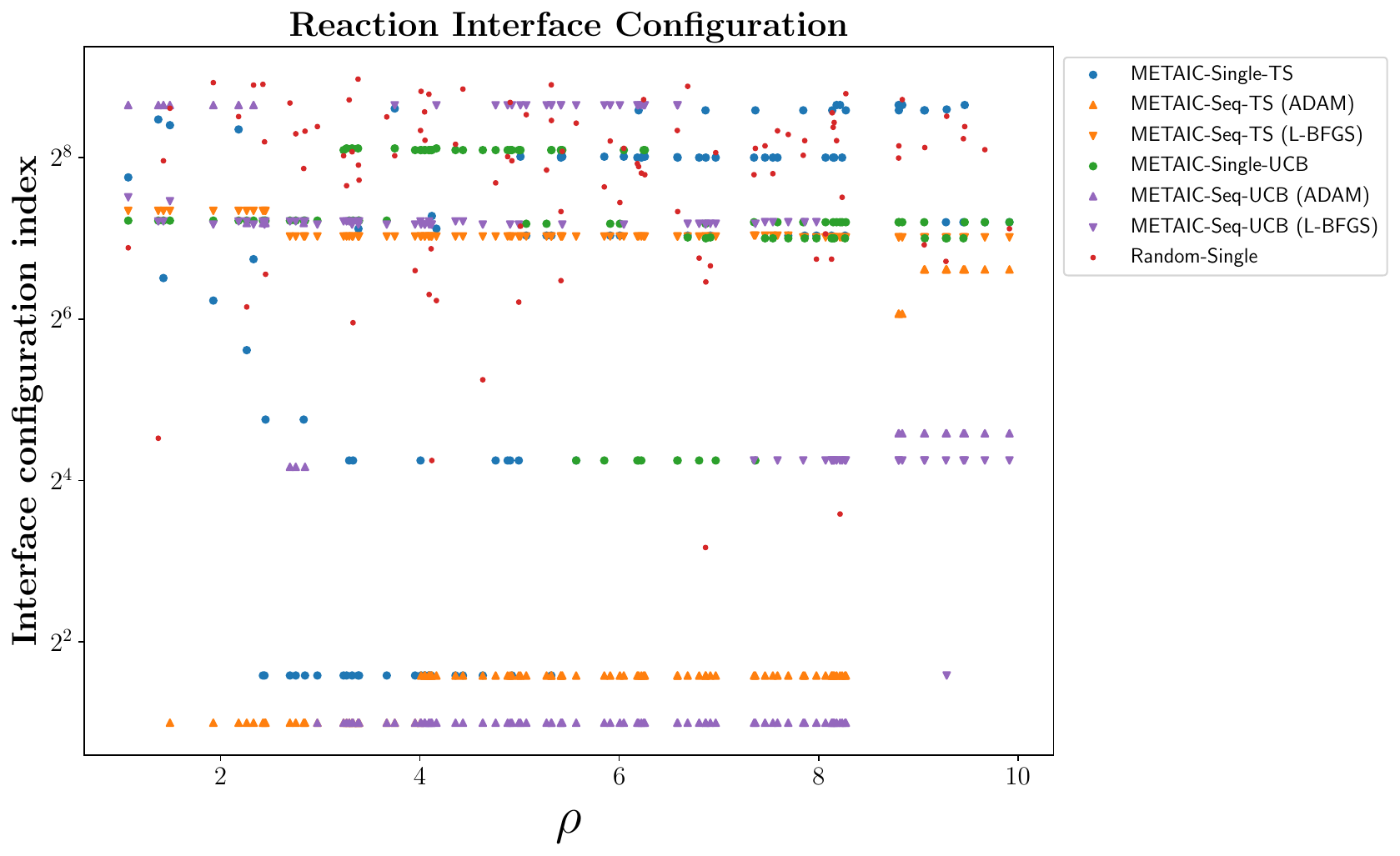}
%	\vspace{-0.25in}
	\caption{\small Scatter plot of interface configuration \textit{vs.} the equation parameter $\rho$.}
	\label{fig:reac1}
\end{figure}

\begin{figure}[H]
	\captionsetup{width=.75\linewidth}
	\centering
	\includegraphics[width=.75\linewidth]{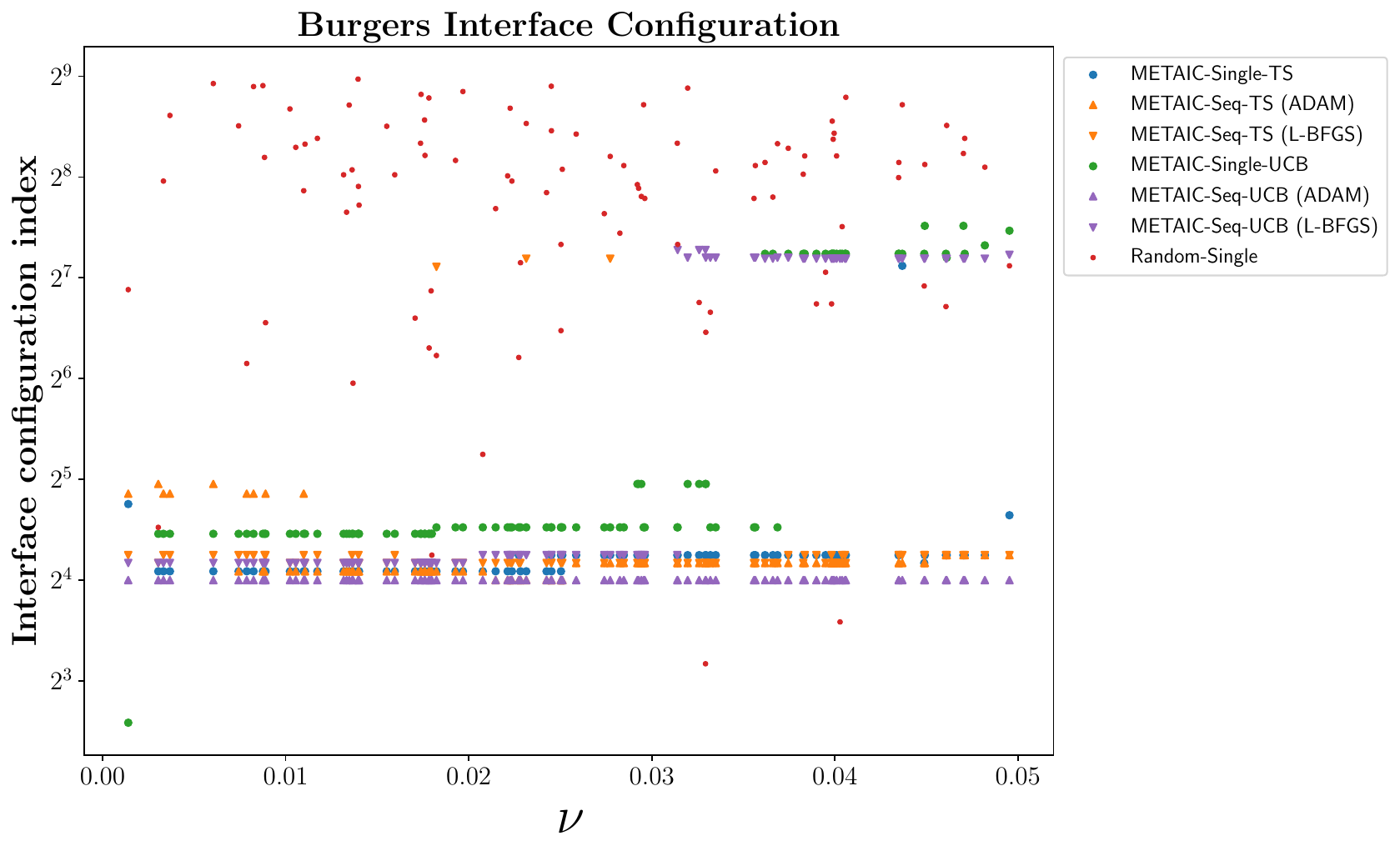}
%	\vspace{-0.25in}
	\caption{\small Scatter plot of interface configuration \textit{vs.} the equation parameter $\nu$.}
	\label{fig:burg1}
\end{figure}

There are $2^9 = 512$ possible combinations of the interface conditions.   For convenience, we use an integer to index each configuration (combination), $
\text{index} = \sum_{i = 0}^{n} 2^{i} \c[i]$ where $\c$ is a list of binaries, and $\c[i] = 1$ means $i$-th interface condition is turned on. We therefore can show how different sets of interface conditions are selected along with the equation parameters (see Figures \ref{fig:poiss1}, \ref{fig:advec1}, \ref{fig:reac1}, and \ref{fig:burg1}). 
%To conveniently examine the selection of these combinations, we use 
%To obtain the interface configuration index in Figures [\ref{fig:poiss1}, \ref{fig:advec1}, \ref{fig:reac1}, \ref{fig:burg1}], we convert the binary list of interface terms to an interface configuration index defined by $
%index = \sum_{i = 0}^{n} 2^{i} \boldsymbol{x}[i]$ where $\boldsymbol{x}$ is a list of binaries.

%\subsection{Supplementary Poisson Forcing Plot}

\subsection{Poisson Equation}
\begin{figure}[H]
%	\captionsetup{width=.75\linewidth}
	\centering
	\includegraphics[width=.75\linewidth]{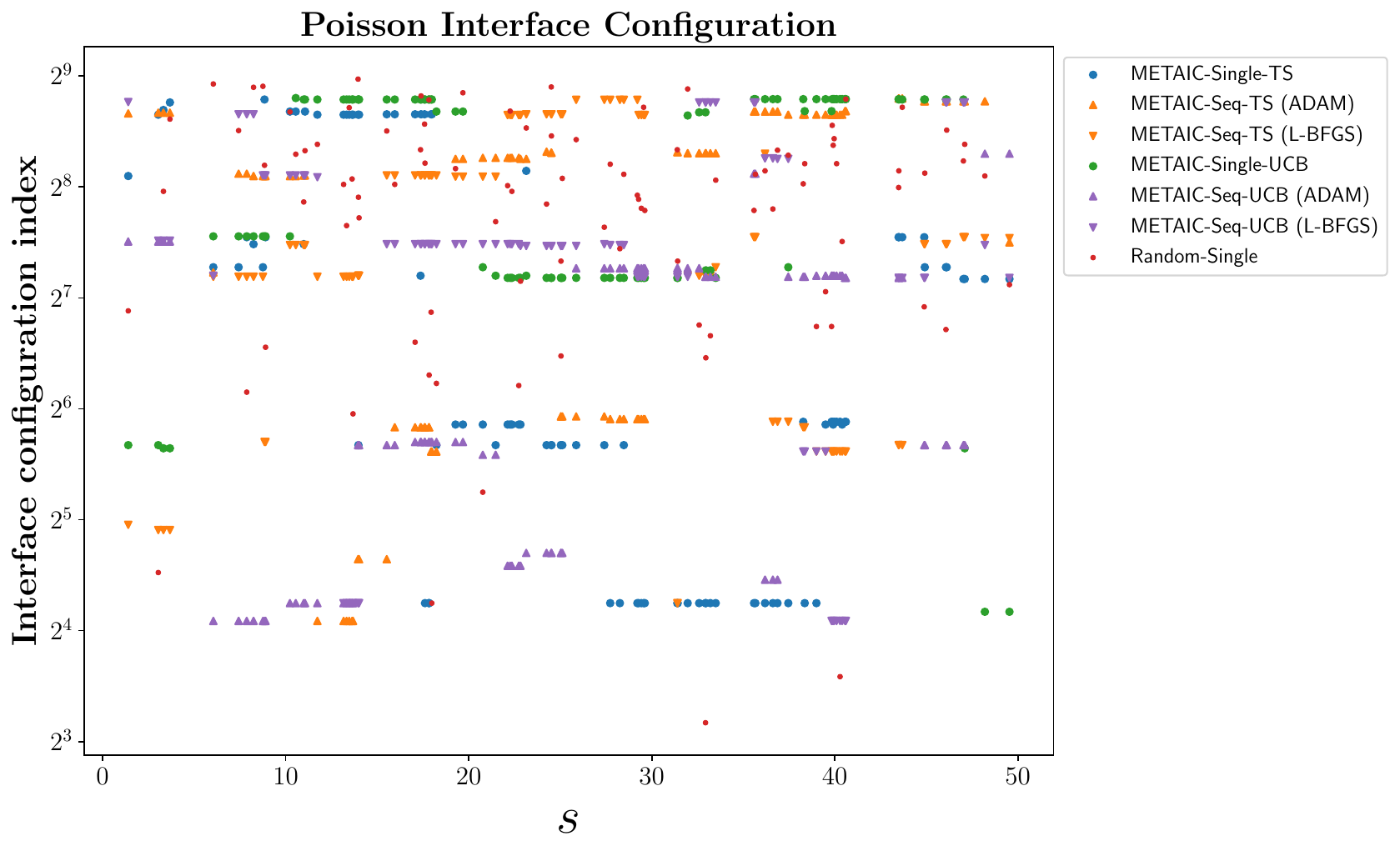}
	%\vspace{-0.25in}
	\caption{\small Scatter plot of interface configuration \textit{vs.} the equation parameter $s$.}
	\label{fig:poiss1}
\end{figure}

For each PDE test case, we provide three analysis plots to better understand the METALIC results. For the Poisson problem, Figure \ref{fig:poiss1} provides an overview of the interface configuration groupings as the equation parameter $s$ varies. As opposed to Random-Single, the various METALIC methods predict interface configurations in groupings based on parameter $s$. This indicates that for these ranges, the PDE solution behaves similarly across the interfaces. It can also be seen that the configurations chosen between METALIC-Single and METALIC-Seq are different, indicating that the optimization is an important factor. This is logical since at the beginning of training, the PINN must first propagate information from the initial and boundary conditions inward to the entire domain. Therefore, interface conditions during this phase may in fact make learning more difficult in terms of the loss landscape as the network is trying to enforce continuity at a location which has no information but is simply a set of random predictions given the random initialization of weights and bias of the network. 

\begin{figure}[H]
	%\captionsetup{width=.75\linewidth}
	\centering
	\includegraphics[width=.75\linewidth]{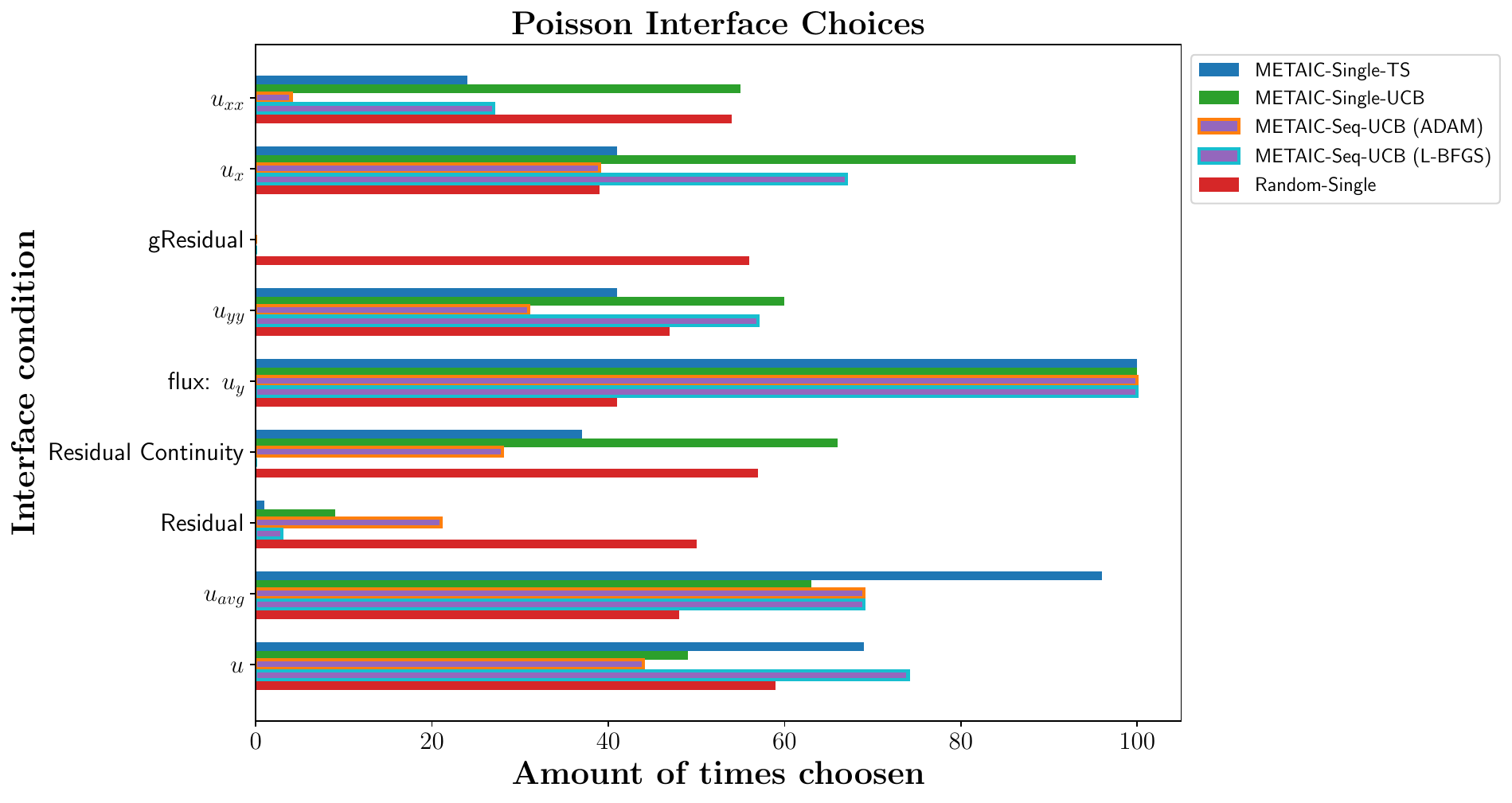}
	%\vspace{-0.25in}
	\caption{\small Horizontal bar plot of the quantity of interfaces chosen throughout testing over 100 randomly drawn equation parameters.}
	\label{fig:poiss2}
\end{figure}

In Figure \ref{fig:poiss2}, we can see the number of times the interface conditions are selected over the 100 test cases. Random-Single serves as a baseline with each interface being chosen roughly half of the time. The two most noticeable trends are that the gradient-enhanced residual term is almost never chosen and the flux continuity which is equivalent to $u_y$ for this case is always chosen by METALIC. This is interesting as the gradient term in the original gPINN paper was shown to be beneficial to PINN training but appears to be a poor choice on a set of interface points, possibly because with all the other terms, it simply make the loss landscape more complex and does not provide a significant accuracy benefit compared to the other more theoretically sound terms such as flux. This result is novel as it showcases the robustness of METALIC in being able to distinguish between valid and invalid terms, something that would take a user doing manual tuning of these terms much trail and error to determine. We also note that the METALIC choices align with our results in Section \ref{sssec:combine} that the flux conditions from cPINN greatly outperforms the residual continuity conditions in XPINN for Poisson's equation. Flux continuity is a well studied conservation term rooted in traditional methods whereas residual continuity is a term devised with the convenience  of PINNs and automatic-differentiation(AD) in mind. We also note that when comparing the METALIC-Seq-UCB ADAM and L-BFGS choices, L-BFGS uses more terms on average than ADAM. This confirms our hypothesis from Figure \ref{fig:poiss1} that more interface terms at the start of training may in-fact make training more difficult. This validates the result that not only is a sequential interface predictor more accurate, but also faster as it adds in terms when needed which would reduce computational cost. We also not that including the residual points in the overall set of collocation points is rarely chosen, likely due to the fact that the interface point set is an order of magnitude smaller than the collocation point set so assuming it is well sampled, its contribution is negligible. All these insights further confirm the method is working well and is consistent with our intuition and the properties of the equation being solved. 

\begin{figure}[H]
	%\captionsetup{width=.75\linewidth}
	\centering
	\includegraphics[width=.75\linewidth]{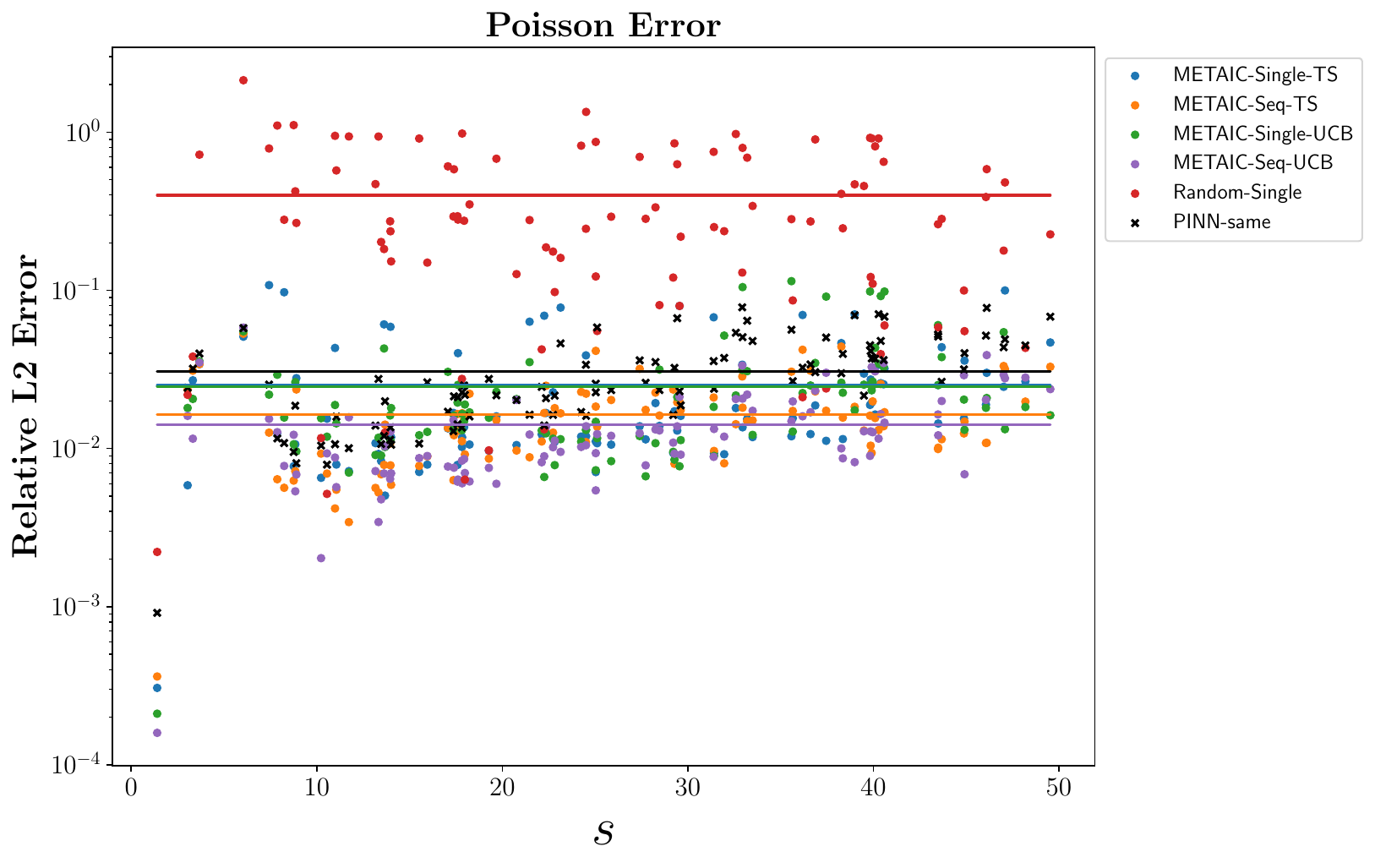}
%	\vspace{-0.25in}
	\caption{\small Scatter plot of the relative $L_2$ error \textit{vs.} the equation parameter $s$.}
	\label{fig:poiss3}
\end{figure}

Finally, in Figure \ref{fig:poiss3}, we show the $L^2$ relative error as a function of $s$. This is a more detailed version of the error table in the manuscript which tells us how the problem difficulty changes over the parameterization of the problem. For this Poisson problem, it is quite consistent other than the lower bound around $s=1$ in which the forcing term is very smooth and as we expect the problem is quite simple, as reflected by the lower errors there. 

\subsection{Advection Equation}
\begin{figure}[H]
	\captionsetup{width=.75\linewidth}
	\centering
	\includegraphics[width=.75\linewidth]{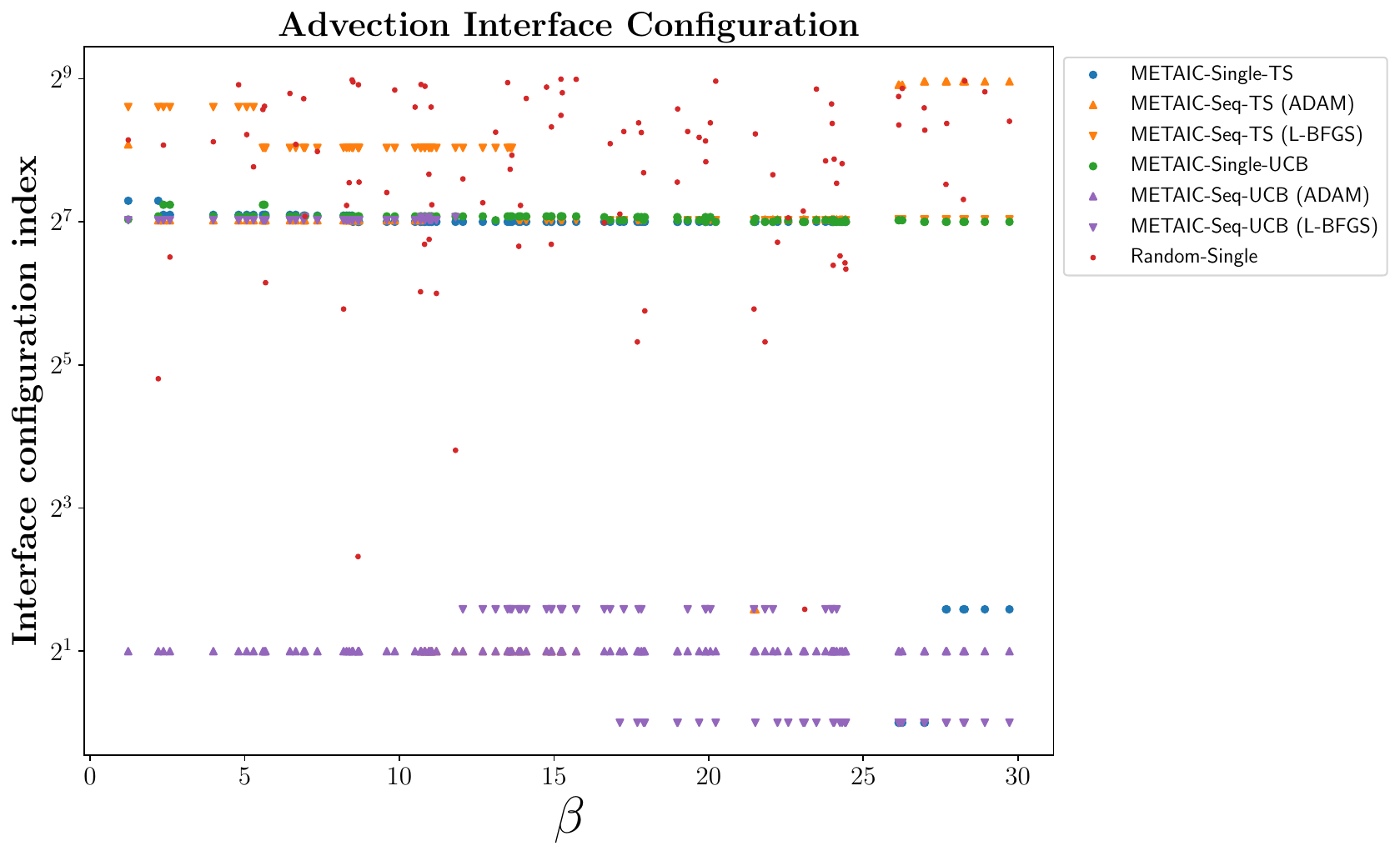}
%	\vspace{-0.25in}
	\caption{\small Scatter plot of interface configuration \textit{vs.} the equation parameter $\beta$.}
	\label{fig:advec1}
\end{figure}

\begin{figure}[H]
	\captionsetup{width=.75\linewidth}
	\centering
	\includegraphics[width=.75\linewidth]{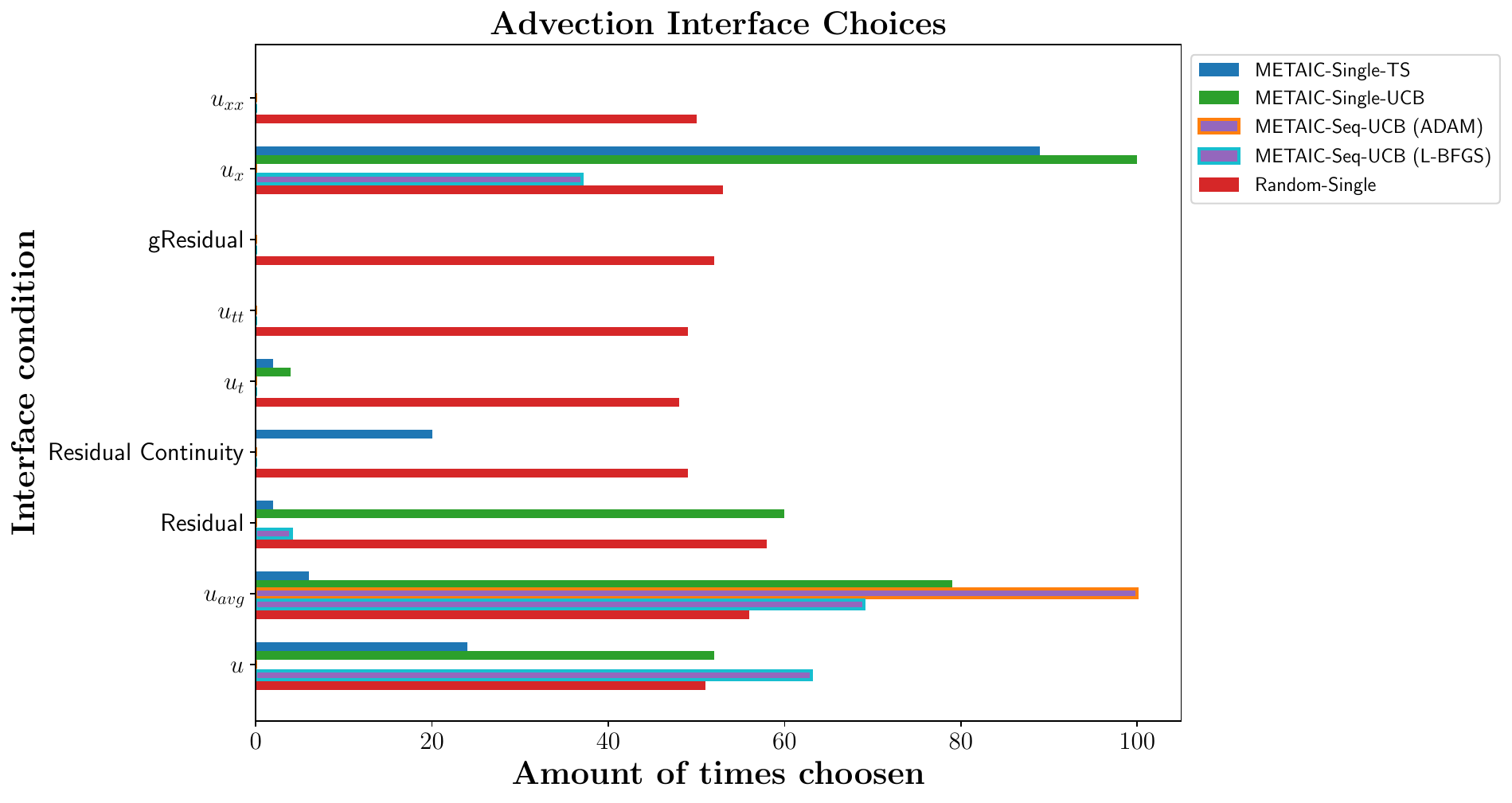}
%	\vspace{-0.25in}
	\caption{\small Horizontal bar plot of the quantity of interfaces chosen throughout testing over 100 randomly drawn equation parameters.}
	\label{fig:advec2}
\end{figure}

\begin{figure}[H]
	\captionsetup{width=.75\linewidth}
	\centering
	\includegraphics[width=.75\linewidth]{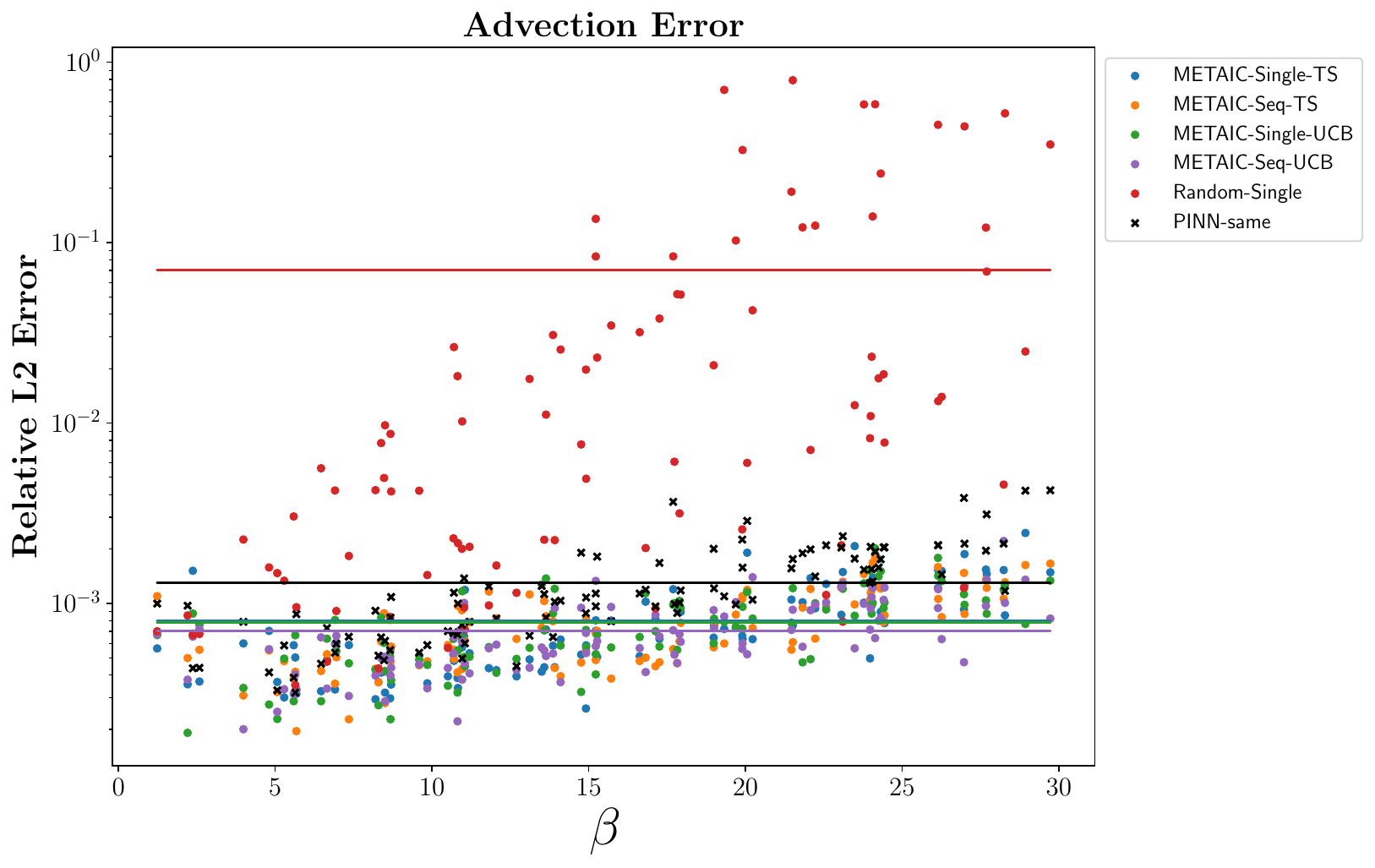}
	%\vspace{-0.25in}
	\caption{\small Scatter plot the relative $L_2$ error \textit{vs.} the equation parameter $\beta$.}
	\label{fig:advec3}
\end{figure}

For the Advection problem, Figure \ref{fig:advec2} indicates that  very few terms where needed. So much so that the ADAM step of METALIC-Seq-UCB has only one term, $I_{u_{avg}}$, the weaker form of the solution continuity. We again point out the benefit of METALIC in being able to sub-select few terms out of many while still resulting in the best accuracy as seen in Figures \ref{fig:advec1} \& \ref{fig:advec3} which show that METALIC-Seq-UCB uses the fewest number of terms but has the best error. This emphasizes that more interface terms are not always better since the loss landscape can become more complex from an optimization standpoint. Another interesting feature is that the first derivative in space ($u_x$) is chosen more than the first derivative in time ($u_t$) despite the subdomain split being in time. This is opposite of the Poisson results in which the derivative normal to the interface ($u_y$), representing the flux, was chosen in all cases. Both terms, $u_t$ and $u_x$ are part of the PDE with flux simply being $u$, but the tangential derivative $u_x$ appears to be a much more meaningful term when it comes to propagating the wave through the interface. There are also no second order terms which was the case with Poisson. 

\subsection{Reaction Equation}
\begin{figure}[H]
	\captionsetup{width=.75\linewidth}
	\centering
	\includegraphics[width=.75\linewidth]{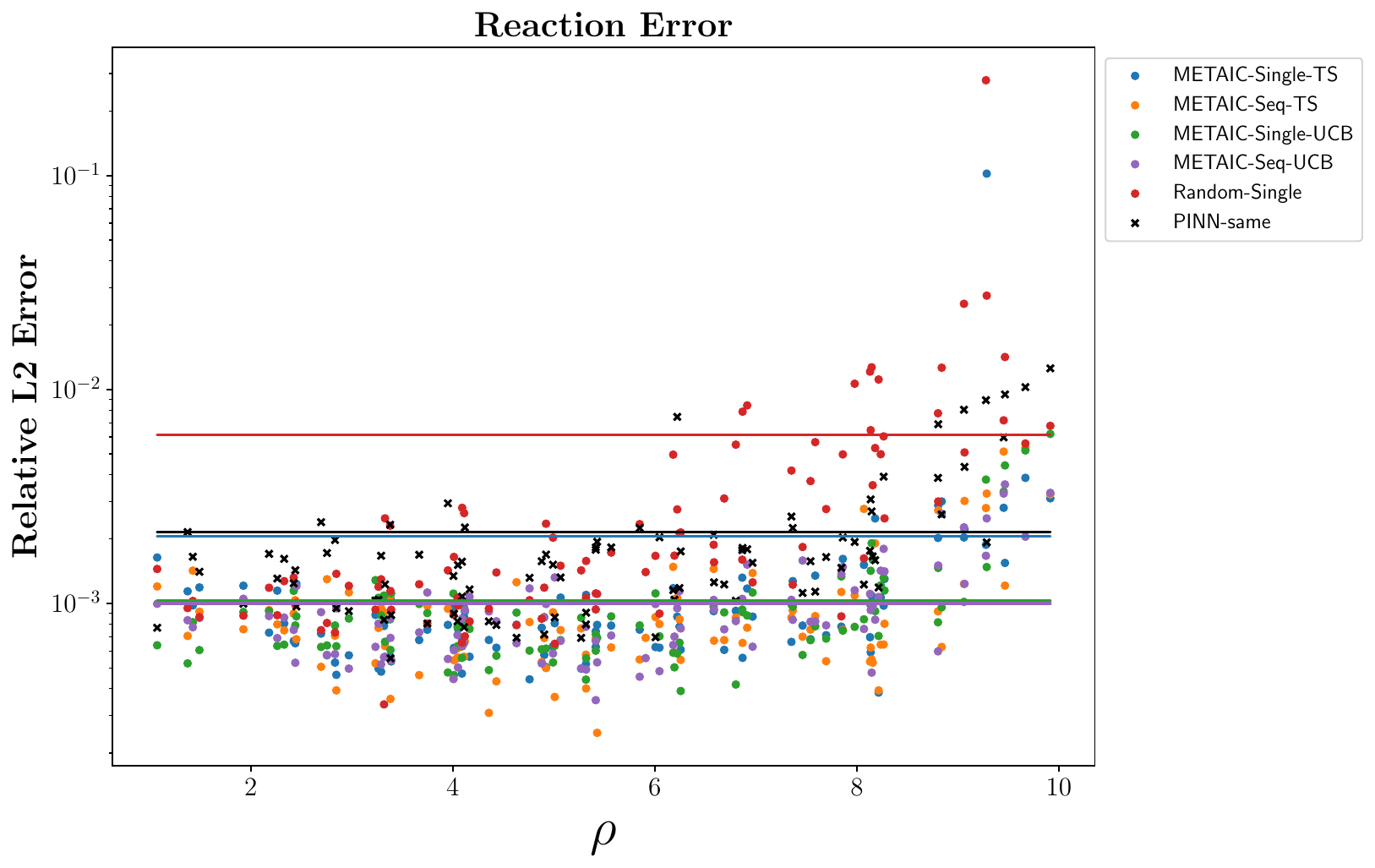}
%	\vspace{-0.25in}
	\caption{\small Scatter plot of the relative $L_2$ error \textit{vs.} the equation parameter $\rho$.}
	\label{fig:reac3}
\end{figure}

For the Reaction problem, we note that while this is an ODE, with no spatial derivatives, they were counter-intuitively chosen as interfaces. This emphasizes the fact that PINNs, and machine learning techniques in general, do not work the same as traditional methods since these terms are not necessary for well-posedness of an ODE. Given this, Figure \ref{fig:reac3} shows that METALIC outperformed the PINN while using these conditions. This is an interesting line of investigation for future work as it shows counter-intuitive terms can provide a training benefit to PINNs even in contrast to the previous Advection problem where almost no terms where chosen. It is not clear why in some cases only the most basic of terms are used while in others terms which do not make physical sense are chosen, but in both, the accuracy is quite good on their respective problems. This shows that METALIC learns something about PINN training that is not evidently clear to the human user.

\begin{figure}[H]
	\captionsetup{width=.75\linewidth}
	\centering
	\includegraphics[width=.75\linewidth]{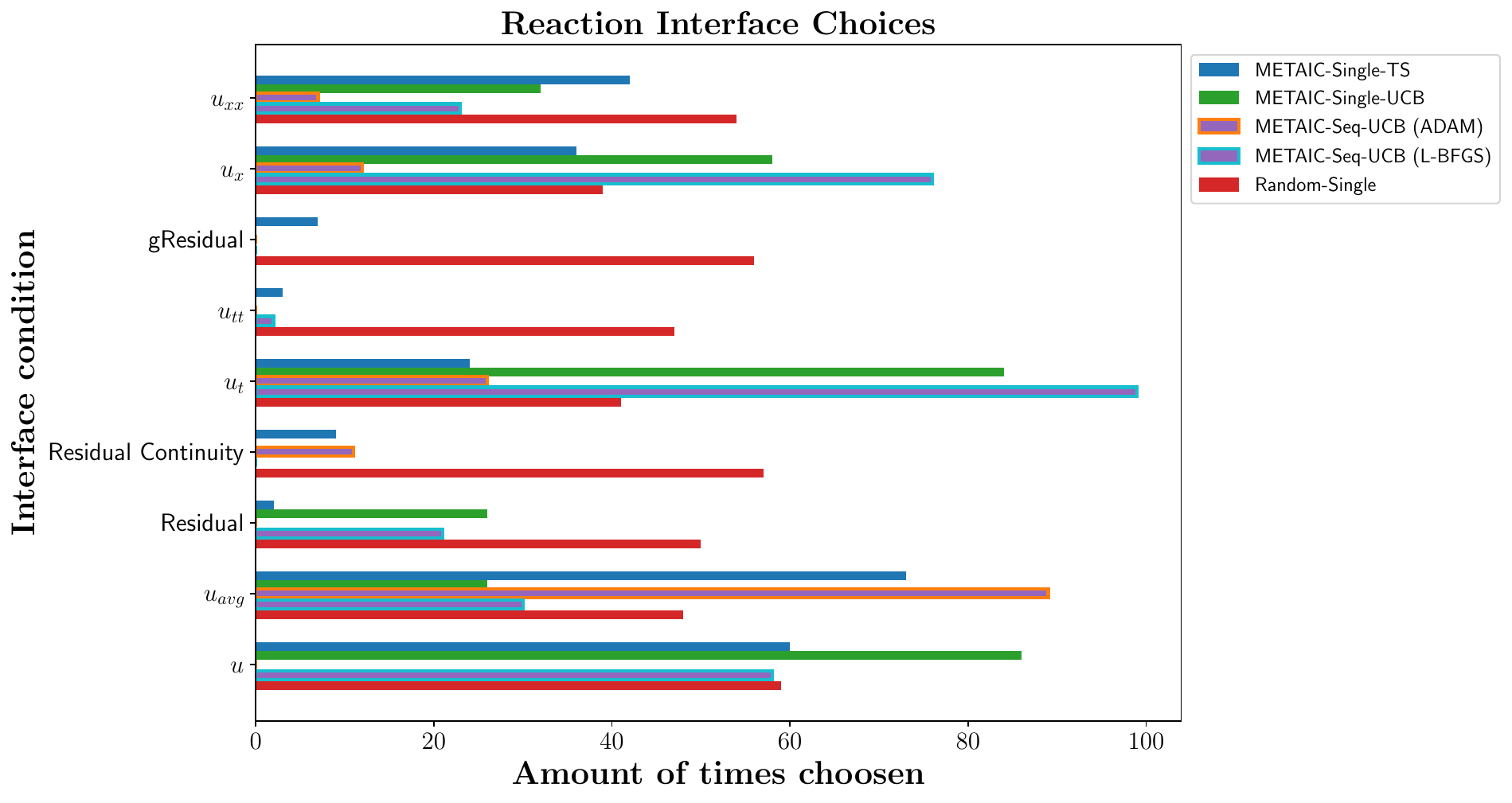}
%	\vspace{-0.25in}
	\caption{\small Horizontal bar plot of the quantity of interfaces chosen throughout testing over 100 randomly drawn equation parameters.}
	\label{fig:reac2}
\end{figure}

\subsection{Burger's Equation}
\begin{figure}[H]
	\captionsetup{width=.75\linewidth}
	\centering
	\includegraphics[width=.75\linewidth]{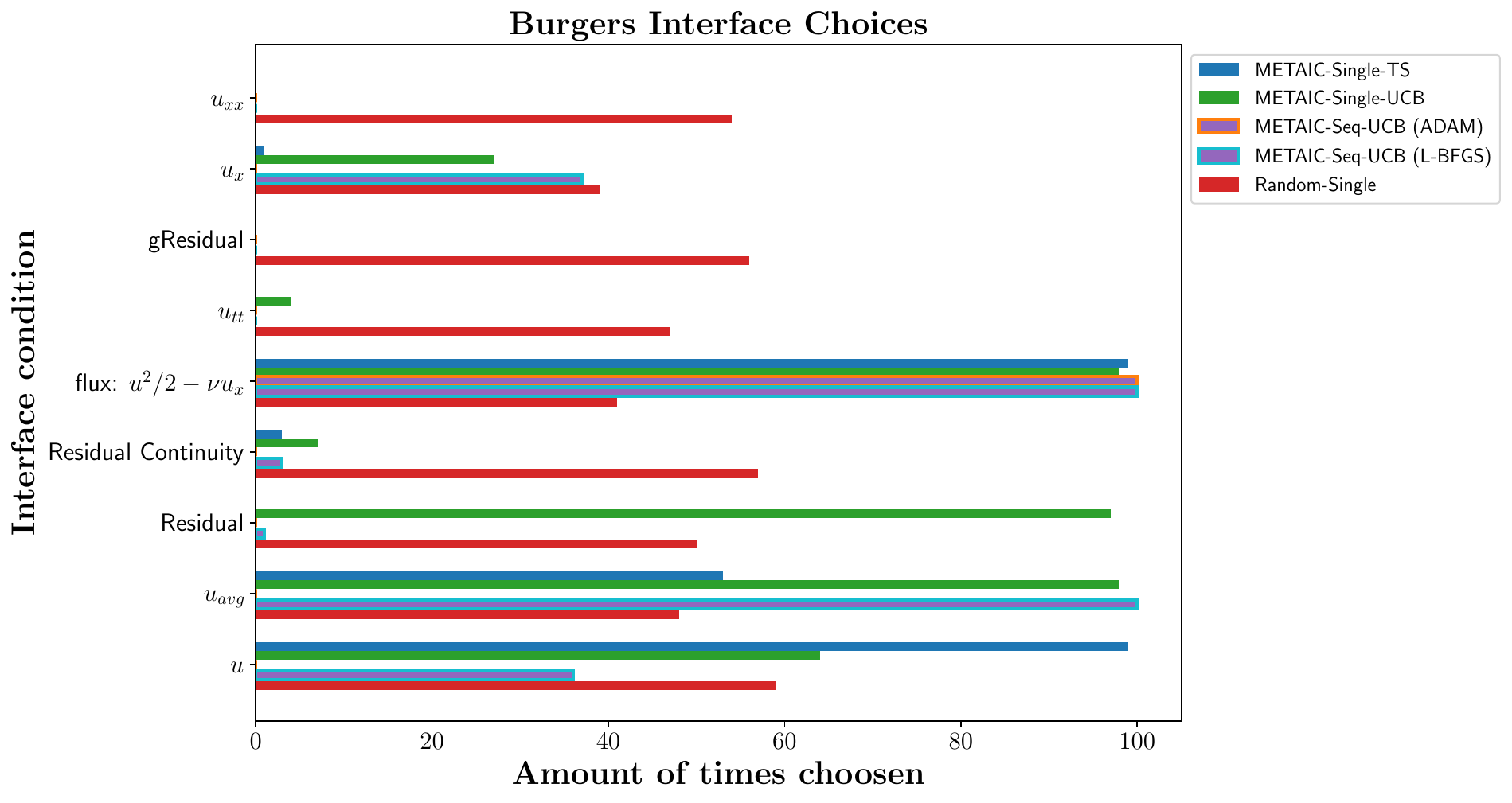}
	%\vspace{-0.25in}
	\caption{\small Horizontal bar plot of the quantity of interfaces chosen throughout testing over 100 randomly drawn equation parameters.}
	\label{fig:burg2}
\end{figure}

\begin{figure}[H]
	\captionsetup{width=.75\linewidth}
	\centering
	\includegraphics[width=.75\linewidth]{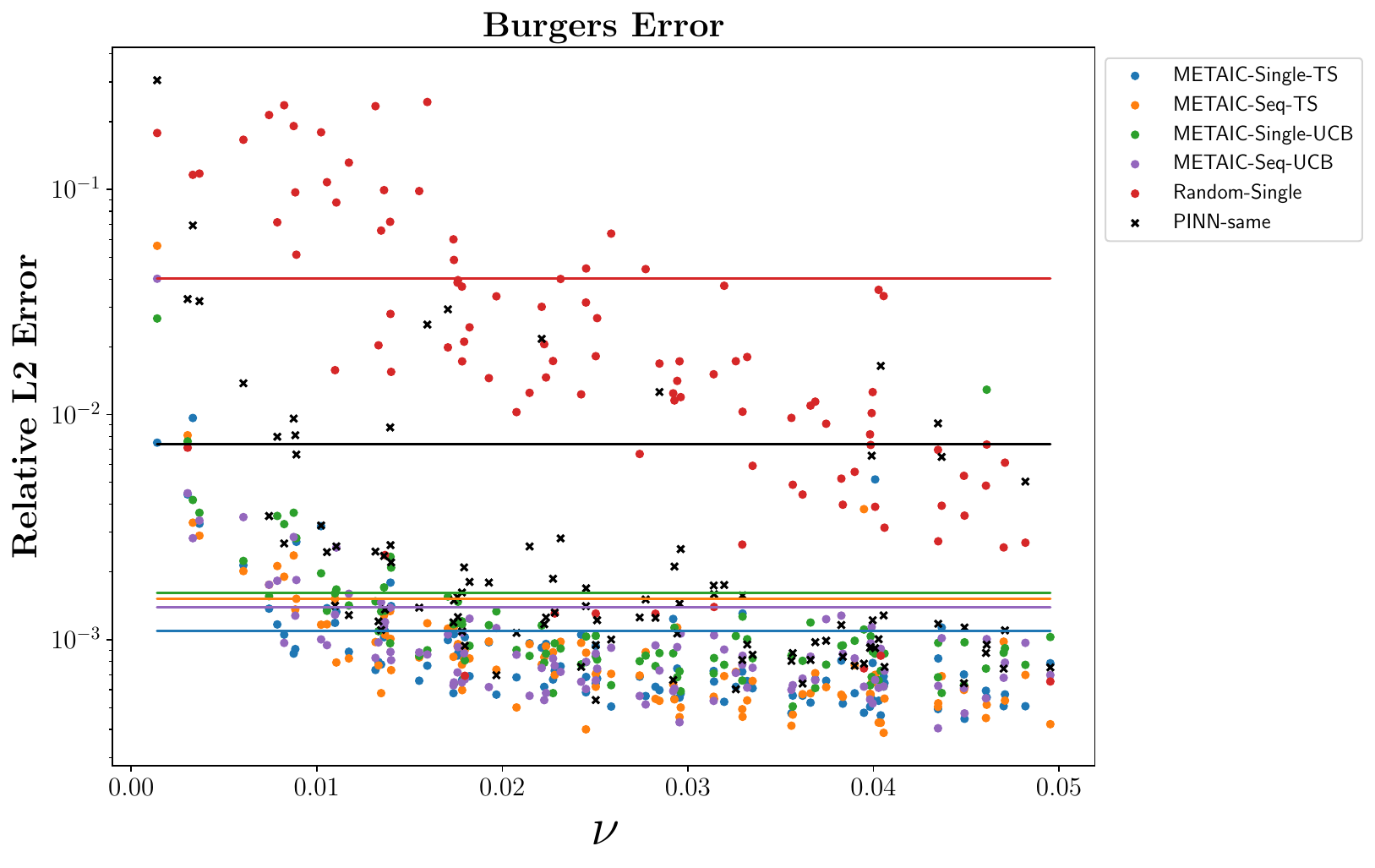}
%	\vspace{-0.25in}
	\caption{\small Scatter plot of the relative $L_2$ error \textit{vs.} the equation parameter $\nu$.}
	\label{fig:burg3}
\end{figure}

For the Burger's problem, we see more of what one might expect from traditional interface continuity terms. Figure \ref{fig:burg2} shows that the flux term is predominately chosen, just as in Poisson's equation. Although here we see the flux is not equivalent to the first order derivative, enforcing the idea that it is in fact the flux providing the training benefit and not a coincidence of the first-order derivative and flux being the same for Poisson. We also see the largest improvement in error of METALIC over PINNs as seen in Figure \ref{fig:burg3}. The trend is also consistent with our physical understanding, as viscosity ($\nu$) increases the problem becomes more simple. This is because at lower viscosities a shock forms and creates a discontinuity in the solution which is difficult for PINNs to resolve. The decomposition of this problem is therefore the most sound, in that we allow one network to handle the sharp discontinuity in the center, and another to handle the relatively simple solution around it. This allows for the network in the center to learn a higher frequency basis with which to approximate the discontinuity instead of also having to fit the lower frequencies around it which has been delegated to the second network. Given this, it makes sense that a multi-domain PINN with the METALIC method greatly outperform PINNs here. It also emphasizes that in the less physically motivated decompositions for Poisson, Advection, and Reaction, we still see improvement using multi-domain PINNs and METALIC. 

\vfill

% % Note: in this sample, the section number is hard-coded in. Following
% % proper LaTeX conventions, it should properly be coded as a reference:

% %In this appendix we prove the following theorem from
% %Section~\ref{sec:textree-generalization}:

% In this appendix we prove the following theorem from
% Section~6.2:

% \noindent
% {\bf Theorem} {\it Let $u,v,w$ be discrete variables such that $v, w$ do
% not co-occur with $u$ (i.e., $u\neq0\;\Rightarrow \;v=w=0$ in a given
% dataset $\dataset$). Let $N_{v0},N_{w0}$ be the number of data points for
% which $v=0, w=0$ respectively, and let $I_{uv},I_{uw}$ be the
% respective empirical mutual information values based on the sample
% $\dataset$. Then
% \[
% 	N_{v0} \;>\; N_{w0}\;\;\Rightarrow\;\;I_{uv} \;\leq\;I_{uw}
% \]
% with equality only if $u$ is identically 0.} \hfill\BlackBox

% \noindent
% {\bf Proof}. We use the notation:
% \[
% P_v(i) \;=\;\frac{N_v^i}{N},\;\;\;i \neq 0;\;\;\;
% P_{v0}\;\equiv\;P_v(0)\; = \;1 - \sum_{i\neq 0}P_v(i).
% \]
% These values represent the (empirical) probabilities of $v$
% taking value $i\neq 0$ and 0 respectively.  Entropies will be denoted
% by $H$. We aim to show that $\fracpartial{I_{uv}}{P_{v0}} < 0$....\\

% {\noindent \em Remainder omitted in this sample. See http://www.jmlr.org/papers/ for full paper.}

% \vskip 0.2in
% \bibliography{sample}

% \bibliographystyle{apalike}
\bibliography{XPINNMAB}

\end{document}